\documentclass[11pt, a4paper, logo]{deepmind}

\usepackage{graphicx}\usepackage{multirow}\usepackage{amsmath,amssymb,amsfonts}\usepackage{amsthm}\usepackage{mathrsfs}\usepackage[title]{appendix}\usepackage{xcolor}\usepackage{textcomp}\usepackage{manyfoot}\usepackage{booktabs}\usepackage{algpseudocode}\usepackage{listings}\usepackage{array}

\usepackage{cleveref}
\usepackage{subfig}
\usepackage[boxruled,vlined,noend]{algorithm2e}
\usepackage{float}
\newcommand{\refsecbyname}[1]{\Cref{#1}}

\begin{abstract}
    Privacy-preserving machine learning aims to train models on private data without leaking sensitive information.
Differential privacy (DP) is considered the gold standard framework for privacy-preserving training, as it provides formal privacy guarantees. However, compared to their non-private counterparts, models trained with DP often have significantly reduced accuracy. Private classifiers are also believed to exhibit larger performance disparities across subpopulations, raising fairness concerns.
The poor performance of classifiers trained with DP has prevented the widespread adoption of privacy preserving machine learning in industry.
Here we show that pre-trained foundation models fine-tuned with DP can achieve similar accuracy to non-private classifiers, even in the presence of significant distribution shifts between pre-training data and downstream tasks. We achieve private accuracies within a few percent of the non-private state of the art across four datasets, including two medical imaging benchmarks. Furthermore, our private medical classifiers do not exhibit larger performance disparities across demographic groups than non-private models.
This milestone to make DP training a practical and reliable technology has the potential to widely enable machine learning practitioners to train safely on sensitive datasets while protecting individuals' privacy. \end{abstract}

\begin{document}

\title{Unlocking Accuracy and Fairness in Differentially Private Image Classification}

\renewcommand{\today}{}

\author[*,1]{Leonard Berrada}
\author[*,1]{Soham De}
\author[*,2,$\dagger$]{Judy Hanwen Shen}
\author[1]{Jamie Hayes}
\author[1]{Robert Stanforth}
\author[1]{David Stutz}
\author[1]{Pushmeet Kohli}
\author[1]{Samuel L. Smith}
\author[1]{Borja Balle}

\affil[*]{Equal contributions}
\correspondingauthor{ \{lberrada | bballe\}@google.com}
\affil[1]{Google DeepMind, London, UK}
\affil[2]{Computer Science Department, Stanford University, Palo Alto, California, USA}
\affil[$\dagger$]{Work done while at Google DeepMind}

\maketitle

\section{Introduction}

Neural networks containing hundreds of millions or even billions of parameters achieve state-of-the-art performance across a wide range of machine learning tasks \cite{DBLP:journals/nature/LeCunBH15,DBLP:conf/nips/BrownMRSKDNSSAA20,zhai2022scaling}. However, these large models are known to memorize parts of their training data \cite{DBLP:journals/cacm/ZhangBHRV21}, opening the door to leakage of sensitive information. Privacy attacks capable of extracting memorized training data have been demonstrated on language models \cite{DBLP:conf/uss/CarliniTWJHLRBS21}, diffusion models \cite{DBLP:journals/corr/abs-2301-13188} and image classification models \cite{balle2022reconstructing}, while membership attacks which detect whether a datapoint was used to train a particular model are successful on multiple architectures and data modalities \cite{DBLP:conf/sp/CarliniCN0TT22}. Mitigating the leakage of sensitive training data is a critical concern in applications where access to private, confidential or proprietary data is key to improving machine learning capabilities (e.g. healthcare, recommendation systems, mobility, etc). Models trained on such data cannot be safely deployed unless these concerns are addressed.

Designing effective mitigations to prevent leakage of sensitive training data is far from straightforward. The failures of classical anonymization techniques against re-identification attacks when adversaries have access to side-knowledge \cite{ohm2009broken} highlight the importance of considering exceedingly pessimistic threat models to ensure privacy mitigations are future-proof. Differential privacy (DP) \cite{dwork2006calibrating} has emerged as the gold standard for protecting individual privacy in data processing algorithms, including machine learning methods. The information-theoretic guarantee provided by DP limits the amount of information any attacker will be able to extract about individual input datapoints after observing the algorithm's output, regardless of what side knowledge about the data they obtain from other sources or how much computational power they have access to. The increasing adoption of DP across governmental \cite{uscensusreport} and industrial \cite{rapporblog,appleblog,flblog} applications underlines its versatility and stems, in part, from its ability to provide quantifiable worst-case privacy guarantees. However, there is no free lunch: models with strong DP guarantees achieve lower accuracy than their non-private counterparts.

\begin{figure}[htpb]
    \centering
    \hspace{ \stretch{1} }
    \begin{minipage}{.35\textwidth}
    \includegraphics{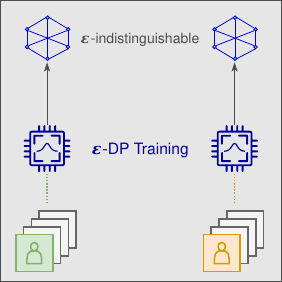}
\end{minipage}
    \hspace{ \stretch{1} }
    \begin{minipage}{.35\textwidth}
    \includegraphics{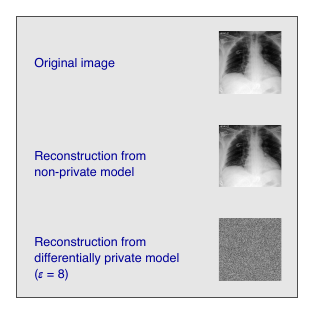}
\end{minipage}
    \hspace{ \stretch{1} }\\[1em]
    \includegraphics{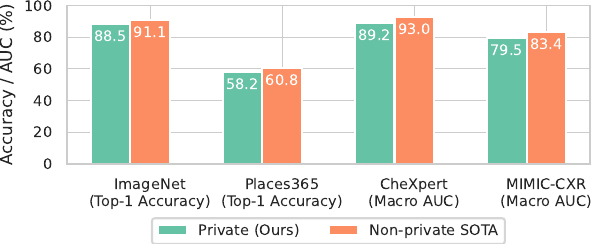}\\[1em]
    \hspace{ \stretch{1} }
    \includegraphics{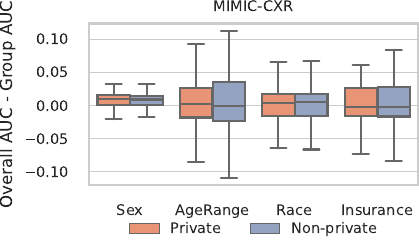}
    \hspace{ \stretch{1} }
    \includegraphics{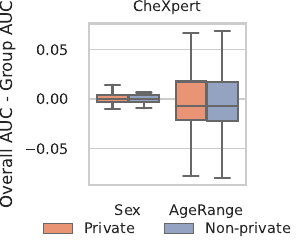}
    \hspace{ \stretch{1} }
    \caption{\small
    Overview of differential privacy and our results. (Top/Left) Differential privacy protects individuals in the training dataset by enforcing training outputs to be indistinguishable (up to a maximum privacy loss parameter $\varepsilon$) on pairs of datasets differing in a single individual. (Top/Right) Effect of the DP protection on reconstruction attacks against a classification model trained with and without DP on medical chest X-ray images. (Middle) We demonstrate it is possible to obtain highly accurate models with DP (at $\varepsilon$=8) that closely match the accuracy of the best models trained without privacy constraint. Results evaluated on the standard test set for each task. (Bottom) Our most accurate private models for chest X-ray classification exhibit AUC disparities (i.e. differences between population and group AUC) across demographic attributes, such as sex and age range, comparable to those of non-private classifiers. Distributions over subgroups and training randomness (20 seeds), metrics evaluated on internal test set split. Details in Appendices.
    }
    \label{fig:main-dp-high-level}
\end{figure}

Unfortunately for deep learning, achieving strong privacy protections with DP is harder when models are larger and when the data dimension is higher: more information needs to be hidden and there are more parameters which might reveal this information. As a consequence, existing practical deployments of DP learning are currently restricted to relatively small models and simple tasks. The large gap between the accuracy attainable with private and non-private learning is one of two main obstacles to unlock the routine use of DP for training models on sensitive data. The second obstacle is the observation that DP training can exacerbate disparities in model accuracy across subpopulations. Challenges in data underrepresentation and quality are known to induce unjustified disparities in accuracy across certain subpopulations in non-private models, and researchers have expressed concerns that DP training can further exacerbate such disparities \cite{bagdasaryan2019differential}. The disparities observed in models trained with DP highlight a potential, undesirable trade-off between privacy and fairness; especially in applications like healthcare where models inform consequential decisions and require access to private training data.

Our work takes a major step towards resolving these two problems in the context of image classification, a task where deep neural networks are ubiquitously used and often handle sensitive data (e.g. personal pictures, scanned documents or medical diagnostic images). To achieve this, we follow the dominant paradigm in the deep learning community consisting of fine-tuning networks pre-trained on large general purpose datasets \cite{bommasani2022opportunities}. Our first contribution is a reliable and accurate method for DP fine-tuning of large vision models pre-trained on non-sensitive data, which we evaluate on four challenging image classification benchmarks. We demonstrate that our method achieves substantially better accuracy than previously thought possible, often reaching the practical performance of previously deployed (non-private) models. On all four tasks, we achieve private accuracies within a few percent of the non-private state of the art (\Cref{fig:main-dp-high-level}). Our evaluation includes two medical imaging classification benchmarks where we are not only able to substantially reduce the gap between the best private and non-private models, but also illustrate our second contribution: demonstrating that our highly-accurate private models exhibit disparities across subpopulations which are no larger than those we observe in non-private models with comparable accuracy. Moreover, our results on medical imaging also highlight the remarkable effectiveness of privately fine-tuning foundation models pre-trained on non-sensitive data from a distribution significantly different to that of the private data. This observation is critical, since there is often very little public data from a similar distribution to private data in practical applications \cite{tramer2022considerations}.

Demonstrating the effectiveness of DP training techniques in image classification has far-reaching implications to ensure the benefits provided by models trained on sensitive data can be leveraged without compromising the privacy of the training data. Importantly, our approach aligns closely with standard practices used in industrial deep learning applications, including the use of models, algorithms and pre-training protocols which follow the same paradigms as standard deep learning frameworks \cite{deepmind2020jax,DBLP:conf/nips/PaszkeGMLBCKLGA19}. This ensures that our methods will remain relevant by being able to incorporate future advances in deep learning, including improved network architectures or pre-trained models. Altogether, this represents a significant milestone towards a new paradigm where deep learning practitioners can routinely leverage the formal privacy guarantees offered by DP to protect sensitive data when training machine learning models.

\section{Challenges of learning under differential privacy}

\begin{figure}[htpb]
    \centering
    \includegraphics[width=0.95\textwidth]{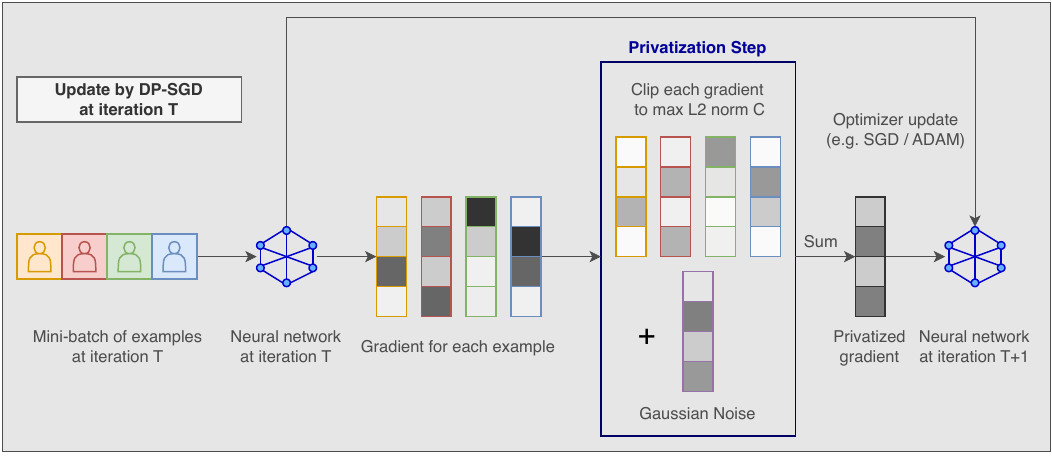}\\[2em]
    \includegraphics{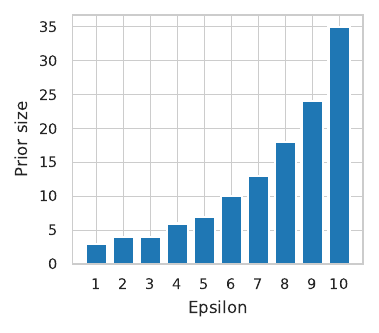}
    \caption{\small
    The DP-SGD algorithm and how to interpret its privacy guarantees. (Top) Differentially private SGD updates model parameters using a similar procedure to standard SGD training, with the difference that gradients are privatized through per-example clipping and the addition of isotropic Gaussian noise. These privatized gradients can also be used with other standard optimizers such as Adam. After training for T iterations with noise $\sigma$ and batch-size B on a dataset with N individuals, DP-SGD provides a DP guarantee with $\varepsilon \approx B \sqrt{T} / \sigma N$. (Bottom) We can interpret the privacy loss $\varepsilon$ of DP-SGD as the amount of side knowledge required by an adversary whose goal is to successfully identify which individual from a number of (equally likely) choices was part of the training data. More side knowledge means the adversary managed to narrow down the choice to a smaller set of candidate individuals - we plot the minimal number of choices that provably prevent successful identification (with probability $> 50\%$) as a function of $\varepsilon$. For example, at $\varepsilon$=8 a target individual in a dataset is protected as long as the adversary is unable to narrow down the individual's identity to a group with fewer than 18 individuals.
    }
    \label{fig:main-dp-detail}
\end{figure}

The strength of the DP guarantee is controlled through a privacy parameter $\varepsilon$; the smaller the value of $\varepsilon$, the smaller the risk that information about a training example will be revealed or that it is even memorized.\footnote{The term “memorization” is used throughout as a shorthand convenience to describe a broad range of phenomena whereby the weights of a model capture enough information to enable inferences about some individual training data points. There is active discussion within the technical and legal communities about whether the presence of this type of “memorization” suggests that neural networks “contain” their training data.} This parameter controls the worst-case privacy loss experienced by any individual in the training dataset; typical values used in practice are in the range $1 \leq \varepsilon \leq 10$ \cite{damienblog}. A DP guarantee can only be obtained if the training algorithm is stochastic, and the formal definition of the guarantee involves the log-likelihood ratio of any potential output model over pairs of datasets differing in a single individual, but more intuitive interpretations bounding the success of worst-case privacy attacks can also be obtained. We provide one such interpretation in \Cref{fig:main-dp-detail}, where we interpret the guarantee in terms of the strength of the side knowledge the adversary must obtain from other sources before being able to identify a training example (a detailed discussion of the privacy guarantee provided by DP is offered in the Appendices).

The most popular DP training technique in deep learning is differentially private stochastic gradient descent (DP-SGD) \cite{abadi2016deep}, an iterative gradient-based method where gradients used to update the model parameters are privatized by 
obfuscating the contribution of each individual example in the mini-batch through clipping and noise addition (see \Cref{fig:main-dp-detail}). The scale of the noise controls how much the contribution of each example to the mean gradient is obfuscated, and the strength of the privacy guarantee $\varepsilon$ depends on this noise scale, the batch-size, the number of training samples and the number of training iterations. Since DP-SGD is almost a drop-in replacement for SGD, it is currently the best candidate to train a wide range of machine learning models with meaningful privacy guarantees with minimal modifications to existing pipelines. However, it presents two major difficulties that have so far precluded its widespread adoption by practitioners.

The first difficulty is to attain high accuracy with models trained with differential privacy. Achieving the DP guarantee requires the injection of carefully crafted random noise into the training algorithm, and the maximum number of training iterations needs to be constrained. Training with noisy gradients in combination with a limited number of iterations makes optimization very challenging \cite{DBLP:conf/iclr/TramerB21}. This noisy regime also affects the adequacy of hyper-parameter choices and other best practices that have been carefully selected by deep learning practitioners to train accurate models without privacy constraints \cite{anil2021large}. Standard hyper-parameter choices need to be re-thought to be able to obtain high accuracy under the challenging conditions of DP training. Furthermore, since the noise is added independently to each coordinate of the gradient, its Euclidean norm grows with the number of parameters in the model. As a result, privacy researchers believe that DP-SGD will perform increasingly poorly as the model size increases \cite{DBLP:conf/iclr/TramerB21,yu2021large}.  This is a major challenge, since highly over-parameterized deep neural networks are currently dominant in the artificial intelligence community, achieving excellent performance across a wide range of tasks and data domains.

The second major difficulty is overcoming fairness issues associated with deep neural networks, one type of which is manifested through unjustified accuracy disparities across subpopulations. Even without differential privacy, when automatically detecting diseases in chest X-rays, underserved patient groups based on race, age, sex, or insurance type can experience underdiagnosis by machine learning models \cite{Seyyed-Kalantari21}. In the context of private models, prior works have suggested that differentially privacy mechanisms can incur a significant additional fairness penalty by increasing the accuracy disparity between different subgroups \cite{bagdasaryan2019differential,Santos-Lozada2020-go}. While multiple hypotheses have been proposed to explain the precise cause of these disparities, unbalanced and complex subgroup data have been commonly highlighted as obstacles to achieving both privacy and fairness \cite{farrand2020neither,sanyal2022unfair}. This phenomenon is perhaps not surprising since DP limits the amount of information that can be extracted from individual data points while still enabling learning across larger populations; strong differential privacy guarantees may therefore be in tension with the ability to accurately learn from small subgroups \cite{cummings2019compatibility}. In real world medical datasets where minority groups are often underrepresented and intersectional groups can be arbitrarily small, disparities that arise in non-private models may be exacerbated when using differentially private training \cite{suriyakumar2021chasing}. 

\section{Training highly accurate models with differential privacy}

Our first goal is to demonstrate that DP training can achieve practical levels of accuracy, i.e., accuracies close to the state-of-the-art achieved by non-private training. We identify four main elements that improve the accuracy of models trained with DP. In order of importance, these are (1) ensuring good signal propagation in models without batch-normalization \cite{DBLP:conf/icml/IoffeS15} (a popular component of vision models which is incompatible with DP-SGD), (2) using significantly larger batch sizes than standard in non-private training (as corroborated in other contexts by \cite{Li2021,anil2021large}), (3) carefully tuning the noise multiplier hyperparameter in DP-SGD (but not the clipping norm), and (4) improving model convergence with parameter averaging (see details in Appendices).

We first demonstrate that our approach is able to learn highly accurate image classifiers on medical images with strong privacy guarantees (a domain where privacy concerns naturally arise). This landmark result shows the ability of DP training to mitigate privacy concerns while offering high levels of accuracy in a practical and challenging scenario. To that end, we pre-train an NFNet-F0 model (72M parameters) \cite{DBLP:conf/icml/BrockDSS21} on ImageNet-21K, a public superset of the popular ImageNet dataset containing 14 million images \cite{DBLP:conf/nips/RidnikBNZ21}.
The NFNet model family is ideally suited to private training, since it was specifically designed to achieve high accuracies without batch-normalization. We fine-tune the model with DP-SGD on CheXpert \cite{IrvinRKYCCMHBSS19} and MIMIC-CXR \cite{Johnson2019-yh}. CheXpert is a public classification benchmark containing 224k chest X-rays, labeled with up to 14 clinically relevant observations. MIMIC-CXR is a similar dataset with 377k chest X-rays with the same label categories as CheXpert.

\begin{figure}[htpb]
    \centering
    \includegraphics{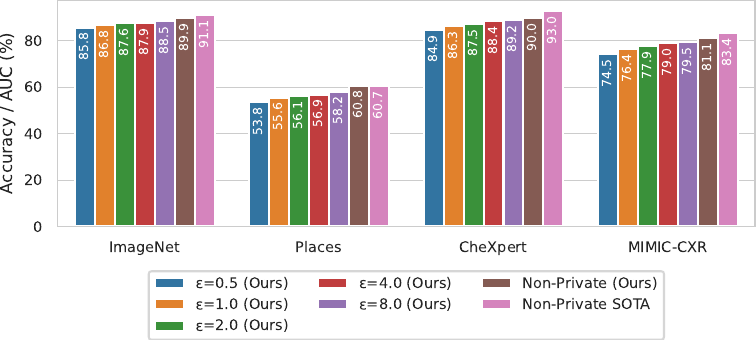}
    \caption{\small
    Performance of our approach across benchmarks and privacy levels. We achieve practical levels of performance, almost matching our non-private baselines, across several challenging image classification benchmarks and values for the privacy budget $\varepsilon$.
    }
    \label{fig:main-accuracy}
\end{figure}

As shown in \Cref{fig:main-accuracy}, our fine-tuned image classifiers are able to reach high accuracies. Specifically, on CheXpert, our model fine-tuned with differential privacy at $\varepsilon$=8 obtains 89.24\% AUC, to be compared to 89.97\% for the pre-trained model fine-tuned without privacy, and to 93.0\% for the state-of-the-art, achieved by ensembles of models trained without any privacy constraints \cite{DBLP:conf/iccv/Yuan0SY21}. Similar results also hold for MIMIC-CXR where our private model obtains 79.53\% AUC at $\varepsilon$=8, 81.14\% without privacy, and for which the published state-of-the-art obtains 84.04\% \cite{KamalZNH22} by using a more complicated architecture that segments the images as part of its processing. Remarkably, when imposing stringent privacy guarantees, e.g. a privacy budget of only $\varepsilon$=1, it is still possible to obtain useful levels of accuracy, with an AUC of 86.34\% on CheXpert and 76.40\% on MIMIC-CXR. 
These results demonstrate that it is possible to achieve practical levels of accuracy with strong differentially private guarantees on large-scale medical image classification tasks that are significantly more challenging than previous work \cite{ziller2021medical,sanyal2022unfair}.

To demonstrate that these results are not confined to the medical image domain, we further consider two popular academic benchmarks: ImageNet and Places-365. ImageNet is a dataset of 1.3 million images from 1000 classes, corresponding to simple objects like ``car'', ``dolphin'' or ``ocean liner'' \cite{ILSVRC15}. It is widely recognized as one of the most important benchmarks in computer vision, providing extremely strong baselines for image classifiers trained without differential privacy against which we can compare our private models. Places-365 is a large-scale dataset for the recognition of 365 different scenes \cite{DBLP:journals/pami/ZhouLKO018}, which requires the model to identify diverse locations such as a science museum or a martial arts gym. We again use the NFNet family of deep convolutional networks \cite{DBLP:conf/icml/BrockDSS21}. We pre-train two versions of our models: one using ImageNet-21K as above, and another using JFT \cite{sun2017revisiting}, a proprietary labeled dataset comprising 4 billion images collected from public internet pages. We then fine-tune these models with DP-SGD on Places-365 and ImageNet, obtaining formal privacy guarantees on those downstream datasets.

When using the NFNet-F7+ model (947M parameters) pre-trained on JFT, we achieve a top-1 accuracy of 88.5\% on ImageNet under a DP guarantee of $\varepsilon$=8, which is only 1.4\% lower than the accuracy of this same pre-trained model when fine-tuned without differential privacy, and just 2.6\% below the overall ImageNet state-of-the-art \cite{chen2023symbolic}.
This also slightly exceeds the previously best existing results with DP \cite{mehta2022large},
which reach 88\% at $\varepsilon$=8.
We also achieve high accuracy under stricter privacy guarantees (lower $\varepsilon$). For example, at $\varepsilon$ = 1, we achieve 86.8\% top-1 accuracy. To put this in perspective, this is significantly higher than the 77\% accuracy of the ResNet-50 architecture trained without any privacy guarantees \cite{he2016deep}, a popular model that has been widely used in computer vision systems. This is also significantly better than fine-tuning the smaller NFNet-F3 (255M parameters) with DP, which achieves 87\% top-1 accuracy under $\varepsilon$=8, and thereby further demonstrates the value of using strong pre-trained foundation models.

We achieve similarly strong results on Places-365. Using the NFNet-F3 pre-trained on JFT, we achieve an accuracy of 58.2\% with $\varepsilon$=8.0; within 3\% of our non-private baseline of 60.8\%, for which we fine-tune the same pre-trained model without privacy. This non-private baseline slightly outperforms the non-private state of the art on this task (60.7\%).  We also achieve 56.6\% with $\varepsilon$=8.0 with the same model when pre-trained on the significantly smaller public dataset ImageNet-21K, demonstrating that strong private classifiers can be obtained without access to proprietary pre-training data.

Previous research only demonstrated that DP fine-tuning of image classifiers is effective on data that is very similar to the pre-training dataset \cite{Kurakin22,mehta2022large,arasteh2023preserving}. In contrast, we attain very high accuracies on two chest X-ray datasets and a scene recognition dataset despite pre-training only on natural images of simple objects. This demonstrates the applicability of DP fine-tuning to situations where it is impossible to find public or non-sensitive pre-training data closely related to the private task.

\section{Accurate private training need not exacerbate disparities between subgroups}

\begin{figure}[htpb]
    \centering
    \includegraphics{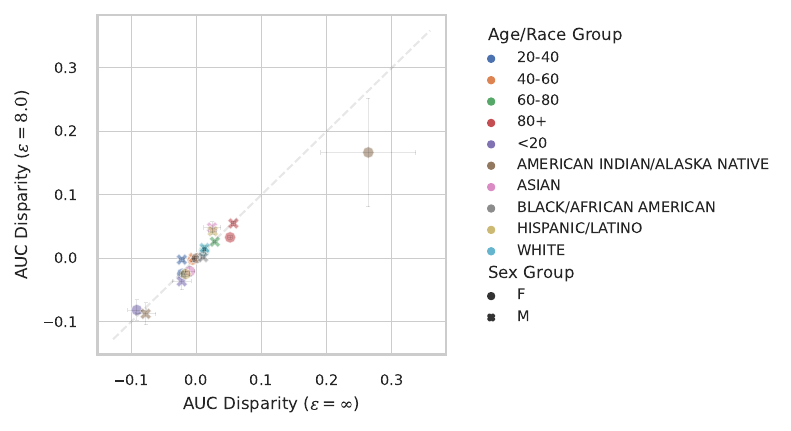}
    \includegraphics{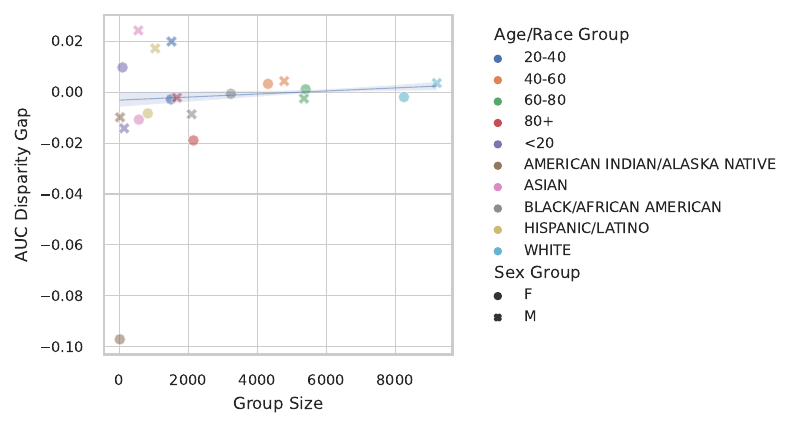}
    \caption{\small
    AUC disparities (i.e. population AUC - subgroup AUC) of private models trained on MIMIC-CXR are comparable to disparities observed on non-private models with comparable accuracy. (Top) For the private ($\varepsilon$=8) and non-private baseline models from \Cref{fig:main-accuracy}, comparing AUC disparities (averaged over 20 independent runs) by subgroup. Gray crosses represent standard deviation. (Bottom) Stratification of differences in disparities (averaged over 20 independent runs) by subgroup size. Blue line represents OLS regression predicting disparity gap as a function of group size (and 95\% confidence interval based on 1000-fold bootstrap). The slope of the regression model lies in the 95\% confidence interval $[5.7 \cdot 10^{-7}, 3.23 \cdot 10^{-6}]$.
    }
    \label{fig:main-analysis}
\end{figure}

Our second goal is to illustrate that private models do not necessarily exacerbate existing disparities across subgroups. Focusing on the setting of models for X-ray diagnostics, we examine demographic intersectional subgroups defined on age, sex, and race in the MIMIC-CXR dataset. We follow prior work in measuring AUC disparity as the difference in AUC between the overall population and the subgroup \cite{larrazabal2020gender, zhang2022improving}. In \Cref{fig:main-analysis}, we compare the disparities exhibited by our private models at $\varepsilon$=8 and the non-private baselines presented in the previous section. We observe that private and non-private models have a tendency to exhibit similar disparities across the considered subgroups, and that, overall, AUC disparities are not systematically worse for private than non-private models. Furthermore, while we observe that the differences in disparities between subgroups exhibit larger variation between the private and non-private models on smaller subgroups, the average difference in disparities is strongly concentrated around zero regardless of group size. We also observe a similar pattern of comparable disparity between private and non-private models in the CheXpert dataset \cite{IrvinRKYCCMHBSS19}, another commonly studied dataset for this task (details in Appendices).

In contrast to what prior works suggest, we observe that our accurate private models do not cause worse group fairness outcomes than non-private baselines. This result also extends to smaller values of $\varepsilon$ as well as intersectional subgroups. Our experiments also demonstrate that this finding is stable across both fine-tuned medical imaging models presented in the main text as well as models trained from scratch on smaller datasets used in prior works (see details in Appendices). These results motivate our hypothesis that the substantially increased disparities observed in private models in prior works are not intrinsic to private models, especially not to those with accuracy comparable to strong non-private models.

We believe that these results represent a significant advancement for differentially private training, since they open the door to providing strong privacy guarantees without further negative impacts on disparities between subgroups. We note, however, that fairness in medical decision-making remains a complex pursuit, since multiple different fairness metrics should be considered concurrently depending on how a system is deployed in practice \cite{Seyyed-Kalantari21}.

\section{Discussion}

In this work, we trained large-scale image classifiers with strong privacy guarantees, while achieving accuracies competitive with state-of-the-art non-private models and observing levels of disparity across subpopulations comparable to those of accurate non-private models. By demonstrating such results on widely used benchmarks and public medical imaging datasets, we provide compelling evidence that differentially private training is a practical tool for deploying accurate machine learning models while providing strong privacy guarantees for their training data. Importantly, we show that differentially private fine-tuning for image classification can leverage large-scale foundation models pre-trained on public datasets, which has proved to be an effective recipe for practical advances in machine learning. These pre-trained models significantly aid private training, even when the private data and the pre-training dataset come from very different distributions. We believe our methodology is relevant to a large number of scenarios where task-specific sensitive datasets can be used to fine-tune readily available pre-trained models.

From the perspective of practitioners looking to adopt our methodology, there are two important considerations to be aware of. First, the privacy guarantees only apply to the fine-tuning data: data used when pre-training the model is not covered by DP guarantees. This is not a major limitation in our view, since non-sensitive public datasets are widely available, and pre-training and fine-tuning data can come from different distributions. However, it implies that care needs to be taken when compiling pre-training datasets based on publicly available data. The second consideration is that our analysis on disparities focuses on pre-existing demographic subgroups from particular datasets. In general, a minimal subgroup size might be required to achieve similar disparities to non-private models, although such a threshold can also depend on the nature of the data and the similarities between individuals from different subpopulations. Our analysis suggests that accurate DP training will not necessarily exacerbate disparities, but we recommend that practitioners assess this issue on their particular application, taking into account the context and potential impact of such disparities.

We hope our results open the door for impactful and responsible deployments of machine learning systems with rigorous privacy guarantees. We believe that DP is ready for widespread adoption in machine learning production systems, and governmental applications. Our codebase (including models pre-trained on ImageNet-21K) has been open-sourced to enable reproducibility of our results \cite{github}, verify the correctness of our DP-SGD implementation, and to help practitioners adopt our techniques in their machine learning pipelines.

\subsubsection*{Acknowledgments}

We would like to thank: Ira Ktena, Stevie Bergman, Jessica Schrouff and Andrew Trask for detailed feedback that helped improve the present manuscript; Matthias Bauer and Sven Gowal for insightful comments that helped improve an earlier version of the paper; Zahra Ahmed and Kitty Stacpoole for project management support; Danielle Belgrave, Sahra Ghalebikesabi, Alan Karthikesalingam, Nenad Tomašev and Thomas Steinke for insightful discussions; Rudy Bunel for code reviews; John Aslanides for code quality reviews and open-sourcing support; Alison Reid and Jon Small for support during the open-sourcing process; Abhradeep Thakurta, Florian Tramèr and Harsh Mehta for discussions on related works; and Andrew Brock, Taylan Cemgil, Raia Hadsell, Koray Kavukcuoglu, Razvan Pascanu and Yee Whye Teh for advice throughout the project. We also want to thank the Stanford Center for Artificial Intelligence in Medicine and Imaging for providing access to CheXpert.

\bibliographystyle{plain}
\bibliography{biblio}

\begin{thebibliography}{100}

\bibitem{abadi2016deep}
Martin Abadi, Andy Chu, Ian Goodfellow, H~Brendan McMahan, Ilya Mironov, Kunal
  Talwar, and Li~Zhang.
\newblock Deep learning with differential privacy.
\newblock In {\em Proceedings of the 2016 {ACM} {SIGSAC} Conference on Computer
  and Communications Security}. {ACM}, oct 2016.

\bibitem{anil2021large}
Rohan Anil, Badih Ghazi, Vineet Gupta, Ravi Kumar, and Pasin Manurangsi.
\newblock Large-scale differentially private {BERT}.
\newblock Preprint arXiv:2108.01624 [cs.LG], 2021.

\bibitem{arasteh2023preserving}
Soroosh~Tayebi Arasteh, Mahshad Lotfinia, Teresa Nolte, Marwin Saehn, Peter
  Isfort, Christiane Kuhl, Sven Nebelung, Georgios Kaissis, and Daniel Truhn.
\newblock Preserving privacy in domain transfer of medical ai models comes at
  no performance costs: The integral role of differential privacy.
\newblock Preprint arXiv:2306.06503 [cs.LG], 2023.

\bibitem{arasteh2023private}
Soroosh~Tayebi Arasteh, Alexander Ziller, Christiane Kuhl, Marcus Makowski,
  Sven Nebelung, Rickmer Braren, Daniel Rueckert, Daniel Truhn, and Georgios
  Kaissis.
\newblock Private, fair and accurate: Training large-scale, privacy-preserving
  ai models in medical imaging.
\newblock Preprint arXiv:2302.01622 [eess.IV], 2023.

\bibitem{deepmind2020jax}
Igor Babuschkin, Kate Baumli, Alison Bell, Surya Bhupatiraju, Jake Bruce, Peter
  Buchlovsky, David Budden, Trevor Cai, Aidan Clark, Ivo Danihelka, Claudio
  Fantacci, Jonathan Godwin, Chris Jones, Tom Hennigan, Matteo Hessel, Steven
  Kapturowski, Thomas Keck, Iurii Kemaev, Michael King, Lena Martens, Vladimir
  Mikulik, Tamara Norman, John Quan, George Papamakarios, Roman Ring, Francisco
  Ruiz, Alvaro Sanchez, Rosalia Schneider, Eren Sezener, Stephen Spencer,
  Srivatsan Srinivasan, Wojciech Stokowiec, and Fabio Viola, The {D}eep{M}ind
  {JAX} {E}cosystem, GitHub, 2020; \url{http://github.com/deepmind}.

\bibitem{bagdasaryan2019differential}
Eugene Bagdasaryan, Omid Poursaeed, and Vitaly Shmatikov.
\newblock Differential privacy has disparate impact on model accuracy.
\newblock In Hanna~M. Wallach, Hugo Larochelle, Alina Beygelzimer, Florence
  d'Alch{\'{e}}{-}Buc, Emily~B. Fox, and Roman Garnett, editors, {\em Advances
  in Neural Information Processing Systems 32: Annual Conference on Neural
  Information Processing Systems 2019, NeurIPS 2019, December 8-14, 2019,
  Vancouver, BC, Canada}, pages 15453--15462, 2019.

\bibitem{github}
Borja Balle, Leonard Berrada, Soham De, Jamie Hayes, Samuel~L Smith, and Robert
  Stanforth, {JAX}-{P}rivacy: Algorithms for privacy-preserving machine
  learning in jax, 0.1.0, 2022; \url{http://github.com/deepmind/jax_privacy}.

\bibitem{balle2022reconstructing}
Borja Balle, Giovanni Cherubin, and Jamie Hayes.
\newblock Reconstructing training data with informed adversaries.
\newblock In {\em Symposium on Security and Privacy ({SP})}. {IEEE}, may 2022.

\bibitem{barocas-hardt-narayanan}
Solon Barocas, Moritz Hardt, and Arvind Narayanan.
\newblock {\em Fairness and Machine Learning}.
\newblock fairmlbook.org, 2019.

\bibitem{DBLP:conf/focs/BassilyST14}
Raef Bassily, Adam~D. Smith, and Abhradeep Thakurta.
\newblock Private empirical risk minimization: Efficient algorithms and tight
  error bounds.
\newblock In {\em 55th Annual Symposium on Foundations of Computer Science}.
  {IEEE}, oct 2014.

\bibitem{benz2021robustness}
Philipp Benz, Chaoning Zhang, Adil Karjauv, and In~So Kweon.
\newblock Robustness may be at odds with fairness: An empirical study on
  class-wise accuracy.
\newblock In Luca Bertinetto, Jo{\~{a}}o~F. Henriques, Samuel Albanie, Michela
  Paganini, and G{\"{u}}l Varol, editors, {\em NeurIPS 2020 Workshop on
  Pre-registration in Machine Learning, 11 December 2020, Virtual Event},
  volume 148 of {\em Proceedings of Machine Learning Research}, pages 325--342.
  {PMLR}, 2020.

\bibitem{bommasani2022opportunities}
Rishi Bommasani, Drew~A. Hudson, Ehsan Adeli, Russ Altman, Simran Arora, Sydney
  von Arx, Michael~S. Bernstein, Jeannette Bohg, Antoine Bosselut, Emma
  Brunskill, Erik Brynjolfsson, Shyamal Buch, Dallas Card, Rodrigo Castellon,
  Niladri Chatterji, Annie Chen, Kathleen Creel, Jared~Quincy Davis, Dora
  Demszky, Chris Donahue, Moussa Doumbouya, Esin Durmus, Stefano Ermon, John
  Etchemendy, Kawin Ethayarajh, Li~Fei-Fei, Chelsea Finn, Trevor Gale, Lauren
  Gillespie, Karan Goel, Noah Goodman, Shelby Grossman, Neel Guha, Tatsunori
  Hashimoto, Peter Henderson, John Hewitt, Daniel~E. Ho, Jenny Hong, Kyle Hsu,
  Jing Huang, Thomas Icard, Saahil Jain, Dan Jurafsky, Pratyusha Kalluri,
  Siddharth Karamcheti, Geoff Keeling, Fereshte Khani, Omar Khattab, Pang~Wei
  Koh, Mark Krass, Ranjay Krishna, Rohith Kuditipudi, Ananya Kumar, Faisal
  Ladhak, Mina Lee, Tony Lee, Jure Leskovec, Isabelle Levent, Xiang~Lisa Li,
  Xuechen Li, Tengyu Ma, Ali Malik, Christopher~D. Manning, Suvir Mirchandani,
  Eric Mitchell, Zanele Munyikwa, Suraj Nair, Avanika Narayan, Deepak
  Narayanan, Ben Newman, Allen Nie, Juan~Carlos Niebles, Hamed Nilforoshan,
  Julian Nyarko, Giray Ogut, Laurel Orr, Isabel Papadimitriou, Joon~Sung Park,
  Chris Piech, Eva Portelance, Christopher Potts, Aditi Raghunathan, Rob Reich,
  Hongyu Ren, Frieda Rong, Yusuf Roohani, Camilo Ruiz, Jack Ryan, Christopher
  Ré, Dorsa Sadigh, Shiori Sagawa, Keshav Santhanam, Andy Shih, Krishnan
  Srinivasan, Alex Tamkin, Rohan Taori, Armin~W. Thomas, Florian Tramèr,
  Rose~E. Wang, William Wang, Bohan Wu, Jiajun Wu, Yuhuai Wu, Sang~Michael Xie,
  Michihiro Yasunaga, Jiaxuan You, Matei Zaharia, Michael Zhang, Tianyi Zhang,
  Xikun Zhang, Yuhui Zhang, Lucia Zheng, Kaitlyn Zhou, and Percy Liang.
\newblock On the opportunities and risks of foundation models.
\newblock Preprint arXiv:2108.07258 [cs.LG], 2022.

\bibitem{jax2018github}
James Bradbury, Roy Frostig, Peter Hawkins, Matthew~James Johnson, Chris Leary,
  Dougal Maclaurin, George Necula, Adam Paszke, Jake Vander{P}las, Skye
  Wanderman-{M}ilne, and Qiao Zhang, {JAX}: composable transformations of
  {P}ython+{N}um{P}y programs, GitHub, 2018;
  \url{http://github.com/google/jax}.

\bibitem{DBLP:conf/icml/BrockDSS21}
Andy Brock, Soham De, Samuel~L. Smith, and Karen Simonyan.
\newblock High-performance large-scale image recognition without normalization.
\newblock In Marina Meila and Tong Zhang, editors, {\em Proceedings of the 38th
  International Conference on Machine Learning, {ICML} 2021, 18-24 July 2021,
  Virtual Event}, volume 139 of {\em Proceedings of Machine Learning Research},
  pages 1059--1071. {PMLR}, 2021.

\bibitem{DBLP:conf/nips/BrownMRSKDNSSAA20}
Tom~B. Brown, Benjamin Mann, Nick Ryder, Melanie Subbiah, Jared Kaplan,
  Prafulla Dhariwal, Arvind Neelakantan, Pranav Shyam, Girish Sastry, Amanda
  Askell, Sandhini Agarwal, Ariel Herbert{-}Voss, Gretchen Krueger, Tom
  Henighan, Rewon Child, Aditya Ramesh, Daniel~M. Ziegler, Jeffrey Wu, Clemens
  Winter, Christopher Hesse, Mark Chen, Eric Sigler, Mateusz Litwin, Scott
  Gray, Benjamin Chess, Jack Clark, Christopher Berner, Sam McCandlish, Alec
  Radford, Ilya Sutskever, and Dario Amodei.
\newblock Language models are few-shot learners.
\newblock In Hugo Larochelle, Marc'Aurelio Ranzato, Raia Hadsell,
  Maria{-}Florina Balcan, and Hsuan{-}Tien Lin, editors, {\em Advances in
  Neural Information Processing Systems 33: Annual Conference on Neural
  Information Processing Systems 2020, NeurIPS 2020, December 6-12, 2020,
  virtual}, 2020.

\bibitem{uscensusreport}
U.S.~Census Bureau.
\newblock Disclosure avoidance for the 2020 census: An introduction.
\newblock Technical report, U.S. Government Publishing Office, 2021;
  \url{https://www2.census.gov/library/publications/decennial/2020/2020-census-disclosure-avoidance-handbook.pdf}.

\bibitem{DBLP:conf/sp/CarliniCN0TT22}
Nicholas Carlini, Steve Chien, Milad Nasr, Shuang Song, Andreas Terzis, and
  Florian Tram{\`{e}}r.
\newblock Membership inference attacks from first principles.
\newblock In {\em 43rd {IEEE} Symposium on Security and Privacy, {SP} 2022, San
  Francisco, CA, USA, May 22-26, 2022}, pages 1897--1914. {IEEE}, 2022.

\bibitem{carlini2019distribution}
Nicholas Carlini, Ulfar Erlingsson, and Nicolas Papernot.
\newblock Distribution density, tails, and outliers in machine learning:
  Metrics and applications.
\newblock Preprint arXiv:1910.13427 [cs.LG], 2019.

\bibitem{DBLP:journals/corr/abs-2301-13188}
Nicholas Carlini, Jamie Hayes, Milad Nasr, Matthew Jagielski, Vikash Sehwag,
  Florian Tramèr, Borja Balle, Daphne Ippolito, and Eric Wallace.
\newblock Extracting training data from diffusion models.
\newblock Preprint arXiv:2301.13188 [cs.CR], 2023.

\bibitem{DBLP:conf/uss/CarliniTWJHLRBS21}
Nicholas Carlini, Florian Tram{\`{e}}r, Eric Wallace, Matthew Jagielski, Ariel
  Herbert{-}Voss, Katherine Lee, Adam Roberts, Tom~B. Brown, Dawn Song,
  {\'{U}}lfar Erlingsson, Alina Oprea, and Colin Raffel.
\newblock Extracting training data from large language models.
\newblock In Michael Bailey and Rachel Greenstadt, editors, {\em 30th {USENIX}
  Security Symposium, {USENIX} Security 2021, August 11-13, 2021}, pages
  2633--2650. {USENIX} Association, 2021.

\bibitem{cattan2022fine}
Yannis Cattan, Christopher~A Choquette-Choo, Nicolas Papernot, and Abhradeep
  Thakurta.
\newblock Fine-tuning with differential privacy necessitates an additional
  hyperparameter search.
\newblock Preprint arXiv:2210.02156 [cs.LG], 2023.

\bibitem{chang2021privacy}
Hongyan Chang and Reza Shokri.
\newblock On the privacy risks of algorithmic fairness.
\newblock In {\em European Symposium on Security and Privacy (EuroS\&P)}, pages
  292--303. IEEE, {IEEE}, sep 2021.

\bibitem{chen2023symbolic}
Xiangning Chen, Chen Liang, Da~Huang, Esteban Real, Kaiyuan Wang, Yao Liu, Hieu
  Pham, Xuanyi Dong, Thang Luong, Cho-Jui Hsieh, Yifeng Lu, and Quoc~V. Le.
\newblock Symbolic discovery of optimization algorithms.
\newblock Preprint arXiv:2302.06675 [cs.LG], 2023.

\bibitem{cummings2019compatibility}
Rachel Cummings, Varun Gupta, Dhamma Kimpara, and Jamie Morgenstern.
\newblock On the compatibility of privacy and fairness.
\newblock In {\em Adjunct Publication of the 27th Conference on User Modeling,
  Adaptation and Personalization}, pages 309--315. {ACM}, jun 2019.

\bibitem{de2022unlocking}
Soham De, Leonard Berrada, Jamie Hayes, Samuel~L Smith, and Borja Balle.
\newblock Unlocking high-accuracy differentially private image classification
  through scale.
\newblock Preprint arXiv:2204.13650 [cs.LG], 2022.

\bibitem{dehghani2023scaling}
Mostafa Dehghani, Josip Djolonga, Basil Mustafa, Piotr Padlewski, Jonathan
  Heek, Justin Gilmer, Andreas Steiner, Mathilde Caron, Robert Geirhos, Ibrahim
  Alabdulmohsin, Rodolphe Jenatton, Lucas Beyer, Michael Tschannen, Anurag
  Arnab, Xiao Wang, Carlos Riquelme, Matthias Minderer, Joan Puigcerver, Utku
  Evci, Manoj Kumar, Sjoerd van Steenkiste, Gamaleldin~F. Elsayed, Aravindh
  Mahendran, Fisher Yu, Avital Oliver, Fantine Huot, Jasmijn Bastings,
  Mark~Patrick Collier, Alexey Gritsenko, Vighnesh Birodkar, Cristina
  Vasconcelos, Yi~Tay, Thomas Mensink, Alexander Kolesnikov, Filip Pavetić,
  Dustin Tran, Thomas Kipf, Mario Lučić, Xiaohua Zhai, Daniel Keysers,
  Jeremiah Harmsen, and Neil Houlsby.
\newblock Scaling vision transformers to 22 billion parameters.
\newblock Preprint arXiv:2302.05442 [cs.CV], 2023.

\bibitem{DengDSLL009}
Jia Deng, Wei Dong, Richard Socher, Li{-}Jia Li, Kai Li, and Li~Fei{-}Fei.
\newblock Imagenet: {A} large-scale hierarchical image database.
\newblock In {\em Conference on Computer Vision and Pattern Recognition}.
  {IEEE}, jun 2009.

\bibitem{damienblog}
Damien Desfontaines.
\newblock A list of real-world uses of differential privacy.
\newblock Technical report, Personal Blog, 2021;
  \url{https://desfontain.es/privacy/real-world-differential-privacy.html}.

\bibitem{devlin2018bert}
Jacob Devlin, Ming{-}Wei Chang, Kenton Lee, and Kristina Toutanova.
\newblock {BERT:} pre-training of deep bidirectional transformers for language
  understanding.
\newblock In Jill Burstein, Christy Doran, and Thamar Solorio, editors, {\em
  Proceedings of the 2019 Conference of the North American Chapter of the
  Association for Computational Linguistics: Human Language Technologies,
  {NAACL-HLT} 2019, Minneapolis, MN, USA, June 2-7, 2019, Volume 1 (Long and
  Short Papers)}, pages 4171--4186. Association for Computational Linguistics,
  2019.

\bibitem{dong2022gaussian}
Jinshuo Dong, Aaron Roth, and Weijie~J Su.
\newblock Gaussian differential privacy.
\newblock {\em Journal of the Royal Statistical Society Series B: Statistical
  Methodology}, 84(1):3--37, 2022.

\bibitem{DormannFAP21}
Friedrich D{\"{o}}rmann, Osvald Frisk, Lars~N{\o}rvang Andersen, and
  Christian~Fischer Pedersen.
\newblock Not all noise is accounted equally: How differentially private
  learning benefits from large sampling rates.
\newblock In {\em 31st International Workshop on Machine Learning for Signal
  Processing ({MLSP})}. {IEEE}, oct 2021.

\bibitem{DBLP:journals/popets/DoroshenkoGKKM22}
Vadym Doroshenko, Badih Ghazi, Pritish Kamath, Ravi Kumar, and Pasin
  Manurangsi.
\newblock Connect the dots: Tighter discrete approximations of privacy loss
  distributions.
\newblock In {\em Proceedings on Privacy Enhancing Technologies}, volume 2022,
  pages 552--570. Privacy Enhancing Technologies Symposium Advisory Board, oct
  2022.

\bibitem{dwork2006calibrating}
Cynthia Dwork, Frank McSherry, Kobbi Nissim, and Adam Smith.
\newblock Calibrating noise to sensitivity in private data analysis.
\newblock In {\em Theory of cryptography conference}, volume~7, pages 17--51.
  Journal of Privacy and Confidentiality, may 2006.

\bibitem{TCS-042}
Cynthia Dwork and Aaron Roth.
\newblock The algorithmic foundations of differential privacy.
\newblock {\em Foundations and Trends{\textregistered} in Theoretical Computer
  Science}, 9(3-4):211--407, 2014.

\bibitem{farrand2020neither}
Tom Farrand, Fatemehsadat Mireshghallah, Sahib Singh, and Andrew Trask.
\newblock Neither private nor fair: Impact of data imbalance on utility and
  fairness in differential privacy.
\newblock In {\em Proceedings of the 2020 Workshop on Privacy-Preserving
  Machine Learning in Practice}, pages 15--19. {ACM}, nov 2020.

\bibitem{DBLP:conf/stoc/Feldman20}
Vitaly Feldman.
\newblock Does learning require memorization? a short tale about a long tail.
\newblock In Konstantin Makarychev, Yury Makarychev, Madhur Tulsiani, Gautam
  Kamath, and Julia Chuzhoy, editors, {\em Proccedings of the 52nd Annual {ACM}
  {SIGACT} Symposium on Theory of Computing, {STOC} 2020, Chicago, IL, USA,
  June 22-26, 2020}, pages 954--959. {ACM}, 2020.

\bibitem{feldman2020neural}
Vitaly Feldman and Chiyuan Zhang.
\newblock What neural networks memorize and why: Discovering the long tail via
  influence estimation.
\newblock In Hugo Larochelle, Marc'Aurelio Ranzato, Raia Hadsell,
  Maria{-}Florina Balcan, and Hsuan{-}Tien Lin, editors, {\em Advances in
  Neural Information Processing Systems 33: Annual Conference on Neural
  Information Processing Systems 2020, NeurIPS 2020, December 6-12, 2020,
  virtual}, 2020.

\bibitem{Fort2021}
Stanislav Fort, Andrew Brock, Razvan Pascanu, Soham De, and Samuel~L. Smith.
\newblock Drawing multiple augmentation samples per image during training
  efficiently decreases test error.
\newblock Preprint arXiv:2105.13343 [cs.LG], 2021.

\bibitem{ganesh2023public}
Arun Ganesh, Mahdi Haghifam, Milad Nasr, Sewoong Oh, Thomas Steinke,
  Om~Thakkar, Abhradeep Thakurta, and Lun Wang.
\newblock Why is public pretraining necessary for private model training?
\newblock Preprint arXiv:2302.09483 [cs.LG], 2023.

\bibitem{NEURIPS2020_c4ede56b}
Jonas Geiping, Hartmut Bauermeister, Hannah Dr{\"{o}}ge, and Michael Moeller.
\newblock Inverting gradients - how easy is it to break privacy in federated
  learning?
\newblock In Hugo Larochelle, Marc'Aurelio Ranzato, Raia Hadsell,
  Maria{-}Florina Balcan, and Hsuan{-}Tien Lin, editors, {\em Advances in
  Neural Information Processing Systems 33: Annual Conference on Neural
  Information Processing Systems 2020, NeurIPS 2020, December 6-12, 2020,
  virtual}, 2020.

\bibitem{googledplibrary}
Google, Differential privacy accounting library, GitHub, 2023;
  \url{https://github.com/google/differential-privacy}.

\bibitem{DBLP:journals/corr/abs-2210-13662}
Chuan Guo, Alexandre Sablayrolles, and Maziar Sanjabi.
\newblock Analyzing privacy leakage in machine learning via multiple hypothesis
  testing: {A} lesson from fano.
\newblock Preprint arXiv:2210.13662 [cs.LG], 2022.

\bibitem{DBLP:journals/jmlr/HallRW13}
Rob Hall, Alessandro Rinaldo, and Larry~A. Wasserman.
\newblock Differential privacy for functions and functional data.
\newblock {\em Journal of Machine Learning Research}, 14:703--727, 2013.

\bibitem{hardt2016equality}
Moritz Hardt, Eric Price, and Nati Srebro.
\newblock Equality of opportunity in supervised learning.
\newblock In Daniel~D. Lee, Masashi Sugiyama, Ulrike von Luxburg, Isabelle
  Guyon, and Roman Garnett, editors, {\em Advances in Neural Information
  Processing Systems 29: Annual Conference on Neural Information Processing
  Systems 2016, December 5-10, 2016, Barcelona, Spain}, pages 3315--3323, 2016.

\bibitem{DBLP:journals/corr/abs-2302-07225}
Jamie Hayes, Saeed Mahloujifar, and Borja Balle.
\newblock Bounding training data reconstruction in {DP-SGD}.
\newblock Preprint arXiv:2302.07225 [cs.CR], 2023.

\bibitem{he2022exploring}
Jiyan He, Xuechen Li, Da~Yu, Huishuai Zhang, Janardhan Kulkarni, Yin~Tat Lee,
  Arturs Backurs, Nenghai Yu, and Jiang Bian.
\newblock Exploring the limits of differentially private deep learning with
  group-wise clipping.
\newblock Preprint arXiv:2212.01539 [cs.LG], 2022.

\bibitem{he2016deep}
Kaiming He, Xiangyu Zhang, Shaoqing Ren, and Jian Sun.
\newblock Deep residual learning for image recognition.
\newblock In {\em Conference on Computer Vision and Pattern Recognition
  ({CVPR})}. {IEEE}, jun 2016.

\bibitem{hoffer19}
Elad Hoffer, Tal Ben{-}Nun, Itay Hubara, Niv Giladi, Torsten Hoefler, and
  Daniel Soudry.
\newblock Augment your batch: Better training with larger batches.
\newblock Preprint arXiv:1901.09335 [cs.LG], 2019.

\bibitem{holzl2023equivariant}
Florian~A. H{\"{o}}lzl, Daniel Rueckert, and Georgios Kaissis.
\newblock Equivariant differentially private deep learning.
\newblock Preprint arXiv:2301.13104 [cs.CV], 2023.

\bibitem{huang2021evaluating}
Yangsibo Huang, Samyak Gupta, Zhao Song, Kai Li, and Sanjeev Arora.
\newblock Evaluating gradient inversion attacks and defenses in federated
  learning.
\newblock In Marc'Aurelio Ranzato, Alina Beygelzimer, Yann~N. Dauphin, Percy
  Liang, and Jennifer~Wortman Vaughan, editors, {\em Advances in Neural
  Information Processing Systems 34: Annual Conference on Neural Information
  Processing Systems 2021, NeurIPS 2021, December 6-14, 2021, virtual}, pages
  7232--7241, 2021.

\bibitem{humphries2020differentially}
Thomas Humphries, Matthew Rafuse, Lindsey Tulloch, Simon Oya, Ian Goldberg, Urs
  Hengartner, and Florian Kerschbaum.
\newblock Differentially private learning does not bound membership inference.
\newblock Preprint arXiv:2010.12112 [cs.CR], 2020.

\bibitem{DBLP:conf/icml/IoffeS15}
Sergey Ioffe and Christian Szegedy.
\newblock Batch normalization: Accelerating deep network training by reducing
  internal covariate shift.
\newblock In Francis~R. Bach and David~M. Blei, editors, {\em Proceedings of
  the 32nd International Conference on Machine Learning, {ICML} 2015, Lille,
  France, 6-11 July 2015}, volume~37 of {\em {JMLR} Workshop and Conference
  Proceedings}, pages 448--456. JMLR.org, 2015.

\bibitem{IrvinRKYCCMHBSS19}
Jeremy Irvin, Pranav Rajpurkar, Michael Ko, Yifan Yu, Silviana Ciurea{-}Ilcus,
  Chris Chute, Henrik Marklund, Behzad Haghgoo, Robyn~L. Ball, Katie~S.
  Shpanskaya, Jayne Seekins, David~A. Mong, Safwan~S. Halabi, Jesse~K.
  Sandberg, Ricky Jones, David~B. Larson, Curtis~P. Langlotz, Bhavik~N. Patel,
  Matthew~P. Lungren, and Andrew~Y. Ng.
\newblock {CheXpert}: A large chest radiograph dataset with uncertainty labels
  and expert comparison.
\newblock In {\em AAAI Conference on Artificial Intelligence}, volume~33, pages
  590--597. Association for the Advancement of Artificial Intelligence
  ({AAAI}), jul 2019.

\bibitem{jagielski2020auditing}
Matthew Jagielski, Jonathan~R. Ullman, and Alina Oprea.
\newblock Auditing differentially private machine learning: How private is
  private sgd?
\newblock In Hugo Larochelle, Marc'Aurelio Ranzato, Raia Hadsell,
  Maria{-}Florina Balcan, and Hsuan{-}Tien Lin, editors, {\em Advances in
  Neural Information Processing Systems 33: Annual Conference on Neural
  Information Processing Systems 2020, NeurIPS 2020, December 6-12, 2020,
  virtual}, 2020.

\bibitem{jeon2021gradient}
Jinwoo Jeon, Jaechang Kim, Kangwook Lee, Sewoong Oh, and Jungseul Ok.
\newblock Gradient inversion with generative image prior.
\newblock In Marc'Aurelio Ranzato, Alina Beygelzimer, Yann~N. Dauphin, Percy
  Liang, and Jennifer~Wortman Vaughan, editors, {\em Advances in Neural
  Information Processing Systems 34: Annual Conference on Neural Information
  Processing Systems 2021, NeurIPS 2021, December 6-14, 2021, virtual}, pages
  29898--29908, 2021.

\bibitem{jiang2020characterizing}
Ziheng Jiang, Chiyuan Zhang, Kunal Talwar, and Michael~C Mozer.
\newblock Characterizing structural regularities of labeled data in
  overparameterized models.
\newblock In Marina Meila and Tong Zhang, editors, {\em Proceedings of the 38th
  International Conference on Machine Learning}, volume 139 of {\em Proceedings
  of Machine Learning Research}, pages 5034--5044. PMLR, 18--24 Jul 2021.

\bibitem{jin2021cafe}
Xiao Jin, Pin-Yu Chen, Chia-Yi Hsu, Chia-Mu Yu, and Tianyi Chen.
\newblock Cafe: Catastrophic data leakage in vertical federated learning.
\newblock In M.~Ranzato, A.~Beygelzimer, Y.~Dauphin, P.S. Liang, and J.~Wortman
  Vaughan, editors, {\em Advances in Neural Information Processing Systems},
  volume~34, pages 994--1006. Curran Associates, Inc., 2021.

\bibitem{mimic-demo}
A.~Johnson, L.~Bulgarelli, T.~Pollard, S.~Horng, L.~A. Celi, and R.~Mark,
  Mimic-iv, version 0.4, PhysioNet, 2020;
  \url{https://doi.org/10.13026/a3wn-hq05}.

\bibitem{mimic-cxr-data}
A.~Johnson, T.~Pollard, R.~Mark, S.~Berkowitz, and S.~Horng, Mimic-cxr
  database, version 2.0.0, PhysioNet, 2019;
  \url{https://doi.org/10.13026/C2JT1Q}.

\bibitem{Johnson2019-yh}
Alistair E~W Johnson, Tom~J Pollard, Seth~J Berkowitz, Nathaniel~R Greenbaum,
  Matthew~P Lungren, Chih-Ying Deng, Roger~G Mark, and Steven Horng.
\newblock {MIMIC-CXR}, a de-identified publicly available database of chest
  radiographs with free-text reports.
\newblock {\em Sci Data}, 6(1):317, December 2019.

\bibitem{pmlr-v134-kairouz21a}
Peter Kairouz, M{\'{o}}nica~Ribero Diaz, Keith Rush, and Abhradeep Thakurta.
\newblock (nearly) dimension independent private {ERM} with adagrad rates via
  publicly estimated subspaces.
\newblock In Mikhail Belkin and Samory Kpotufe, editors, {\em Conference on
  Learning Theory, {COLT} 2021, 15-19 August 2021, Boulder, Colorado, {USA}},
  volume 134 of {\em Proceedings of Machine Learning Research}, pages
  2717--2746. {PMLR}, 2021.

\bibitem{DBLP:journals/natmi/KaissisZPRUTLMJ21}
Georgios Kaissis, Alexander Ziller, Jonathan Passerat-Palmbach, Th{\'{e}}o
  Ryffel, Dmitrii Usynin, Andrew Trask, Ion{\'{e}}sio Lima, Jason Mancuso,
  Friederike Jungmann, Marc-Matthias Steinborn, Andreas Saleh, Marcus Makowski,
  Daniel Rueckert, and Rickmer Braren.
\newblock End-to-end privacy preserving deep learning on multi-institutional
  medical imaging.
\newblock {\em Nature Machine Intelligence}, 3(6):473--484, may 2021.

\bibitem{KamalZNH22}
Uday Kamal, Mohammad Zunaed, Nusrat~Binta Nizam, and Taufiq Hasan.
\newblock Anatomy-{XNet}: An anatomy aware convolutional neural network for
  thoracic disease classification in chest x-rays.
\newblock {\em {IEEE} Journal of Biomedical and Health Informatics},
  26(11):5518--5528, nov 2022.

\bibitem{kaplun2022deconstructing}
Gal Kaplun, Nikhil Ghosh, Saurabh Garg, Boaz Barak, and Preetum Nakkiran.
\newblock Deconstructing distributions: A pointwise framework of learning.
\newblock Preprint arXiv:2202.09931 [cs.LG], 2022.

\bibitem{klause2022differentially}
Helena Klause, Alexander Ziller, Daniel Rueckert, Kerstin Hammernik, and
  Georgios Kaissis.
\newblock Differentially private training of residual networks with scale
  normalisation.
\newblock Preprint arXiv:2203.00324 [cs.LG], 2022.

\bibitem{kolesnikov2020big}
Alexander Kolesnikov, Lucas Beyer, Xiaohua Zhai, Joan Puigcerver, Jessica Yung,
  Sylvain Gelly, and Neil Houlsby.
\newblock Big transfer (bit): General visual representation learning.
\newblock In {\em European Conference on Computer Vision}, pages 491--507.
  Springer International Publishing, 2020.

\bibitem{kulynych2022disparate}
Bogdan Kulynych, Mohammad Yaghini, Giovanni Cherubin, Michael Veale, and
  Carmela Troncoso.
\newblock Disparate vulnerability to membership inference attacks.
\newblock In {\em Proc. Priv. Enhancing Technol.}, volume 2022, pages 460--480,
  2022.

\bibitem{Kurakin22}
Alexey Kurakin, Steve Chien, Shuang Song, Roxana Geambasu, Andreas Terzis, and
  Abhradeep Thakurta.
\newblock Toward training at imagenet scale with differential privacy.
\newblock Preprint arXiv:2201.12328 [cs.LG], 2022.

\bibitem{kwegyir2023misuse}
Kweku Kwegyir-Aggrey, Marissa Gerchick, Malika Mohan, Aaron Horowitz, and
  Suresh Venkatasubramanian.
\newblock The misuse of auc: What high impact risk assessment gets wrong.
\newblock In {\em Proceedings of the 2023 ACM Conference on Fairness,
  Accountability, and Transparency}, pages 1570--1583, 2023.

\bibitem{larrazabal2020gender}
Agostina~J Larrazabal, Nicol{\'a}s Nieto, Victoria Peterson, Diego~H Milone,
  and Enzo Ferrante.
\newblock Gender imbalance in medical imaging datasets produces biased
  classifiers for computer-aided diagnosis.
\newblock {\em Proceedings of the National Academy of Sciences},
  117(23):12592--12594, 2020.

\bibitem{DBLP:journals/nature/LeCunBH15}
Yann LeCun, Yoshua Bengio, and Geoffrey~E. Hinton.
\newblock Deep learning.
\newblock {\em Nat.}, 521(7553):436--444, 2015.

\bibitem{lecun1998gradient}
Yann LeCun, L{\'e}on Bottou, Yoshua Bengio, and Patrick Haffner.
\newblock Gradient-based learning applied to document recognition.
\newblock {\em Proceedings of the {IEEE}}, 86(11):2278--2324, 1998.

\bibitem{NEURIPS2022_b75ce884}
Xuechen Li, Daogao Liu, Tatsunori~B. Hashimoto, Huseyin~A. Inan, Janardhan
  Kulkarni, Yin{-}Tat Lee, and Abhradeep~Guha Thakurta.
\newblock When does differentially private learning not suffer in high
  dimensions?
\newblock In {\em NeurIPS}, 2022.

\bibitem{Li2021}
Xuechen Li, Florian Tram{\`{e}}r, Percy Liang, and Tatsunori Hashimoto.
\newblock Large language models can be strong differentially private learners.
\newblock In {\em The Tenth International Conference on Learning
  Representations, {ICLR} 2022, Virtual Event, April 25-29, 2022}.
  OpenReview.net, 2022.

\bibitem{loshchilov2016sgdr}
Ilya Loshchilov and Frank Hutter.
\newblock {SGDR:} stochastic gradient descent with warm restarts.
\newblock In {\em 5th International Conference on Learning Representations,
  {ICLR} 2017, Toulon, France, April 24-26, 2017, Conference Track
  Proceedings}. OpenReview.net, 2017.

\bibitem{LuoW0021}
Zelun Luo, Daniel~J. Wu, Ehsan Adeli, and Li~Fei{-}Fei.
\newblock Scalable differential privacy with sparse network finetuning.
\newblock In {\em Conference on Computer Vision and Pattern Recognition
  ({CVPR})}. {IEEE}, jun 2021.

\bibitem{flblog}
Brendan McMahan and Abhradeep Thakurta.
\newblock Federated learning with formal differential privacy guarantees.
\newblock Technical report, Google Research Blog, 2022;
  \url{https://ai.googleblog.com/2022/02/federated-learning-with-formal.html}.

\bibitem{mehta2022differentially}
Harsh Mehta, Walid Krichene, Abhradeep Thakurta, Alexey Kurakin, and Ashok
  Cutkosky.
\newblock Differentially private image classification from features.
\newblock Preprint arXiv:2211.13403 [cs.LG], 2022.

\bibitem{mehta2022large}
Harsh Mehta, Abhradeep Thakurta, Alexey Kurakin, and Ashok Cutkosky.
\newblock Large scale transfer learning for differentially private image
  classification.
\newblock Preprint arXiv:2205.02973 [cs.LG], 2022.

\bibitem{mozannar2020fair}
Hussein Mozannar, Mesrob~I. Ohannessian, and Nathan Srebro.
\newblock Fair learning with private demographic data.
\newblock In {\em Proceedings of the 37th International Conference on Machine
  Learning, {ICML} 2020, 13-18 July 2020, Virtual Event}, volume 119 of {\em
  Proceedings of Machine Learning Research}, pages 7066--7075. {PMLR}, 2020.

\bibitem{nasr2021adversary}
Milad Nasr, Shuang Song, Abhradeep Thakurta, Nicolas Papernot, and Nicholas
  Carliniz.
\newblock Adversary instantiation: Lower bounds for differentially private
  machine learning.
\newblock In {\em Symposium on Security and Privacy (SP)}. {IEEE}, may 2021.

\bibitem{nielsen2018guaranteed}
Frank Nielsen and Ke~Sun.
\newblock Guaranteed deterministic bounds on the total variation distance
  between univariate mixtures.
\newblock In {\em 28th International Workshop on Machine Learning for Signal
  Processing ({MLSP})}. {IEEE}, sep 2018.

\bibitem{ohm2009broken}
Paul Ohm.
\newblock Broken promises of privacy: Responding to the surprising failure of
  anonymization.
\newblock {\em UCLA l. Rev.}, 57:1701, 2009.

\bibitem{DBLP:conf/cvpr/OyallonM15}
Edouard Oyallon and St{\'{e}}phane Mallat.
\newblock Deep roto-translation scattering for object classification.
\newblock In {\em Conference on Computer Vision and Pattern Recognition}.
  {IEEE}, jun 2015.

\bibitem{DBLP:conf/aaai/PapernotT0CE21}
Nicolas Papernot, Abhradeep Thakurta, Shuang Song, Steve Chien, and {\'{U}}lfar
  Erlingsson.
\newblock Tempered sigmoid activations for deep learning with differential
  privacy.
\newblock In {\em AAAI Conference on Artificial Intelligence}, volume~35, pages
  9312--9321. Association for the Advancement of Artificial Intelligence
  ({AAAI}), may 2021.

\bibitem{DBLP:conf/nips/PaszkeGMLBCKLGA19}
Adam Paszke, Sam Gross, Francisco Massa, Adam Lerer, James Bradbury, Gregory
  Chanan, Trevor Killeen, Zeming Lin, Natalia Gimelshein, Luca Antiga, Alban
  Desmaison, Andreas K{\"{o}}pf, Edward~Z. Yang, Zachary DeVito, Martin Raison,
  Alykhan Tejani, Sasank Chilamkurthy, Benoit Steiner, Lu~Fang, Junjie Bai, and
  Soumith Chintala.
\newblock Pytorch: An imperative style, high-performance deep learning library.
\newblock In Hanna~M. Wallach, Hugo Larochelle, Alina Beygelzimer, Florence
  d'Alch{\'{e}}{-}Buc, Emily~B. Fox, and Roman Garnett, editors, {\em Advances
  in Neural Information Processing Systems 32: Annual Conference on Neural
  Information Processing Systems 2019, NeurIPS 2019, December 8-14, 2019,
  Vancouver, BC, Canada}, pages 8024--8035, 2019.

\bibitem{pham2021meta}
Hieu Pham, Zihang Dai, Qizhe Xie, and Quoc~V Le.
\newblock Meta pseudo labels.
\newblock In {\em Conference on Computer Vision and Pattern Recognition
  ({CVPR})}. {IEEE}, jun 2021.

\bibitem{pinto2023pillar}
Francesco Pinto, Yaxi Hu, Fanny Yang, and Amartya Sanyal.
\newblock Pillar: How to make semi-private learning more effective.
\newblock Preprint arXiv:2306.03962 [cs.LG], 2023.

\bibitem{rastegarpanah2020fair}
Bashir Rastegarpanah, Mark Crovella, and Krishna~P Gummadi.
\newblock Fair inputs and fair outputs: The incompatibility of fairness in
  privacy and accuracy.
\newblock In {\em Adjunct Publication of the 28th ACM Conference on User
  Modeling, Adaptation and Personalization}, pages 260--267. {ACM}, jul 2020.

\bibitem{DBLP:conf/nips/RidnikBNZ21}
Tal Ridnik, Emanuel~Ben Baruch, Asaf Noy, and Lihi Zelnik.
\newblock Imagenet-21k pretraining for the masses.
\newblock In Joaquin Vanschoren and Sai{-}Kit Yeung, editors, {\em Proceedings
  of the Neural Information Processing Systems Track on Datasets and Benchmarks
  1, NeurIPS Datasets and Benchmarks 2021, December 2021, virtual}, 2021.

\bibitem{ILSVRC15}
Olga Russakovsky, Jia Deng, Hao Su, Jonathan Krause, Sanjeev Satheesh, Sean Ma,
  Zhiheng Huang, Andrej Karpathy, Aditya Khosla, Michael Bernstein,
  Alexander~C. Berg, and Li~Fei-Fei.
\newblock {ImageNet} large scale visual recognition challenge.
\newblock {\em International Journal of Computer Vision}, 115(3):211--252, apr
  2015.

\bibitem{sander2022tan}
Tom Sander, Pierre Stock, and Alexandre Sablayrolles.
\newblock Tan without a burn: Scaling laws of {DP-SGD}.
\newblock Preprint arXiv:2210.03403 [cs.LG], 2022.

\bibitem{Santos-Lozada2020-go}
Alexis~R Santos-Lozada, Jeffrey~T Howard, and Ashton~M Verdery.
\newblock How differential privacy will affect our understanding of health
  disparities in the united states.
\newblock {\em Proc. Natl. Acad. Sci. U. S. A.}, 117(24):13405--13412, June
  2020.

\bibitem{sanyal2022unfair}
Amartya Sanyal, Yaxi Hu, and Fanny Yang.
\newblock How unfair is private learning?
\newblock In James Cussens and Kun Zhang, editors, {\em Proceedings of the
  Thirty-Eighth Conference on Uncertainty in Artificial Intelligence}, volume
  180 of {\em Proceedings of Machine Learning Research}, pages 1738--1748.
  PMLR, 01--05 Aug 2022.

\bibitem{Seyyed-Kalantari21}
Laleh Seyyed{-}Kalantari, Guanxiong Liu, Matthew B.~A. McDermott, Irene~Y.
  Chen, and Marzyeh Ghassemi.
\newblock Chexclusion: Fairness gaps in deep chest x-ray classifiers.
\newblock In {\em Biocomputing 2021: Proceedings of the Pacific Symposium,
  Kohala Coast, Hawaii, USA, January 3-7, 2021}. WorldScientific, 2021.

\bibitem{seyyed2021underdiagnosis}
Laleh Seyyed-Kalantari, Haoran Zhang, Matthew McDermott, Irene~Y Chen, and
  Marzyeh Ghassemi.
\newblock Underdiagnosis bias of artificial intelligence algorithms applied to
  chest radiographs in under-served patient populations.
\newblock {\em Nature Medicine}, 27(12):2176--2182, dec 2021.

\bibitem{shorten2019survey}
Connor Shorten and Taghi~M Khoshgoftaar.
\newblock A survey on image data augmentation for deep learning.
\newblock {\em Journal of Big Data}, 6(1):60, jul 2019.

\bibitem{DBLP:journals/corr/SimonyanZ14a}
Karen Simonyan and Andrew Zisserman.
\newblock Very deep convolutional networks for large-scale image recognition.
\newblock In Yoshua Bengio and Yann LeCun, editors, {\em 3rd International
  Conference on Learning Representations, {ICLR} 2015, San Diego, CA, USA, May
  7-9, 2015, Conference Track Proceedings}, 2015.

\bibitem{pmlr-v130-song21a}
Shuang Song, Thomas Steinke, Om~Thakkar, and Abhradeep Thakurta.
\newblock Evading the curse of dimensionality in unconstrained private glms.
\newblock In Arindam Banerjee and Kenji Fukumizu, editors, {\em The 24th
  International Conference on Artificial Intelligence and Statistics, {AISTATS}
  2021, April 13-15, 2021, Virtual Event}, volume 130 of {\em Proceedings of
  Machine Learning Research}, pages 2638--2646. {PMLR}, 2021.

\bibitem{sun2017revisiting}
Chen Sun, Abhinav Shrivastava, Saurabh Singh, and Abhinav Gupta.
\newblock Revisiting unreasonable effectiveness of data in deep learning era.
\newblock In {\em International Conference on Computer Vision ({ICCV})}.
  {IEEE}, oct 2017.

\bibitem{suriyakumar2021chasing}
Vinith~M Suriyakumar, Nicolas Papernot, Anna Goldenberg, and Marzyeh Ghassemi.
\newblock Chasing your long tails: Differentially private prediction in health
  care settings.
\newblock In {\em Proceedings of the {ACM} Conference on Fairness,
  Accountability, and Transparency}. {ACM}, mar 2021.

\bibitem{szegedy2016rethinking}
Christian Szegedy, Vincent Vanhoucke, Sergey Ioffe, Jon Shlens, and Zbigniew
  Wojna.
\newblock Rethinking the inception architecture for computer vision.
\newblock In {\em Conference on Computer Vision and Pattern Recognition
  ({CVPR})}, pages 2818--2826. {IEEE}, jun 2016.

\bibitem{tan2019efficientnet}
Mingxing Tan and Quoc~V. Le.
\newblock Efficientnet: Rethinking model scaling for convolutional neural
  networks.
\newblock In Kamalika Chaudhuri and Ruslan Salakhutdinov, editors, {\em
  Proceedings of the 36th International Conference on Machine Learning, {ICML}
  2019, 9-15 June 2019, Long Beach, California, {USA}}, volume~97 of {\em
  Proceedings of Machine Learning Research}, pages 6105--6114. {PMLR}, 2019.

\bibitem{tang2023differentially}
Xinyu Tang, Ashwinee Panda, Vikash Sehwag, and Prateek Mittal.
\newblock Differentially private image classification by learning priors from
  random processes.
\newblock Preprint arXiv:2306.06076 [cs.CV], 2023.

\bibitem{appleblog}
Differential~Privacy Team.
\newblock Learning with privacy at scale.
\newblock Technical report, Machine Learning Research at Apple, 2017;
  \url{https://machinelearning.apple.com/research/learning-with-privacy-at-scale}.

\bibitem{DBLP:conf/iclr/TramerB21}
Florian Tram{\`{e}}r and Dan Boneh.
\newblock Differentially private learning needs better features (or much more
  data).
\newblock In {\em 9th International Conference on Learning Representations,
  {ICLR} 2021, Virtual Event, Austria, May 3-7, 2021}. OpenReview.net, 2021.

\bibitem{tramer2022considerations}
Florian Tram{\`e}r, Gautam Kamath, and Nicholas Carlini.
\newblock Considerations for differentially private learning with large-scale
  public pretraining.
\newblock Preprint arXiv:2212.06470 [cs.LG], 2022.

\bibitem{tramer2022debugging}
Florian Tramer, Andreas Terzis, Thomas Steinke, Shuang Song, Matthew Jagielski,
  and Nicholas Carlini.
\newblock Debugging differential privacy: A case study for privacy auditing.
\newblock Preprint arXiv:2202.12219 [cs.LG], 2022.

\bibitem{tran2021differentially}
Cuong Tran, My~H Dinh, and Ferdinando Fioretto.
\newblock Differentially private deep learning under the fairness lens.
\newblock Preprint arXiv:2106.02674 [cs.LG], 2021.

\bibitem{uniyal2021dp}
Archit Uniyal, Rakshit Naidu, Sasikanth Kotti, Sahib Singh, Patrik~Joslin
  Kenfack, Fatemehsadat Mireshghallah, and Andrew Trask.
\newblock {DP-SGD} vs {PATE}: Which has less disparate impact on model
  accuracy?
\newblock Preprint arXiv:2106.12576 [cs.LG], 2021.

\bibitem{DBLP:journals/corr/abs-2007-05089}
Laurens van~der Maaten and Awni~Y. Hannun.
\newblock The trade-offs of private prediction.
\newblock Preprint arXiv:2007.05089 [cs.LG], 2020.

\bibitem{van2018inaturalist}
Grant Van~Horn, Oisin Mac~Aodha, Yang Song, Yin Cui, Chen Sun, Alex Shepard,
  Hartwig Adam, Pietro Perona, and Serge Belongie.
\newblock The inaturalist species classification and detection dataset.
\newblock In {\em Conference on Computer Vision and Pattern Recognition}.
  {IEEE}, jun 2018.

\bibitem{vaswani2017attention}
Ashish Vaswani, Noam Shazeer, Niki Parmar, Jakob Uszkoreit, Llion Jones,
  Aidan~N. Gomez, Lukasz Kaiser, and Illia Polosukhin.
\newblock Attention is all you need.
\newblock In Isabelle Guyon, Ulrike von Luxburg, Samy Bengio, Hanna~M. Wallach,
  Rob Fergus, S.~V.~N. Vishwanathan, and Roman Garnett, editors, {\em Advances
  in Neural Information Processing Systems 30: Annual Conference on Neural
  Information Processing Systems 2017, December 4-9, 2017, Long Beach, CA,
  {USA}}, pages 5998--6008, 2017.

\bibitem{welford1962note}
BP~Welford.
\newblock Note on a method for calculating corrected sums of squares and
  products.
\newblock {\em Technometrics}, 4(3):419--420, aug 1962.

\bibitem{xu2020removing}
Depeng Xu, Wei Du, and Xintao Wu.
\newblock Removing disparate impact of differentially private stochastic
  gradient descent on model accuracy.
\newblock Preprint arXiv:2003.03699 [cs.LG], 2020.

\bibitem{xu2021robust}
Han Xu, Xiaorui Liu, Yaxin Li, Anil~K. Jain, and Jiliang Tang.
\newblock To be robust or to be fair: Towards fairness in adversarial training.
\newblock In Marina Meila and Tong Zhang, editors, {\em Proceedings of the 38th
  International Conference on Machine Learning, {ICML} 2021, 18-24 July 2021,
  Virtual Event}, volume 139 of {\em Proceedings of Machine Learning Research},
  pages 11492--11501. {PMLR}, 2021.

\bibitem{yin2021see}
Hongxu Yin, Arun Mallya, Arash Vahdat, Jose~M Alvarez, Jan Kautz, and Pavlo
  Molchanov.
\newblock See through gradients: Image batch recovery via gradinversion.
\newblock In {\em Proceedings of the IEEE/CVF Conference on Computer Vision and
  Pattern Recognition}, pages 16337--16346. {IEEE}, jun 2021.

\bibitem{you2019does}
Kaichao You, Mingsheng Long, Jianmin Wang, and Michael~I Jordan.
\newblock How does learning rate decay help modern neural networks?
\newblock Preprint arXiv:1908.01878 [cs.LG], 2019.

\bibitem{Yu2021lm}
Da~Yu, Saurabh Naik, Arturs Backurs, Sivakanth Gopi, Huseyin~A. Inan, Gautam
  Kamath, Janardhan Kulkarni, Yin~Tat Lee, Andre Manoel, Lukas Wutschitz,
  Sergey Yekhanin, and Huishuai Zhang.
\newblock Differentially private fine-tuning of language models.
\newblock In {\em The Tenth International Conference on Learning
  Representations, {ICLR} 2022, Virtual Event, April 25-29, 2022}.
  OpenReview.net, 2022.

\bibitem{YuZ0L21}
Da~Yu, Huishuai Zhang, Wei Chen, and Tie{-}Yan Liu.
\newblock Do not let privacy overbill utility: Gradient embedding perturbation
  for private learning.
\newblock In {\em 9th International Conference on Learning Representations,
  {ICLR} 2021, Virtual Event, Austria, May 3-7, 2021}. OpenReview.net, 2021.

\bibitem{yu2021large}
Da~Yu, Huishuai Zhang, Wei Chen, Jian Yin, and Tie{-}Yan Liu.
\newblock Large scale private learning via low-rank reparametrization.
\newblock In Marina Meila and Tong Zhang, editors, {\em Proceedings of the 38th
  International Conference on Machine Learning, {ICML} 2021, 18-24 July 2021,
  Virtual Event}, volume 139 of {\em Proceedings of Machine Learning Research},
  pages 12208--12218. {PMLR}, 2021.

\bibitem{yu2023vip}
Yaodong Yu, Maziar Sanjabi, Yi~Ma, Kamalika Chaudhuri, and Chuan Guo.
\newblock Vip: {A} differentially private foundation model for computer vision.
\newblock Preprint arXiv:2306.08842 [cs.CV], 2023.

\bibitem{DBLP:conf/iccv/Yuan0SY21}
Zhuoning Yuan, Yan Yan, Milan Sonka, and Tianbao Yang.
\newblock Large-scale robust deep {AUC} maximization: {A} new surrogate loss
  and empirical studies on medical image classification.
\newblock In {\em 2021 {IEEE/CVF} International Conference on Computer Vision,
  {ICCV} 2021, Montreal, QC, Canada, October 10-17, 2021}, pages 3020--3029.
  {IEEE}, 2021.

\bibitem{zhai2022scaling}
Xiaohua Zhai, Alexander Kolesnikov, Neil Houlsby, and Lucas Beyer.
\newblock Scaling vision transformers.
\newblock In {\em Conference on Computer Vision and Pattern Recognition
  ({CVPR})}. {IEEE}, jun 2022.

\bibitem{DBLP:journals/cacm/ZhangBHRV21}
Chiyuan Zhang, Samy Bengio, Moritz Hardt, Benjamin Recht, and Oriol Vinyals.
\newblock Understanding deep learning (still) requires rethinking
  generalization.
\newblock {\em Commun. {ACM}}, 64(3):107--115, 2021.

\bibitem{zhang2022improving}
Haoran Zhang, Natalie Dullerud, Karsten Roth, Lauren Oakden{-}Rayner, Stephen
  Pfohl, and Marzyeh Ghassemi.
\newblock Improving the fairness of chest x-ray classifiers.
\newblock In Gerardo Flores, George~H. Chen, Tom~J. Pollard, Joyce~C. Ho, and
  Tristan Naumann, editors, {\em Conference on Health, Inference, and Learning,
  {CHIL} 2022, 7-8 April 2022, Virtual Event}, volume 174 of {\em Proceedings
  of Machine Learning Research}, pages 204--233. {PMLR}, 2022.

\bibitem{DBLP:journals/pami/ZhouLKO018}
Bolei Zhou, {\`{A}}gata Lapedriza, Aditya Khosla, Aude Oliva, and Antonio
  Torralba.
\newblock Places: {A} 10 million image database for scene recognition.
\newblock {\em {IEEE} Trans. Pattern Anal. Mach. Intell.}, 40(6):1452--1464,
  2018.

\bibitem{zhu2019deep}
Ligeng Zhu and Song Han.
\newblock {\em Deep Leakage from Gradients}, pages 17--31.
\newblock Springer International Publishing, 2020.

\bibitem{ziller2021medical}
Alexander Ziller, Dmitrii Usynin, Rickmer Braren, Marcus Makowski, Daniel
  Rueckert, and Georgios Kaissis.
\newblock Medical imaging deep learning with differential privacy.
\newblock {\em Scientific Reports}, 11(1):13524, 2021.

\bibitem{rapporblog}
Úlfar Erlingsson.
\newblock Learning statistics with privacy, aided by the flip of a coin.
\newblock Technical report, Google Research Blog, 2014;
  \url{https://ai.googleblog.com/2014/10/learning-statistics-with-privacy-aided.html}.

\end{thebibliography}

\pagebreak

\begin{appendices}

\section{Background: Differential Privacy and DP-SGD}

\subsection{Differential Privacy}

Differential privacy (DP) \cite{dwork2006calibrating} is a formal privacy guarantee that applies to randomized data analysis algorithms. 
By construction, differentially private algorithms prevent an adversary that observes the output of a computation from inferring any property pertaining to individual data points in the input data used during the computation. 
The strength of this guarantee is controlled by two parameters: $\varepsilon > 0$ and $\delta \in [0,1]$. 
Intuitively, $\varepsilon$ bounds the log-likelihood ratio of any particular output that can be obtained when running the algorithm on two datasets differing in a single data point, and $\delta$ is a small probability which bounds the occurrence of infrequent outputs that violate this bound. 
The privacy guarantee becomes stronger as both parameters get smaller. 
A standard rule of thumb states that, to obtain meaningful privacy, $\varepsilon$ should be a small constant while $\delta$ should be smaller than $1/ N$, where $N$ is the size of the input dataset. 
More formally, we have the following.

\textbf{Differential Privacy.}
Let $A: \mathcal{D} \to \mathcal{S}$ be a randomized algorithm, and let $\varepsilon > 0$, $\delta \in [0, 1]$.
We say that $A$ is $(\varepsilon, \delta)$-DP if for any two neighboring datasets $D, D' \in \mathcal{D}$ differing by a single element, we have that
\[
\forall \: S \subset \mathcal{S}, \: \Pr[A(D) \in S] \leq \exp(\varepsilon) \Pr[A(D') \in S] + \delta \enspace.
\label{eq:dp_definition}
\]

The privacy protection afforded by DP holds under an exceedingly strong threat model: inferences about individuals are protected even in the face of an adversary that has full knowledge of the DP algorithm, unbounded computational power, and arbitrary side knowledge about the input data. Broadly speaking, by observing the output the adversary cannot learn anything about the input data which they did not already know from their side knowledge.
Furthermore, DP satisfies a number of appealing properties from the algorithm design standpoint, including preservation under post-processing and a smooth degradation with multiple accesses to the same data.
These properties are exploited in the construction of complex DP algorithms based on the combination of small building blocks that inject carefully calibrated noise into operations that access the data.
The magnitude of the noise required to satisfy the privacy guarantee increases with the strength of the privacy parameters, leading to an unavoidable trade-off between utility and privacy (e.g.\ as illustrated by the Fundamental Law of Information Recovery \cite{TCS-042}), implying it is impossible to fully close the performance gap between private and non-private learning.

\subsection{Interpreting the Differential Privacy Guarantee}

\textbf{Threat model.} The standard formulation of DP given above is based on a strong notion of (information-theoretic) indistinguishability between the probability distributions over outputs produced by $A(D)$ and $A(D')$. Implicit in the definition there is an adversary whose goal is to infer whether the individual in which the datasets $D$ and $D'$ differ was actually part of the computation that produced the output observed by the adversary. The set $S$ then corresponds to a test the adversary runs on the output -- the adversary's decision is based on whether the output is included or not in $S$ -- and the quantification over all sets $S$ is akin to assuming the adversary will use the best possible test. In particular, the adversary is free to design the test based on full knowledge of the algorithm $A$ and the pair of datasets $D$ and $D'$ under consideration. Such worst-case assumptions are what makes DP a robust and flexible notion of privacy, but at the same time make the guarantee hard to interpret for non-experts. We now describe two alternative interpretations of the privacy afforded by DP in terms of resilience to specific privacy attacks.

\textbf{Membership inference interpretation.} Membership inference attacks are a class of privacy attacks where the adversary's goal is to determine whether the data of a particular individual was used in a certain computation (e.g.\ training a machine learning model) based on the computation's output. This goal is identical to the one which DP is designed to protect against, but the attack can be formalized in a number of threat models depending on the knowledge available to the adversary, e.g.\ about the other individuals in the training dataset. In the threat model of DP where the adversary has full knowledge, it is possible to directly translate the $(\varepsilon,\delta)$-DP guarantee into protection against membership inference attacks by rephrasing the latter as a simple statistical hypothesis testing problem.

\begin{figure}[htpb]
    \centering
    \includegraphics{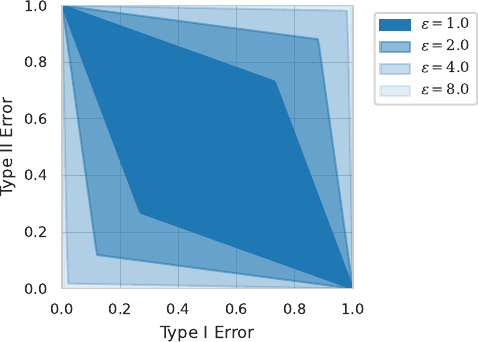}
    \caption{\small Constraints on the error rates of membership inference attacks implied by $(\varepsilon, 10^{-5})$-DP algorithms. For each value of $\varepsilon$, the corresponding shaded region identifies values for the Type I and Type II error rates allowed under the DP constraint. An adversary would like to design an attack that is close as possible to the lower left corner, and DP with small $\varepsilon$ precludes that possibility, thus forcing the adversary to make frequent mistakes no matter how strong their membership inference attack is.}
    \label{fig:privacy-region}
\end{figure}

Assuming that the adversary's null hypothesis is that a target individual was not included in the dataset and the alternative hypothesis that the individual was included, the error rates of Type I ($\alpha$) and Type II ($\beta$) correspond, respectively, to the adversary concluding the individual was included when they actually were not, and the adversary concluding the individual was not included when they actually were. The DP guarantee then implies a constraint that makes it impossible for the adversary to find an attack that simultaneously achieves small Type I and Type II error.
More formally, any membership inference attack against an $(\varepsilon, \delta)$-DP algorithm must satisfy $\ln\left(\frac{1-\beta-\delta}{\alpha}\right) \leq \varepsilon$.
One of the benefits of the strong threat model implicit in the DP definition is that this constraint automatically applies to any other type of adversary with less access to privileged information.
\Cref{fig:privacy-region} illustrates the regions of valid Type I and Type II error rates that are implied by this constraint for DP algorithms across a range of $\varepsilon$'s.

An important limitation of the connection between DP guarantees and resilience to membership inference is that the resulting guarantee becomes meaningless very quickly as $\varepsilon$ grows -- note, for example, that at $\varepsilon = 8$ the constraint illustrated in \Cref{fig:privacy-region} is nearly vacuous. This is not an artifact -- it is known that such membership inference guarantees are tight \cite{nasr2021adversary} -- but it does not mean that DP does not provide any protection for $\varepsilon = 8$ either. It is, in fact, a limitation of relying on a threat model where the adversary possesses enough side information to narrow down the choices of the privacy attack to two options: deciding whether the target individual was included or not, or, equivalently, deciding whether the data associated with the target individual is one of two possibilities. Note that this is an extremely strong threat model where the adversary already has enough side information to be 50\% confident that the target individual was included in the training set before observing the trained model.

It is worth noting that some DP mechanisms satisfy stronger notions of privacy corresponding to multiple $(\epsilon, \delta)$ pairs simultaneously. In such cases the trade-off curves from \Cref{fig:privacy-region} can be refined to take all pairs of parameters into account \cite{dong2022gaussian}.

\begin{figure}[htpb]
    \centering
    \begin{minipage}{.45\textwidth}
    \includegraphics{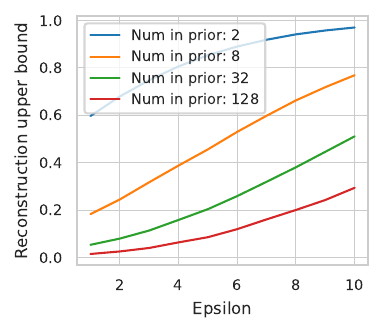}
    \vfill
    \end{minipage}
    \begin{minipage}{.45\textwidth}
    \includegraphics{assets/crowd-size-multi-mia-alt.pdf}
    \vfill
    \end{minipage}
    \caption{\small Constraints on the error rates of $K$-choice membership inference attacks implied by $(\varepsilon, 10^{-5})$-DP algorithms. The left plot shows the maximum accuracy of a membership inference attack for a range of $\varepsilon$ values, which depends on the number of candidate images $K$. The attacker has to identify the target image used during training out of the $K$ candidates, which are assumed to be equally likely under the attacker's side knowledge. The right plot shows the maximum number of points that can be contained in the prior for the attack to be more than 50\% accurate. For instance at $\varepsilon=8$, the membership attack will succeed with higher than 50\% probability if the attacker obtains enough side knowledge to narrow the target image down to one of (at most) 18 equally likely candidates.}
    \label{fig:privacy-k-mia}
\end{figure}

\textbf{Multi-choice membership inference interpretation.} To ascribe meaningful and quantifiable privacy protections to large values of $\varepsilon$ one can consider a mild relaxation of the threat model where the goal of the adversary is to guess which of $K > 2$ equally likely choices corresponds to the target individual. Such a \emph{$K$-choice membership inference} attack captures a scenario where, although the adversary knows the full details of the algorithm and all the dataset except one individual, they have not been able to collect enough side information to narrow down the data of the unknown target individual to only two choices. For example, in a setting where the dataset contains the birth date of individuals, the adversary might already know the year when the target individual was born, but without additional side knowledge they still have 365 equally likely choices for what the correct day and month are. Methods recently developed in \cite{DBLP:journals/corr/abs-2210-13662,DBLP:journals/corr/abs-2302-07225} can be used to obtain the probability that the adversary guesses which of the $K$ choices is correct based on the output of a DP algorithm as a function of its privacy parameters. These are illustrated in \Cref{fig:privacy-k-mia}, where we observe that, for example, at $\varepsilon = 8$ the DP guarantee prevents the adversary from reliably guessing an individual's data correctly as soon as they have at least 18 equally likely options to choose from.
Thus, a DP-classifier at $\varepsilon = 8$ does provide meaningful protection, so long as the attacker is not able to obtain substantial side information from other sources.

\subsection{Differentially Private Stochastic Gradient Descent}
\label{sec:background_dpsgd}

In this work, we work with differentially private algorithms $A$ for supervised deep learning. This means that $A$ maps a training dataset $D=\{(x_i, y_i)\}_{1 \leq i \leq N}$ to a vector of learned neural network parameters $w \in \mathcal{S} = \mathbb{R}^p$.
Let $\mathcal{L}(w, x, y)$ denote the learning objective (e.g., the cross-entropy loss), given the model parameters $w$, input example $x$ and label $y$. For convenience, we use the shorthand notation $l_i(w) = \mathcal{L}(w, x_i, y_i)$.

In the non-private setting, Stochastic Gradient Descent (SGD) provides a standard iterative optimization approach to learning, whereby at each iteration $t$ the algorithm draws $B$ examples at random from the training dataset, and updates the model parameters according to:
\[
w^{(t+1)} = w^{(t)} - \eta_t \frac{1}{B} \sum\limits_{i \in \mathcal{B}_t} \nabla l_i (w^{(t)}) \enspace,
\]
where $\eta_t$ is the step-size for the $t${th} update, $\nabla$ denotes the gradient operator, and $\mathcal{B}_t$ represents the set of examples sampled at iteration $t$ with $|\mathcal{B}_t|=B$.

In order to make this algorithm differentially private, we apply the following modifications. 
First, the gradient for each example in the mini-batch is clipped to a maximal norm $C$ and normalized by $C$, and second, Gaussian noise with an appropriate standard deviation is added to the mean of the clipped gradients.
Let $\texttt{clip}_C(v) = \min\left\{1, \frac{C}{\|v\|_2}\right\} \cdot v$ denote the clipping function which re-scales its input so that the output has a maximal $\ell_2$ norm of $C$.
The new update step is:
\[
w^{(t+1)} = w^{(t)} - \eta_t \left\{ \frac{1}{B} \sum\limits_{i \in \mathcal{B}_t}  \frac{1}{C}\texttt{clip}_C \left(\nabla l_i (w^{(t)}) \right) + \frac{\sigma}{B} \xi \right\} \enspace,
\]
where $\xi \sim \mathcal{N}(0, I_p)$ is a standard $p$-dimensional Gaussian random variable and $\sigma$ specifies the standard deviation of the added noise.
The resulting algorithm is called Differentially Private-Stochastic Gradient Descent (DP-SGD) \cite{abadi2016deep}.
We note that this is a minor re-parameterization of the algorithm in \cite{abadi2016deep} in which the learning rate $\eta_t$ absorbs a factor of $C$.
This has no effect on the privacy guarantees, but ensures the clipping norm does not influence the scale of the update, which simplifies hyper-parameter tuning in practice.

\textbf{Privacy accounting.}
Intuitively, performing a model update using DP-SGD provides differential privacy because adding Gaussian noise with standard deviation proportional to $C$ is sufficient to mask the contribution of any single example whose clipped gradient has norm less than or equal to $C$.
The total privacy guarantee of DP-SGD is determined by three parameters: the standard deviation $\sigma$, the sampling ratio $q = B / N$ and the number of training iterations $T$.
In practice, the privacy budget $(\varepsilon, \delta)$ is usually fixed, and these three hyper-parameters are chosen to provide the best possible performance within this budget. 
The privacy calibration process is performed using a privacy accountant: a numerical algorithm providing tight upper bounds for the privacy budget as a function of the hyper-parameters \cite{abadi2016deep}, which in turn can be combined with numerical optimization routines to optimize one hyper-parameter given the privacy budget and the other two hyper-parameters.
The privacy accounting for DP-SGD relies on a ``composition'' analysis across iterations, which allows for the release not only of the final model, but also of \emph{every} intermediate model obtained during training (under the same privacy budget).

\subsection{DP-SGD: Implementation Details}\label{app:implementation}

All the experiments reported in the paper use a JAX \cite{jax2018github} implementation of DP-SGD based on JAXline \cite{deepmind2020jax}, a re-usable framework for distributed model training and evaluation.
For privacy accounting, our implementation uses a method based on \emph{privacy loss distributions} proposed in \cite{DBLP:journals/popets/DoroshenkoGKKM22}, and implemented in Google's DP library \cite{googledplibrary}.

Besides enabling reproducibility of our results, another important reason for open sourcing our code is to allow the differential privacy community to verify the correctness our implementation of DP-SGD.
To help navigate the code base, we provide an in-depth description of how our code is structured to parallelize the computation of privatized gradients in DP-SGD across many devices when using virtual batching and multiple augmentations (see \cite{de2022unlocking}).
Furthermore, we describe a collection of privacy auditing experiments aimed at obtaining empirical lower bounds for the privacy of our implementation by leveraging membership inference attacks.

Algorithm~\ref{alg:dpsgd-jaxline} provides a high-level description of how our DP-SGD implementation computes the privatized gradient used in each model update step. The structure of the code is informed by the way model training pipelines are implemented in JAXline.
The implementation is parallelized across $N_{dev}$ devices, where each device runs a copy of Algorithm~\ref{alg:dpsgd-jaxline}. To extract the maximum possible throughput from the implementation, each device processes training examples in batches of size $B_{local}$, where this parameter is adjusted depending on the memory available in each device and the size of model gradients for the present architecture. To accommodate settings where the desired batch size for a single model update is larger than $B_{local} \cdot N_{dev}$, our implementation incorporates gradient accumulation (i.e.\ virtual batching) where $N_{acc}$ gradient accumulation steps are performed before each model update, giving a total batch size $B = B_{local} \cdot N_{dev} \cdot N_{acc}$.

\newcommand{\Nacc}{N_{acc}}
\newcommand{\Ndev}{N_{dev}}
\newcommand{\Naug}{K}
\newcommand{\mlocal}{B_{local}}
\begin{algorithm}[t]
\DontPrintSemicolon
\KwIn{
Current model parameters $w$,
clipping norm $C$,
noise multiplier $\sigma$,
device id $d$,
per-device per-step batch-size $\mlocal$,
number of gradient accumulation steps $\Nacc$,
number of devices $\Ndev$,
number of per-example augmentations $\Naug$,
shared noise sample $\xi \sim \mathcal{N}(0,I)$,
training examples $\{(x_{d,s,i}, y_{d,s,i}) : s \in [\Nacc], i \in [\mlocal] \}$.
}
$B \gets \mlocal \cdot \Ndev \cdot \Nacc$\;
$g \gets 0$\;
\For{$s \in \{1,\ldots, \Nacc\}$}{
    \For{$i \in \{1, \ldots, \mlocal\}$}{
        $g_{d,s,i} \gets \frac{1}{C} \texttt{clip}_C\left(\frac{1}{\Naug} \sum_{j=1}^{\Naug} \nabla \mathcal{L}(w, \texttt{augment}(x_{d,s,i}), y_{d,s,i})\right)$\;
    }
    $g_{d,s} \gets \frac{1}{\mlocal} \sum_i g_{d,s,i}$ \tcp*[htpb]{Average over local mini-batch}\;
    $\hat{g}_{d,s} \gets g_{d,s} + \frac{\sigma}{B} \xi$ \tcp*[htpb]{Add the same noise at each aggregation step on each device}\;
    $\bar{g}_{s} \gets \frac{1}{\Ndev} \sum_{d'} \hat{g}_{d',s}$ \tcp*[htpb]{Synchronize average gradient across devices}\;
    $g \gets g + \frac{\bar{g}_{s} - g}{s}$ \tcp*[htpb]{Numerically stable averaging across accumulation steps}\;
}
\KwRet{$g$} \tcp*[htpb]{Each device gets the same gradient}

\caption{\small Private gradient computation across multiple devices with virtual-batching, multiple augmentations, synchronized noise and gradient normalization.}\label{alg:dpsgd-jaxline}
\end{algorithm}

As input to the gradient computation step, each device receives the current model parameters $w$ (which are identical across devices), the desired clipping norm $C$ and noise standard deviation $\sigma$, and their device identifier $d \in \{1, \ldots, N_{dev}\}$. In addition, each device $d$ has access to $N_{acc} \cdot B_{local}$ training examples $\{(x_{d,s,i}, y_{d,s,i}) : s \in [N_{acc}], i \in [B_{local}] \}$, and a copy of a sample $\xi$ from a standard multivariate Gaussian distribution.
Crucially, the noise is resampled independently each time Algorithm~\ref{alg:dpsgd-jaxline} is executed, but it is \emph{shared} across devices and aggregation steps during a single execution.
This is enforced in our implementation by broadcasting the same pseudo-random number generator key to all devices -- this is preferred over having a different PRNG key per-device because it makes the pipeline more reproducible across training infrastructures with different numbers of devices.  

In the innermost loop of Algorithm~\ref{alg:dpsgd-jaxline}, client $d$ computes the individual contribution $g_{d,s,i}$ of a single example $x_{d,s,i}$ (where $s$ indexes the accumulation step and $i$ the local batch-size). 
Our implementation enables the use of multiple random augmentations \cite{Fort2021}, with the $K$ augmentations corresponding to a single data point being denoted by $x_{d,s,i}$. These augmentations are returned by \emph{independent} calls to the $\texttt{augment}$ subroutine. The result of this average is then clipped to a maximal $\ell_2$ norm of $C$ by the $\texttt{clip}_C$ function and then normalized by $C$. This results in a contribution $g_{d,s,i}$ to the model update with norm bounded by 1, which is then averaged locally over $i$ on device $d$ to produce $g_{d,s}$. We note that although the loop over $i \in [B_{local}]$ is presented in Algorithm~\ref{alg:dpsgd-jaxline} as a sequential computation for convenience, in reality our implementation uses hardware parallelism offered by modern accelerators -- this means that as long as the device can process $B_{local} \cdot K$ examples in parallel, the cost of computing $g_{d,s}$ is constant in these parameters.

After computing the average of clipped gradient contributions $g_{d,s}$, each device $d$ adds appropriately calibrated Gaussian noise to obtain $\hat{g}_{d,s}$ -- we emphasize again that all device share the same random noise sample, so the noise adding process is not independent across devices. At this point devices synchronize their updates by computing the average of $\hat{g}_{d,s}$ over $d \in [N_{dev}]$. After this step, each device has the same averaged noisy gradient $\bar{g}_s$. Finally, each device updates their local copy of the accumulated gradient $g$ by using Welford's incremental averaging algorithm for numerical stability \cite{welford1962note}.

That Algorithm~\ref{alg:dpsgd-jaxline} provides a correct implementation of the privatized gradients required by DP-SGD follows by comparing the following result to the formal definition of DP-SGD.

\textbf{Correctness claim.}
Each device participating in Algorithm~\ref{alg:dpsgd-jaxline} obtains the same noisy gradient given by
\[
g = \left\{\frac{1}{B} \sum_{d=1}^{N_{dev}} \sum_{s = 1}^{N_{acc}} \sum_{i = 1}^{B_{local}} \frac{1}{C} \texttt{clip}_C\left(\frac{1}{K} \sum_{j=1}^{K} \nabla \mathcal{L}(w, \texttt{augment}(x_{d,s,i}), y_{d,s,i})\right)\right\} + \frac{\sigma}{B} \xi \enspace,
\]
where $\xi \sim \mathcal{N}(0, I)$.

\textbf{Correctness proof.}
By construction it is clear that each device gets the same gradient.
Now let $g^{(s)}$ be the value of $g$ (on an arbitrary device) at the end of iteration $s$ of the outermost loop. By induction on $s$ we can show that $g^{(s)} = \frac{1}{s} \sum_{s'=1}^{s} \bar{g}_{s'}$: it is clear that $g^{(1)} = \bar{g}_1$, and
\begin{equation}
g^{(s+1)}
=
g^{(s)} + \frac{\bar{g}_{s+1} - g^{(s)}}{s+1}
=
\frac{s g^{(s)} + \bar{g}_{s+1}}{s+1}
=
\frac{1}{s+1} \sum_{s'=1}^{s+1} \bar{g}_{s'} \enspace,
\end{equation}
where the last identity follows by the inductive hypothesis.
Thus, at the end of the algorithm each device gets $g = g^{(N_{acc})} = \frac{1}{N_{acc}} \sum_{s=1}^{N_{acc}} \bar{g}_s$. Unrolling the computations done at every accumulation step on every device we get:
\begin{eqnarray*}
    g
    &=&
    \frac{1}{N_{acc}} \sum_{s=1}^{N_{acc}} \bar{g}_s
    \\
    &=&
    \frac{1}{N_{acc} \cdot N_{dev}} \sum_{s=1}^{N_{acc}} \sum_{d=1}^{N_{dev}} \hat{g}_{d,s}
    \\
    &=&
    \frac{1}{N_{acc} \cdot N_{dev}} \sum_{s=1}^{N_{acc}} \sum_{d=1}^{N_{dev}} g_{d,s} + \frac{\sigma}{N_{acc} \cdot N_{dev} \cdot B} \sum_{s=1}^{N_{acc}} \sum_{d=1}^{N_{dev}} \xi
    \\
    &=&
    \frac{1}{N_{acc} \cdot N_{dev} \cdot B_{local}} \sum_{s=1}^{N_{acc}} \sum_{d=1}^{N_{dev}} \sum_{i=1}^{B_{local}} g_{d,s,i}
    +
    \frac{\sigma}{B} \xi
    \enspace.
\end{eqnarray*}
The result now follows by observing that the first term in the sum above equals the first term in the claim's equation.

\subsection{Computational Efficiency}

We consistently find that private training requires significantly more compute than non-private training to achieve optimal performance.
The computational cost of training with DP-SGD can be broken down into two components: the cost of performing a single parameter update given a batch size, and the number of parameter updates that need to be performed for the model to reach a high accuracy.
The cost of a single DP-SGD update is largely dominated by the cost of computing per-example gradients, which is slower than computing the averaged gradient and also requires more memory.
While recent deep learning frameworks like JAX \cite{jax2018github} have significantly reduced these overheads, DP-SGD remains slower than SGD in our experience.
For example on CheXpert, we observe that the number of examples processed per second by an NFNet-F0 on a TPUv3 is about $6.5\times$ slower with our implementation of DP-SGD than with standard SGD.
Furthermore, we found that DP-SGD also required more training epochs than non-private training to achieve optimal performance.
This observation is exacerbated by the use of large batch sizes and/or large amount of (within batch) data augmentation, both of which further increase the computational cost of training.
We provide additional discussions on this in an earlier version of this work \cite{de2022unlocking}.

When tuning hyper-parameters, this cost is further added over the multiple experiments.
However, \cite{sander2022tan} demonstrate that this can be mitigated by exploring the values of hyper-parameters in a data-efficient regime and by extrapolating these values to a final expensive run that is optimized for best accuracy.

Finally, we note that the computational cost is significantly reduced on tasks where fine-tuning only the last layer is sufficient to obtain high accuracy. In our experiments however, we typically found fine-tuning the whole model to be optimal when transferring between datasets other than from JFT-300M/4B to ImageNet.

\subsection{Privacy Auditing Experiments}
\label{app:debugging:lower_bounds}

Membership inference attacks provide a powerful signal for auditing the correctness of differentially private learning algorithms \cite{jagielski2020auditing, nasr2021adversary, tramer2022debugging}.
The extent to which these attacks are successful can be used to provide (empirical) \emph{lower bounds} on the privacy guarantees afforded by an implementation, which can then be compared with the nominal upper bound.
This section reports the results of applying this methodology to our training pipeline.
Note that this type of test is incomplete; a failure to find a violation of the nominal upper bound does not rule out the possibility that one exists.
However, it provides a signal that our implementation of DP-SGD does not contain major issues, which we complement with independent code reviews and unit testing of critical components to ensure overall correctness.

\textbf{DP lower bounds via hypothesis testing.}
Given two datasets, $D$ and $D'=D\cup\{z\}$, that differ by a single data point $z$, a model trained with a DP algorithm reduces an adversary's ability to infer via a simple hypothesis test whether the model was trained on $D$ or $D'$.
More formally, if an algorithm is $(\varepsilon, \delta)$-DP then it must satisfy $\ln{(\frac{1-\beta-\delta}{\alpha})}\leq\varepsilon$
\cite{DBLP:journals/jmlr/HallRW13}, where $\alpha$ and $\beta$ denote the Type I and Type II errors of the hypothesis testing procedure.
If we can construct datasets $D$ and $D'$ such that this bound is violated, we can be sure the algorithm does not provide $(\varepsilon, \delta)$-DP;
evaluating $\ln{(\frac{1-\beta-\delta}{\alpha})}$ through a membership inference attack will give a valid lower bound of $\varepsilon$. In practice, the test relies on comparing the (distribution over) model losses of point $z$ under models trained with $D$ and $D'$.

\textbf{Overview of the attack.}
Note that DP is a worst-case guarantee, and so we are free to design our datasets $D$ and $D'$ in any way we choose. In addition, our algorithm provides the same privacy guarantees for many setting of its hyper-parameters.
Our attack operates in two phases: in phase I, we design a learning problem that we expect will maximize the ability to infer membership, while in phase II, we use the procedure set out by \cite{tramer2022debugging} to run a membership inference attack to find a (statistically valid) lower bound for $\varepsilon$. 
In phase I, we sweep over different hyper-parameter configurations and choices for $z$, to find a choice that is likely to maximize the ratio $\frac{1-\beta-\delta}{\alpha}$ when the model is trained with Algorithm~\ref{alg:dpsgd-jaxline}.
In phase II, we use the best configuration from phase I and train a large number of models to ensure the lower bound is statistically valid with enough confidence.
For completeness, we repeat the same procedure with two dataset sizes ($|D| \in \{100, 60K\}$) and four settings of the nominal privacy parameters: $\varepsilon\in\{1, 2, 4, 8\}$, where we set $\delta=\frac{1}{|D|}$.

\textbf{Experimental details.}
Throughout the auditing procedure, we use a simple LeNet classifier \cite{lecun1998gradient} as the trained model and take $D$ from the MNIST dataset.
The hyper-parameters and choices of $z$ we evaluated in phase I are given in \Cref{tab:hp_mia_config}.
Given a choice of $(\varepsilon, \delta)$, we find a choice of hyperparameters that maximizes distinguishability between models trained on $D$ and $D'$ as follows: for each hyperparameter configuration, we train 1K models on $D$ and 1K models on $D'$, where the only randomness originates from the noise added from DP-SGD. 
In particular, all models are initialized with the same parameters, as prior work has shown random initialization weakens privacy attacks \cite{jagielski2020auditing, balle2022reconstructing}. 
We then record the loss on $z$ for each model trained on $D$ and on $D'$, and fit two Gaussians over these histograms.
After this, we record the total variation distance between these two Gaussians \cite{nielsen2018guaranteed}, and choose the hyperparameter configuration that maximizes this distance.
The choice of learning rate, clipping norm, and number of model updates that maximize our ability to infer if $z$ was included in training varied for each $\varepsilon\in\{1, 2, 4, 8\}$ and $|D|\in\{100, 60K\}$. 
However, we found that selecting $z$ to be a blank image was a consistently better choice than using uniform noise or mislabeling an MNIST test set example.

In phase II, we repeat the procedure set out by \cite{tramer2022debugging} for each choice of privacy parameters, dataset size and best hyper-parameters from phase I.
For both $D$ and $D'$ we train 500K models, reserving the first 10K models on each setting as the set from which we find a threshold that maximizes the ratio between $1-\beta$ and $\alpha$, and the remaining models are used to evaluate this chosen threshold and report the $\varepsilon$ lower bound.
To gain statistical confidence in our $\varepsilon$ lower bounds we compute the lower bound of the true positive rate, $1-\beta$, and upper bound of the false positive rate, $\alpha$, over the remaining 980K models using Clopper-Pearson confidence intervals. 
With an appropriate choice of significance level this gives us a probabilistic $\varepsilon$ lower bound with 99.9\% confidence.

\textbf{Results.}
Our results are given in Table~\ref{tab:mia_results}; we did not find any violation of our reported $(\varepsilon, \delta)$-DP guarantees. 
Note that our lower bounds when using $|D|=100$ are substantially tighter than when training with $|D|=60K$ -- for example at nominal $\varepsilon>1$ training on a small dataset yields $\varepsilon$ lower bounds which are over $2\times$ stronger.
We also provide membership inference AUC and advantage ($1-\beta-\alpha$) scores for inferring if $z$ was or was not included in training over the 980K models, and give upper bounds to the membership advantage that can be derived in closed form from $\varepsilon$ and $\delta$ (cf.\ \cite{humphries2020differentially}).

\begin{table}[htpb]
\centering
\label{tab:hp_mia_config}
\resizebox{\textwidth}{!}{
\begin{tabular}{lc|c}
\toprule
Hyperparameter & \multicolumn{2}{c}{Values}                                                                          \\
\midrule

$z$       & \multicolumn{2}{c}{Uniform noise: \includegraphics[width=0.5cm]{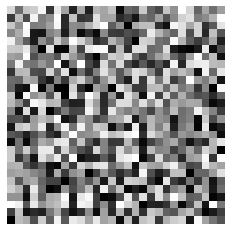}, Blank: \includegraphics[width=0.5cm]{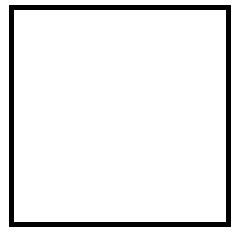}, Label 7 as 8: \includegraphics[width=0.5cm]{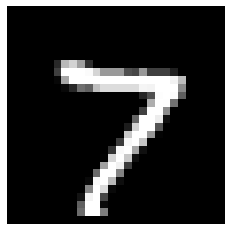},  Label 6 as 7: \includegraphics[width=0.5cm]{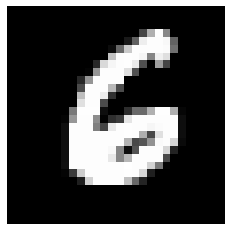}, Label 0 as 1: \includegraphics[width=0.5cm]{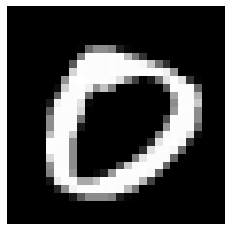}
} \\
Learning rate  & \multicolumn{2}{c}{0.1, 0.5, 1.0}                                                                   \\
Clipping norm  & \multicolumn{2}{c}{0.1, 1.0, 10.0}                                                                  \\
$|D|$ & 60K & 100 \\
Batch size     & 4096 (accumulated over two steps of batch size 2048)              & 64                              \\
Number of updates      & 500, 1K                                                           & 50, 100   \\
\bottomrule
\end{tabular}
}
\caption{\small Phase I of the $\varepsilon$ lower bound experiment: Finding the best choice of hyperparameters that distinguish models trained with and without $z$.}
\end{table}

\begin{table}[htpb]
\centering
\resizebox{\textwidth}{!}{
\begin{tabular}{ccccccc}
\toprule
$|D|$ & Nominal $\varepsilon$ & $\varepsilon$ lower bound & Membership AUC & Membership advantage & Membership advantage upper bound \\
\midrule
\multirow{4}{*}{60K}    & 1                                       & 0.279                                                                                                                               & 0.54                                        & 0.06 & 0.46                                              \\
                        & 2                                       & 0.456                                                                                                                             & 0.57                                        & 0.10    & 0.76                                           \\
                        & 4                                       & 1.139                                                                                                                               & 0.62                                        & 0.17     & 0.96                                         \\
                        & 8                                       & 2.153                                                                                                                             & 0.72                                        & 0.31          & 1.00                                    \\

\cmidrule{2-6}
\multirow{4}{*}{100}    & 1                                       & 0.361                                                                                                                             & 0.59                                        & 0.13                     & 0.47                         \\
                        & 2                                       & 0.837                                                                                                                             & 0.66                                        & 0.22     & 0.76                                         \\
                        & 4                                       & 2.461                                                                                                              & 0.79                                        & 0.43        & 0.96                                      \\
                        & 8                                       & 4.327                                                                                                                              & 0.91                                        & 0.67      & 1.00                                        \\
      
\bottomrule
\end{tabular}
}
\caption{\small Phase II of the $\varepsilon$ lower bound experiment: Reporting probabilistic $\varepsilon$ lower bounds with 99.9\% confidence and the membership inference (inferring if $z$ was or was not used in training) AUC and advantage ($1-\beta-\alpha$).}
\label{tab:mia_results}
\end{table}

\subsection{Reconstruction of CheXpert Training Data: Experimental Details}
\label{app:reconstruction_attack_details}

In \Cref{fig:main-dp-high-level} we demonstrated the one can reconstruct training data given access to gradients from SGD while the attack fails for DP-SGD.
Here we give details for how we designed the attack, when we train a small VGG-11 model~\cite{DBLP:journals/corr/SimonyanZ14a} on the CheXpert training set for simplicity.
We assume the adversary can access intermediate model updates, as permitted under the DP threat model. 
For a target training point $z$, the attacker's goal is to output a reconstruction $\hat{z}$ (that should be close to $z$ for some distance metric) given access to $g_z$, where $g_z$ is the gradient of the loss on $z$ with respect to model parameters.
Note that $g_z$ could be privatized using DP, and so the goal of this experiment is to measure the difference in reconstruction quality given access to a privatized and non-privatized gradient.
Inspired by recent work on gradient-based data reconstruction attacks~\cite{yin2021see, huang2021evaluating, jeon2021gradient, jin2021cafe, zhu2019deep,NEURIPS2020_c4ede56b} we implement an optimization-based attack by initializing $\hat{z}$ to random noise and performing gradient descent based on the loss $\lVert g_z - g_{\hat{z}}\rVert$, where $g_{\hat{z}}$ is the gradient of the loss on $\hat{z}$ with respect to model parameters. 
We optimize this objective for 100,000 iterations, for the case when the gradient is privatized and when the gradient is non-private.
As seen from \Cref{fig:main-dp-high-level}, in the non-private case the reconstruction $\hat{z}$ looks almost identical to $z$, while in the private setting, $\hat{z}$ resembles random noise even after the optimization process has terminated.
We note that there is a choice over which update step to attack in the training process. 
We chose to attack the first update, as prior work has observed that it is easier to attack updates closer to initialization~\cite{jeon2021gradient, zhu2019deep}.

\section{Related Work}

\subsection{Differentially Private Training}

\textbf{Background.}
Differential Privacy (DP) was initially formalized in \cite{dwork2006calibrating}, and was first adapted to deep learning with the work of \cite{abadi2016deep}, who showed how DP-SGD \cite{DBLP:conf/focs/BassilyST14} could be operationalized to train neural networks with differential privacy guarantees.
Since then, multiple threads of research have contributed to the ability to train machine learning models with DP, out of which three categories are relevant to situate our work. 

The first relevant thread of research aims to provide tight upper bounds on the privacy loss for DP-SGD, so that for example, as little noise as possible can be injected during training to fit a given privacy budget of $(\varepsilon, \delta)$.
In our work, we rely on \cite{DBLP:journals/popets/DoroshenkoGKKM22} to compute the $(\varepsilon, \delta)$-guarantees of our differentially private training runs.
This research workstream is complementary to our work: any improvement in that direction can be integrated to our experiments to further improve accuracy at a given privacy budget -- or conversely to reduce the privacy budget at a given level of accuracy.

Another pertinent thread has been dedicated to theoretically understanding the effects of differential privacy on the learning ability of models.
In particular, the highly influential work of \cite{DBLP:conf/focs/BassilyST14} showed that there exist worst-case convex learning problems where the statistical error of any model trained with DP can be lower bounded by a quantity that grows linearly with the number of model parameters.
This result has led many researchers to believe that DP training could not perform well in the highly dimensional regime of large-scale deep learning, and partially explains the focus of the empirical community on developing small models for DP training, as we detail further.
Further theoretical investigations have shown that by imposing some additional assumptions on the learning problem (e.g.\ generalized linear models, gradients in a constant rank subspace) it is possible to bypass these worst-case lower bounds and achieve error bounds which are independent (or grow slowly) with the number of parameters \cite{pmlr-v130-song21a,pmlr-v134-kairouz21a}.
Other relevant theoretical investigations have focused on how public data can aid private training, either by working with models pre-trained on that data (as we do in our work), or using the public data for other purposes (e.g.\ estimating subspaces on which to project gradients obtained during optimizations) \cite{NEURIPS2022_b75ce884,ganesh2023public}.

Finally, the third and most relevant research thread is an empirical approach to DP training that aims to train models that are as private and as accurate as possible.
This is the category most closely related to our work, thus we describe the prior state of the art in greater detail below.

\textbf{Image Classification from Scratch.}
As mentioned earlier, a popular belief in the research community was that smaller models would perform better than large ones when trained with differential privacy.
For instance, examining recent work on the CIFAR-10 benchmark, which has been a popular yardstick to evaluate progress of DP training in recent years,  \cite{Kurakin22}, \cite{DBLP:conf/aaai/PapernotT0CE21} and \cite{DormannFAP21}
use variants of shallow VGG models \cite{DBLP:journals/corr/SimonyanZ14a}, while \cite{DBLP:conf/iclr/TramerB21} use ScatterNets \cite{DBLP:conf/cvpr/OyallonM15} to train linear models on handcrafted features, achieving $69.3\%$ test accuracy under a tight privacy budget of $(3, 10^{-5})$-DP. 
\cite{klause2022differentially} achieved $71.7\%$ test accuracy under $(7.5, 10^{-5})$-DP by training a shallow 9-layer residual network.
Building on an earlier version of this manuscript, \cite{holzl2023equivariant} obtained 81.6\% at $\varepsilon=8$ on CIFAR-10.
In \cite{tang2023differentially}, the authors improved training accuracy by warm-starting the model parameters by pre-training on a small amount of synthetic data.
Thanks to better scaling laws, \cite{sander2022tan} were able to scale up DP training from scratch and obtain 39.2\% top-1 accuracy on ImageNet without additional data.
Finally, \cite{yu2023vip} were able to train a foundational image classifier with DP-SGD, by pre-training on synthetic data and then training the model with differential privacy on a large-scale computer vision dataset.

\textbf{Image Classification Fine-Tuning.}
In the very work introducing differentially private deep learning, \cite{abadi2016deep} already considered the task of fine-tuning with differential privacy. 
Indeed, they used a model pre-trained on the CIFAR-100 dataset without privacy, and they fine-tuned it with differential privacy on CIFAR-10.
As mentioned earlier in the manuscript, this approach assumes that the pre-training data is ``public'', and that only the fine-tuning dataset is ``private'' and should be protected by the guarantee of differential privacy.
This approach has since been used in multiple works \cite{DBLP:conf/iclr/TramerB21, YuZ0L21,cattan2022fine}, with many results focusing on small-scale datasets such as CIFAR-10 or CIFAR-100.

Recently, DP fine-tuning image classifiers at larger scale got more interest. 
In particular, DP fine-tuning on ImageNet was first tackled in \cite{Kurakin22}, where the authors obtained a top-1 accuracy of 47.8\% at a privacy budget of $(10, 10^{-6})$-DP. 
This was the only existing result on ImageNet before our earlier version of this work \cite{de2022unlocking} appeared, where we obtained a considerably improved 81.1\% top-1 accuracy under the tighter privacy budget of $(8, 8 \cdot 10^{-7})$-DP.
Shortly after our work was initially released, \cite{mehta2022large} published a study on differentially private fine-tuning on ImageNet using a large ViT model pre-trained on JFT-4B \cite{sun2017revisiting}, in which they achieved 81.1\% top-1 accuracy under (1, $10^{-6}$)-DP and 81.7$\%$ under (4, $10^{-6}$)-DP.
Subsequently, \cite{mehta2022differentially} further improved on these results by using a stronger pre-trained model and obtained 88\% under $(8, 8 \cdot 10^{-7})$-DP.
Finally, in this manuscript we obtain 88.5\% top-1 accuracy on ImageNet at the same privacy budget of $(8, 8 \cdot 10^{-7})$ with powerful NFNets pre-trained on JFT-4B.

We note however that some researchers have raised concerns about the relevance of these results for practical DP applications \cite{tramer2022considerations}, since JFT and ImageNet images come from similar data distributions (both datasets were collected from natural images publicly available on the internet).
Indeed, a crucial question is whether it is still possible to obtain strong results with DP-SGD when the pre-training data and fine-tuning data come from very different distributions. 
This situation is likely to be common in practical scenarios, since images similar in distribution to private images are often not publicly available.

Our work sets out to precisely resolve this issue, by showing that DP fine-tuning can produce highly accurate models even in the presence of a significant shift between the pre-training and the fine-tuning images.
While previous results have been shown on DP image classification on medical images \cite{ziller2021medical,pinto2023pillar,arasteh2023private,arasteh2023preserving}, to the best of our knowledge, this work is the first one to show high accuracy results (close to externally validated SOTA) on standard medical imaging datasets using only a public dataset of natural images for pre-training.

\textbf{Other Learning Tasks, Data Modalities and Privacy Threat Models.}
Differentially private training has also obtained promising results on natural language processing, with the notable successes of \cite{anil2021large} training a BERT model \cite{devlin2018bert}, and \cite{Li2021,Yu2021lm,he2022exploring} fine-tuning large-scale Transformer language models \cite{vaswani2017attention}.
Small-scale investigations into federated learning approaches for medical imaging applications have been investigated in \cite{DBLP:journals/natmi/KaissisZPRUTLMJ21}.

\subsection{Disparities in Chest X-Ray Classification}

While disparate outcomes based on demographic groups in healthcare have been highlighted for many years, the first systematic study of disparities in chest x-ray diagnosis models in terms of true positive rate (TPR) was presented in  \cite{Seyyed-Kalantari21}. They find that TPR disparities across subgroups exist in state-of-the-art classifiers for multiple datasets and clinical tasks. In followup work, they focus on under-diagnosis (e.g. a patient with a condition not receiving a diagnosis) and find that patients under 20 years old, Black patients, Hispanic patients, and patients with Medicaid insurance receive higher rates of under-diagnosis by machine learning models \cite{seyyed2021underdiagnosis}. The role of data imbalance across sex was studied in \cite{larrazabal2020gender}, where it was found that there can be AUC disparities when a minimum balance of data is not achieved. Further, \cite{zhang2022improving} analyses different techniques for improving predictive disparities and uses a comprehensive set of metrics including TPR and AUC from prior works as well as expected calibration error (ECE), cross entropy loss, recall, and specificity across groups. In the domain of private chest x-ray classifiers in particular, \cite{suriyakumar2021chasing} find that predictions made by private models are influenced by larger subgroups in the population but they do not find systematic biases in classifier performance across sex. Since disparities have been studied and well documented in chest x-ray classification, we continue this line of inquiry when building privacy-preserving chest x-ray models. 

Recent work has also highlighted that using AUC alone is not sufficient for evaluating the quality and consequently, equity, of risk assessment tools \cite{kwegyir2023misuse}. Moreover, the cost of false positives and false negatives should be weighed in a context-specific manner when choosing a metric. However, threshold based metrics such as TPR and underdiagnosis require an additional optimization step. To isolate the effect of the model training step, we will focus on measure AUC and AUC disparities in our work. 

\subsection{Privacy and Fairness}

\textbf{Privacy and fairness trade-offs.}
 The trade-offs between privacy and fairness have been studied by a number of prior works in machine learning. From differential privacy guarantees as the starting point, \cite{cummings2019compatibility} prove that exact fairness in terms of equal opportunity \cite{hardt2016equality} is incompatible with exact differential privacy for a classifier of non-trivial accuracy. From fairness guarantees as the starting point, \cite{chang2021privacy} and \cite{kulynych2022disparate} show that there are disparate privacy risks across populations in fair models using membership inference attacks. These works study models that have been trained with the objective of fair accuracy. Adjacent work has studied the notion of fair privacy; \cite{rastegarpanah2020fair} show that an optimal classifier cannot ensure individuals reveal the same amount of information (fair-privacy) while achieving fair accuracy. Nevertheless, solutions for ameliorating fairness violations such as incorporating fairness constraints \cite{tran2021differentially} and post-processing approaches \cite{mozannar2020fair} have also been proposed.
Finally, \cite{sanyal2022unfair} show that when there exists worst cases (in terms of label assignment or amount of minority subgroups) where a classifier cannot be simultaneously fair, differentially private and accurate.

\textbf{Measuring disparities in private deep learning models.}
Empirical investigations measuring the fairness of deep learning models trained with differential privacy have revealed various levels of disparity; disparity is most often measured by the difference between the worst-off group performance and the overall performance. The performance measure could be a variety of metrics including accuracy, AUC, True Positive Rate, and False Negative Rate depending on the application.  It was empirically demonstrated in \cite{bagdasaryan2019differential} that for underrepresented subgroups in age and gender classification in facial recognition datasets, sentiment analysis of tweets, and classifying species in the wild, the accuracy gap between a underrepresented class and the majority class widens with private training.  Further, in \cite{farrand2020neither} it was shown that a 70/30 imbalance in Celeb A male/female faces can cause large accuracy differences for the task of smile detection. Surprisingly, they also demonstrate that for their specific dataset and task there is a dataset split and clipping bound where a lower epsilon model has a smaller disparity between the male and female groups. Further, with unbalanced MNIST classes it was observed in \cite{uniyal2021dp} that the accuracy of the underrepresented group is much worse in a model trained with DP-SGD than trained with PATE or without privacy.
\Cref{tab:prev_work} summarises the existing literature demonstrating empirical disparities on image classification tasks.    
\begin{table}[htpb]
    \centering
    \small

    \begin{tabularx}{\textwidth}{|l|X|p{8em}|X|} \hline
        \textbf{Prior work} & \textbf{Datasets} & \textbf{Accuracy gap} & \textbf{Attributed reason for disparity} \\ \hline
        \cite{bagdasaryan2019differential}  & MNIST, Diversity in Faces, iNaturalist & ${\sim 12\%}$~(Faces),
        ${\sim 2.5\%}$~(MNIST) &  Clipping and noise addition \\ \hline
        \cite{xu2020removing} & MNIST & ${\sim 10\%}$ & Larger gradients of minority groups \\ \hline
        \cite{farrand2020neither} & Celeb A & ${\sim 20\%}$ & Class imbalance \\ \hline
        \cite{sanyal2022unfair} & CIFAR-10, Celeb A & ${\sim 7\%}$~(Celeb A), ${\sim 10\%}$~(CIFAR-10) & Long-tail data (example difficulty)\\
        \hline
    \end{tabularx}
    \caption{\small Overview of prior work reporting disparity on image classification datasets.}
    \label{tab:prev_work}
\end{table}

\section{Experimental Details}
\label{sec:methods}

\subsection{Additional Dataset Details}

ImageNet-1k is a dataset of approximately 1.3M images, each labelled with one of 1000 mutually exclusive classes \cite{ILSVRC15}.
Due to its popularity in deep learning, it is an extremely competitive benchmark that allows us to compare the accuracy of our privately fine-tuned models against the most performant published image classification models to date without differential privacy \cite{pham2021meta, DBLP:conf/icml/BrockDSS21, zhai2022scaling}.
When fine-tuning on ImageNet, we found that we could obtain performance comparable to full model fine-tuning when solely fine-tuning the final classification layer. This allows us to pre-compute the feature vector corresponding to each image, and to only learn a linear classifier layer, which significantly reduced the computational cost of training, enabling us to evaluate our performance on the larger NFNet-F7+ pre-trained model. AGC is not used when only fine-tuning the last layer. We note however that we found that it was necessary to fine-tune the full model to maximize performance on all the other datasets in this study. We speculate that this may indicate that the distribution shift between JFT and ImageNet is relatively small, when compared to the other datasets we study.

Places-365 is a dataset of approximately 1.8 million images of scenes, labelled into 365 exclusive categories.

CheXpert is a dataset of chest X-ray images annotated with labels of potential diseases \cite{IrvinRKYCCMHBSS19}.
This allows us to evaluate our fine-tuning methods in a scenario where the images protected by DP-SGD during fine-tuning are very different in nature from the images used for pre-training.
In addition, CheXpert is a public competition that has received many entries, which provides us with strong baselines for non-private training.

MIMIC-CXR is another large dataset of chest X-ray (grayscale) images annotated with labels of potential diseases, where the labels were automatically extracted from medical reports.
MIMIC-CXR serves as an additional challenging benchmark to confirm that our findings obtained on CheXpert do generalize to other datasets.

\subsection{Non-Private Pre-Training}
\label{sec:pretraining}

The pre-trained models were based on the NFNet architecture \cite{DBLP:conf/icml/BrockDSS21}, and were pre-trained without differential privacy on two data-sets, JFT-4B \cite{sun2017revisiting, zhai2022scaling} and ImageNet-21K \cite{DengDSLL009}. The choice of NFNets was motivated by their high performance on image classification while avoiding the use of batch normalization \cite{DBLP:conf/icml/IoffeS15}, which is incompatible with DP-SGD. Three models were used in the experiments: F0, F3 and F7+. F0 and and F3 are pre-activation residual networks \cite{he2016deep}, which share the same block design and have the same model width. F3 has double the depth of F1 (excluding input and output layers), which itself has double the depth of F0. F7+ has double the depth of F3, and also has slightly increased width \cite{DBLP:conf/icml/BrockDSS21}.

JFT-4B is an internal proprietary dataset, containing approximately 4 billion images, labelled into 30k (non-exclusive) classes. Networks pre-trained on JFT hold the state of the art for non-private training on a number of popular computer vision benchmarks \cite{pham2021meta, DBLP:conf/icml/BrockDSS21, zhai2022scaling, dehghani2023scaling}. To preserve privacy guarantees when fine-tuning, a script was run to remove all images from JFT that were exact or near-duplicates of images in the ImageNet and Places-365 data sets across common data augmentations. To ensure the absence of subtle duplicates, the script also removed close semantic matches by comparing image embeddings under a pre-trained model \cite{kolesnikov2020big}. The pre-training strategy follows \cite{DBLP:conf/icml/BrockDSS21} and uses an image resolution of 320x320. The NFNet-F0 and F3 were pre-trained for 2 epochs using the cross entropy loss. The larger F7+ model was pre-trained for a total of 4 epochs. For all four models, the learning rate was tuned on a logarithmic grid. No other hyper-parameters were tuned. As described in \cite{DBLP:conf/icml/BrockDSS21}, our pre-training pipeline used SGD with adaptive gradient clipping (AGC). These pre-trained models perform significantly better if the same optimization algorithm is used during fine-tuning (both with and without DP).

In order to provide results using smaller public/non-proprietary pre-training datasets, additional networks were pre-trained on the ImageNet-21k dataset, which comprises 14 million labelled images from 22k classes. ImageNet-1k (referred to as `ImageNet' in the main text) is a subset of ImageNet-21k, and consequently it cannot be used to meaningfully evaluate the performance of privately fine-tuning models pre-trained on ImageNet-21k on ImageNet-1k. Following an identical protocol to \cite{DBLP:conf/icml/BrockDSS21}, the NFNets F0 and F3 were pre-trained for 80 epochs at resolution 224x224.

\subsection{Fine-Tuning on ImageNet}

Hyper-parameters were cross-validated a validation set of 10k examples extracted from the official training set, and as is standard practice, the evaluation accuracy was reported on the official validation set of 50k examples. Images were resized to 320x320 while preserving their aspect ratio, and standardized per channel (using pre-computed average and standard deviation) before being fed to the model. On this dataset, only the last linear layer was fine-tuned.

The privacy parameter $\delta$ was set to $8 \cdot 10^{-7}$, and a clipping norm $C=1$ was employed. The batch size was set to $B = 2^{18} = 262,144$. The model was trained for 1000 updates at $\varepsilon \in \{2, 4, 8\}$ (with corresponding $\sigma \in \{14.75, 7.92, 4.38\}$ to fit the privacy budget), for 750 updates at $\varepsilon=1$ ($\sigma = 24.18$), and for 500 updates at $\varepsilon=0.5$ ($\sigma = 37.67$). More precise tuning of the step budget or noise scale did not improve performance further in our experiments. The learning rate was tuned on a logarithmic grid spaced by powers of 3.3.

\subsection{Fine-Tuning on Places-365}

All images were resized to 256x256 while preserving their aspect ratio, and they were standardized per channel (using pre-computed averages and standard deviations). All parameters of the model were fine-tuned simultaneously using DP-SGD with AGC \cite{DBLP:conf/icml/BrockDSS21}.

Results are provided when fine-tuning the NFNet-F3 pre-trained on either JFT-4B or ImageNet-21K. The privacy parameter $\delta$ was set to $5 \cdot 10^{-7}$. During training, label smoothing \cite{szegedy2016rethinking} was employed with value 0.1. Hyper-parameters were tuned on an internal validation set of 10k images extracted from the official training set. Final results were computed on the official validation set of 36.5k images. When fine-tuning the model pre-trained on JFT-4B, the batch size was set to $B=2^{17} = 131072$. The model was trained for 1000 updates at $\varepsilon \in \{4, 8\}$ ($\sigma \in \{2.70, 1.73\}$), 750 updates at $\varepsilon=2$ ($\sigma=4.72$), 500 updates at $\varepsilon=1$ ($\sigma=7.25$) and 250 updates at $\varepsilon=0.5$ ($\sigma=9.79$). When fine-tuning the model pre-trained on ImageNet-21K at $\varepsilon=8$, the batch size was set to $B=2^{17}$ and the model was trained for 1000 updates ($\sigma=1.73$). The learning rate was tuned on a logarithmic grid spaced by powers of 3.3.

\subsection{Fine-Tuning on CheXpert}
\label{sec:chexpert-details}

Following the standard methodology \cite{IrvinRKYCCMHBSS19}, multi-label binary classifiers were trained on all classes and evaluated on a subset of 5 classes: `Atelectasis', `Cardiomegaly', `Consolidation', `Edema' and `Pleural Effusion' -- the classes available in the official validation and test sets.
The Area Under the Curve (AUC) was computed per class and then averaged over classes.
For simplicity, all examples were treated independently, both at training and evaluation time, instead of aggregating them by patient or study.
Images were rescaled to have values within $[-1, 1]$.
Following \cite{IrvinRKYCCMHBSS19}, during training, uncertain labels were mapped to positive or negative labels: for classes `Atelectasis', `Edema' and `Pleural Effusion', uncertain labels were treated as positive, while for the other classes they were treated as negative.
At training time, label smoothing \cite{szegedy2016rethinking} was used with value $0.2$ for the uncertain labels (and no label smoothing for positive/negative labels).

The model was trained on all 223k training images. Hyper-parameters were cross-validated on the official validation set of 234 images, and the final performance was evaluated on the official test set of 668 images (as done in the CheXpert competition \cite{IrvinRKYCCMHBSS19}).
All layers of the model were fine-tuned, and parameters were accumulated using an Exponential Moving Average with parameter $0.9$.

For all runs with differential privacy, the batch-size was set to 4096, and the clipping-norm to $C=10^{-3}$, $\delta = 1 / N$, where $N$ is the number of training samples, and a learning-rate of $2 / C$.
The noise multiplier $\sigma$ was automatically adjusted to fit the privacy budget $(\varepsilon, \delta)$, number of steps $T$ and batch-size $B$.
For $\varepsilon=0.5$, the model was trained for $T=188$ updates ($\sigma=2.11$), for $\varepsilon=1.0$ $T=375$ ($\sigma=1.64$), for $\varepsilon=2.0$ $T=750$ ($\sigma=1.30$), for $\varepsilon=4.0$ $T=1500$ ($\sigma=1.07$) and for $\varepsilon=8.0$ $T=3000$ ($\sigma=0.91$).

For the non-private baseline, the model was trained for $T=2000$ updates at a batch-size of 1024 (training longer resulted in over-fitting), using a constant learning-rate of 1.0 and a weight decay of 0.0001.

For both private and non-private runs, the final predictions were aggregated from multiple checkpoints stored along the course of a single training run, following \cite{IrvinRKYCCMHBSS19}. 
For each run, ten evenly spaced checkpoints were created, each using the Exponential Moving Average version of the parameters.
Neither the EMA or the checkpoint aggregation incur a cost in terms of differential privacy since DP-SGD allows for the release of all intermediate checkpoints.

The experiments on DP fine-tuning take from 3 TPUv3 hours ($\varepsilon=0.5$) to 42 TPUv3 hours ($\varepsilon=8$), while the non-private fine-tuning experiment takes approximately 1 TPUv3 hour.

\subsection{Fine-Tuning on MIMIC-CXR}
\label{sec:details-mimic}

All classes were used for training and evaluation on MIMIC-CXR, following standard protocol \cite{Seyyed-Kalantari21}.
As in CheXpert, Area Under the Curve (AUC) was computed per class and averaged over the classes.
In order to compare with the results from \cite{Seyyed-Kalantari21,KamalZNH22}, an identical protocol was used to extract validation and test sets from the official training set, forming a 80-10-10 split for training-validation-testing (respectively 259k, 32k and 32k examples).
For simplicity, all examples were treated independently (both at training and evaluation time), instead of aggregating them by patient or study.
The images were down-sampled to 256x256 and rescaled to have values within $[-1, 1]$.
All layers of the model were fine-tuned in this experiment, and the parameters were accumulated using an Exponential Moving Average with parameter 0.9.

For all runs with differential privacy, the batch-size was set to 4096, the clipping-norm to $C=10^{-3}$, $\delta = 1 / N$, where $N$ is the number of training samples, and a learning-rate of $2 / C$.
The noise multiplier $\sigma$ was automatically adjusted to fit the privacy budget $(\varepsilon, \delta)$, number of steps $T$ and batch-size $B$.
At $\varepsilon=0.5$, the model was trained for $T=188$ updates ($\sigma=1.88$), at $\varepsilon=1.0$ for $T=375$ ($\sigma=1.48$), at $\varepsilon=2.0$ for $T=750$ ($\sigma=1.19$), at $\varepsilon=4.0$ for $T=1500$ ($\sigma=0.99$) and at $\varepsilon=8.0$ for $T=3000$ ($\sigma=0.85$).

For the non-private baseline, the model was trained for $T=2000$ updates at a batch-size of 1024  (training longer resulted in over-fitting), using a constant learning-rate of 0.25 and a weight decay of 0.0001.

The experiments on DP fine-tuning take from 2.4 TPUv3 hours ($\varepsilon=0.5$) to 37 TPUv3 hours ($\varepsilon=8$), while the non-private fine-tuning experiment takes approximately 1 TPUv3 hour.

\subsection{Analyzing Fairness Disparities in CXR Classification Models}
\label{sec:details-fairness}

Fairness is analyzed through the lens of accuracy parity \cite{barocas-hardt-narayanan} by measuring disparities through additive differences in model performance on subgroups relative to the overall accuracy as in \cite{Seyyed-Kalantari21}. Area under the ROC curve (AUC) was used as the measure of accuracy for CXR models.
For an evaluation dataset $D$, a subgroup $A$ and a classifier $f$, the disparity in AUC of $f$ on $A$ is given by
$
Disp_A = AUC(f, D) - AUC(f, D_A),
$
where $D_A$ is the dataset obtained by taking all the records in $D$ corresponding to individuals in group $A$.

\textbf{Disparities on MIMIC-CXR.}
Following \cite{seyyed2021underdiagnosis}, sub-groups were defined using demographic groups present in the dataset based on patient's sex (M/F), age bracket (age discretized to one of 18-20, 20-40, 40-60, 60-80, 80+), race (american indian/alaska native, asian, black/african american, hispanic/latino, white, and other), and type of insurance (medicaid, medicare, other). In addition to sub-groups defined by single attributes, intersectional sub-groups based on (sex, race) and (sex, age bracket) were also considered.
Demographic attributes for each individual were obtained by linking the patient ID in the MIMIC-CXR dataset \cite{mimic-cxr-data} with the admissions table in the MIMIC-IV dataset \cite{mimic-demo}.
In a small number of cases, patients had multiple admission records reporting different races or types of insurance -- records from these patients were removed from the dataset before analysis to avoid confounders.

\textbf{Disparities on CheXpert.}
A similar procedure to MIMIC-CXR was followed, although some modifications were required.
First, the private models and non-private baselines were re-trained following the procedure described in Section~\ref{sec:chexpert-details} but using a different dataset split: the official training dataset was partitioned into 3 parts using an 80-10-10 proportion for training, validation and testing, resulting in datasets of size 178k, 22k and 22k respectively. 
This was necessary because the official test set does not contain demographic attributes (while the official training set does). 
The splits were created by assigning the first 10\% of patient IDs to the test set, the following 10\% to the validation set and the remaining 80\% to the training set.
This process did not bias the data distribution of each split because patient IDs were randomly generated when the CheXpert dataset was originally compiled, and at the same time ensures that images from a single patient only occur in one of the splits.
When training private models on this split, the batch size and number of iterations remained identical to the experiments from Section~\ref{sec:chexpert-details}, but the noise multiplier was adapted to fit the privacy budget $\varepsilon$ given the new number of training examples as follows:
$\sigma = 2.52$ for $\varepsilon=0.5$,
$\sigma = 1.94$ for $\varepsilon=1$,
$\sigma = 1.52$ for $\varepsilon=2$,
$\sigma = 1.23$ for $\varepsilon=4$,
and
$\sigma = 1.03$ for $\varepsilon=8$.

Demographic sub-groups in CheXpert were based on the only two demographic attributes present: age range (discretized as above), and sex.

\subsection{Tuning the Hyper-Parameters}

In all our experiments when fine-tuning with DP-SGD, we use a constant learning rate throughout training. Although learning rate schedules significantly improve performance in non-private training \cite{loshchilov2016sgdr, you2019does}, we found that they did not improve performance for private training and required additional hyper-parameter tuning. We also use an Exponential Moving Averaging (EMA) of the model parameters with rate 0.9999 (unless otherwise stated), using the EMA warm-up scheme described in \cite{tan2019efficientnet}. We found that EMA consistently improved model accuracy by increasing the convergence rate when training with the added noise required to achieve privacy guarantees with DP-SGD. 

We use extremely large batch sizes, often only an order of magnitude smaller than the total dataset size. As also observed by other researchers \cite{DBLP:journals/corr/abs-2007-05089, LuoW0021,Kurakin22,yu2021large,Li2021,anil2021large,DormannFAP21}, we found that this significantly improved performance when training with DP-SGD. When we use large batch-sizes, we use gradient accumulation across multiple steps to avoid memory overflow. We note that our DP-SGD implementation encodes this feature carefully so as to ensure that the correct amount of noise is injected to satisfy the DP guarantees. 

We found that the performance of DP-SGD did not depend strongly on the clipping norm $C$ (so long as the clipping norm was not too large). We therefore set this hyper-parameter $C=1$ for all experiments on the ImageNet and Places-365 sets. When fine-tuning on the CheXpert and MIMIC-CXR datasets, we employ a per-class loss function that results in gradients with smaller scale, thus we use $C=0.001$ in these cases. 

The main hyper-parameter specific to DP-SGD that we needed to tune carefully on some (though not all) datasets was the noise multiplier $\sigma$. We found that performance usually improves as $\sigma$ rises, however this also increases the computational cost of training, since the number of training iterations allowed within a fixed privacy budget $\varepsilon$ increases as $\sigma$ rises. We found that values of $\sigma$ close to 2 usually achieve a good trade-off between achieving high accuracy while not requiring too large a computational budget \cite{sander2022tan}.

Data augmentation as conventionally applied, drawing a single random sample from the augmentation procedure for each image in the current batch \cite{lecun1998gradient, shorten2019survey}, consistently reduced the performance of private training with DP-SGD. In order for private training to benefit from data augmentation, we found in \cite{de2022unlocking} that one must apply augmentation multiplicity \cite{hoffer19, Fort2021}, whereby the per-example gradient of each image in the batch is averaged over multiple independent random augmentations. This averaging can be applied before the per-example gradients are clipped, reducing the privacy loss. However the downside of this scheme is that it increases the computational cost of training with DP-SGD substantially. For the experiments provided in the main text we instead simply removed data augmentation from the training pipeline.

As stated above, we found performance improved when the same optimization algorithm used for pre-training was also used during fine-tuning. We therefore applied DP-SGD with AGC \cite{DBLP:conf/icml/BrockDSS21}, to match our pre-training framework described in the \refsecbyname{sec:methods} section.

On ImageNet, only the last linear layer was fine-tuned with differential privacy. On all other datasets, all layers were fine-tuned, which we find to significantly outperform tuning the last layer only. We postulate that this is due to the wider significant shift between the pre-training data and Places-365 as well as the medical image classification tasks, compared to that between the pre-training data and ImageNet.

\section{Disparities in Private Chest X-Ray Classification: Additional Results}

\subsection{MIMIC-CXR}
\begin{figure}[htpb]
    \centering
    \includegraphics{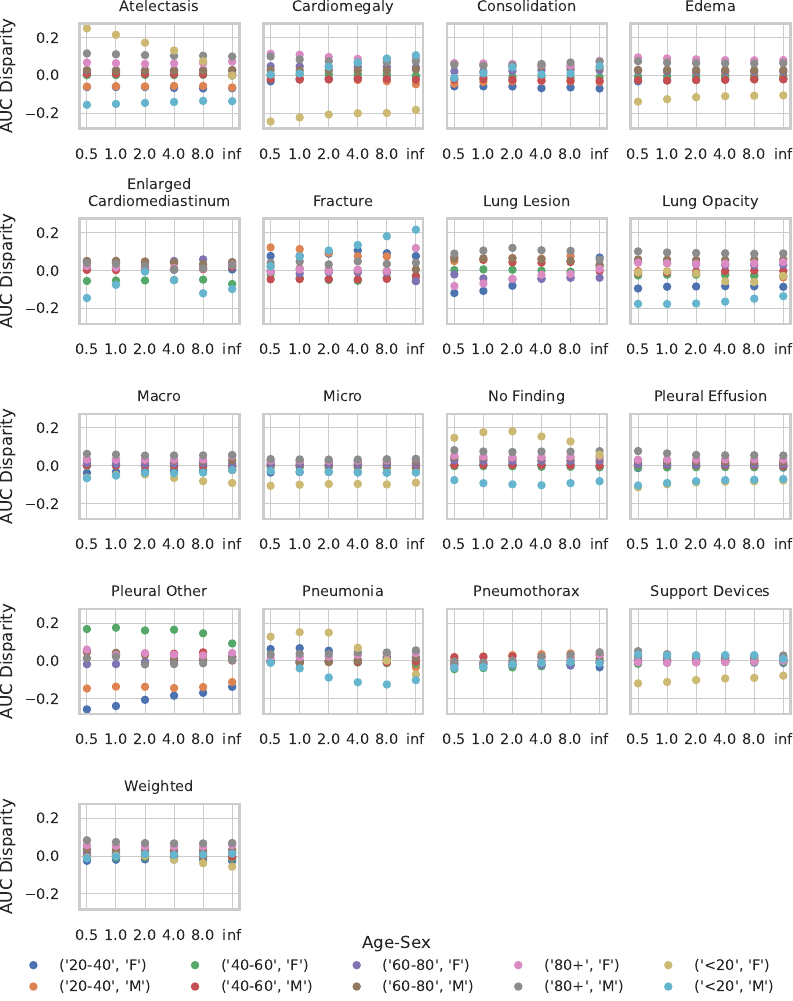}
    \caption{\small
        AUC disparities on MIMIC-CXR for the various age-sex subgroups, as a function of $\varepsilon$. 
        AUC disparities do not consistently increase as the model gets more private (i.e. as $\varepsilon$ decreases).
    }
    \label{fig:cxr_disp_all_mimic_0}
\end{figure}

\begin{figure}[htpb]
    \centering
    \includegraphics{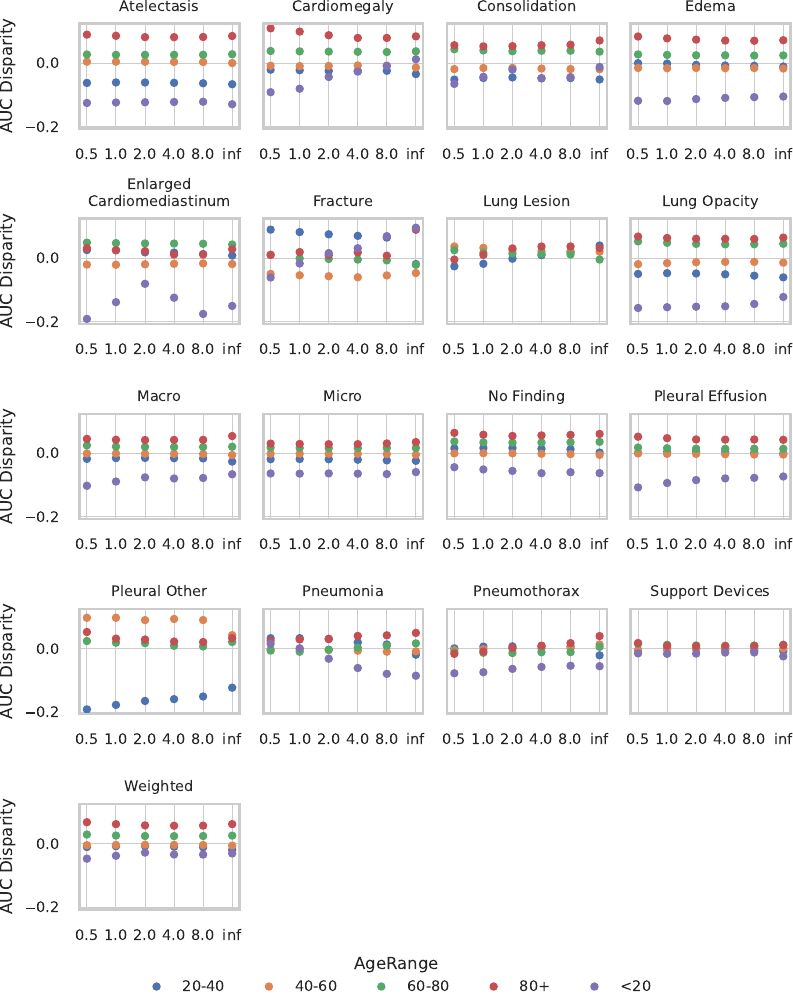}
    \caption{\small
        AUC disparities on MIMIC-CXR for the various age subgroups, as a function of $\varepsilon$. 
    }
    \label{fig:cxr_disp_all_mimic_1}
\end{figure}

\begin{figure}[htpb]
    \centering
    \includegraphics{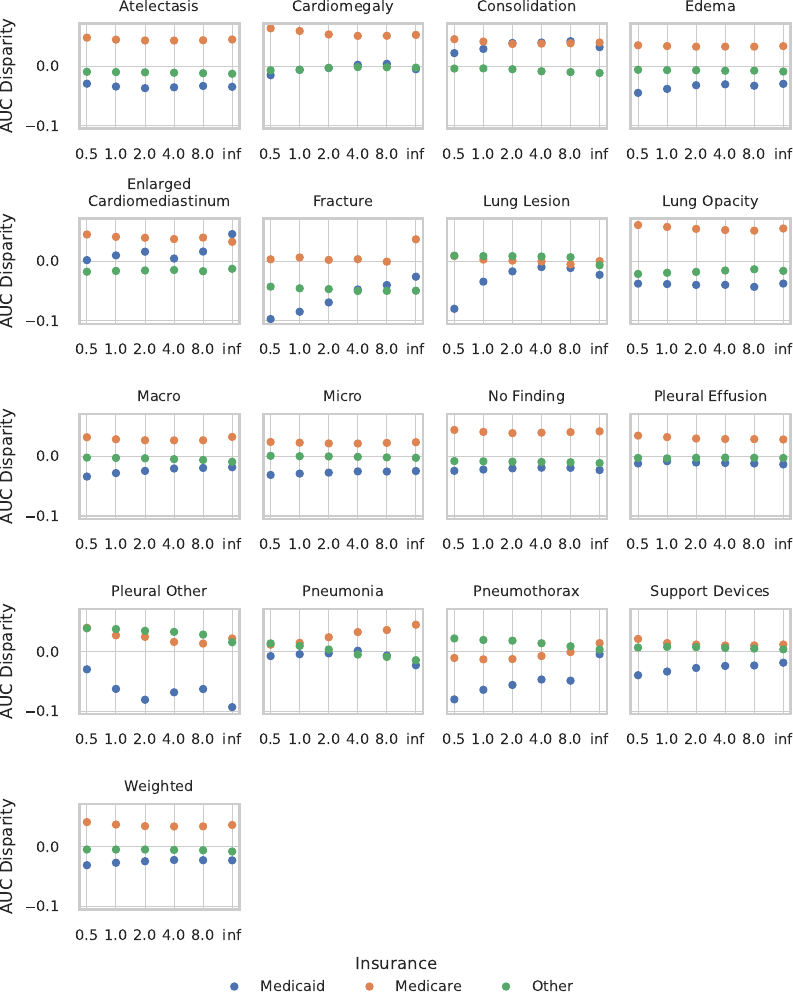}
    \caption{\small
        AUC disparities on MIMIC-CXR for the various insurance subgroups, as a function of $\varepsilon$. 
    }
    \label{fig:cxr_disp_all_mimic_2}
\end{figure}

\begin{figure}[htpb]
    \centering
    \includegraphics{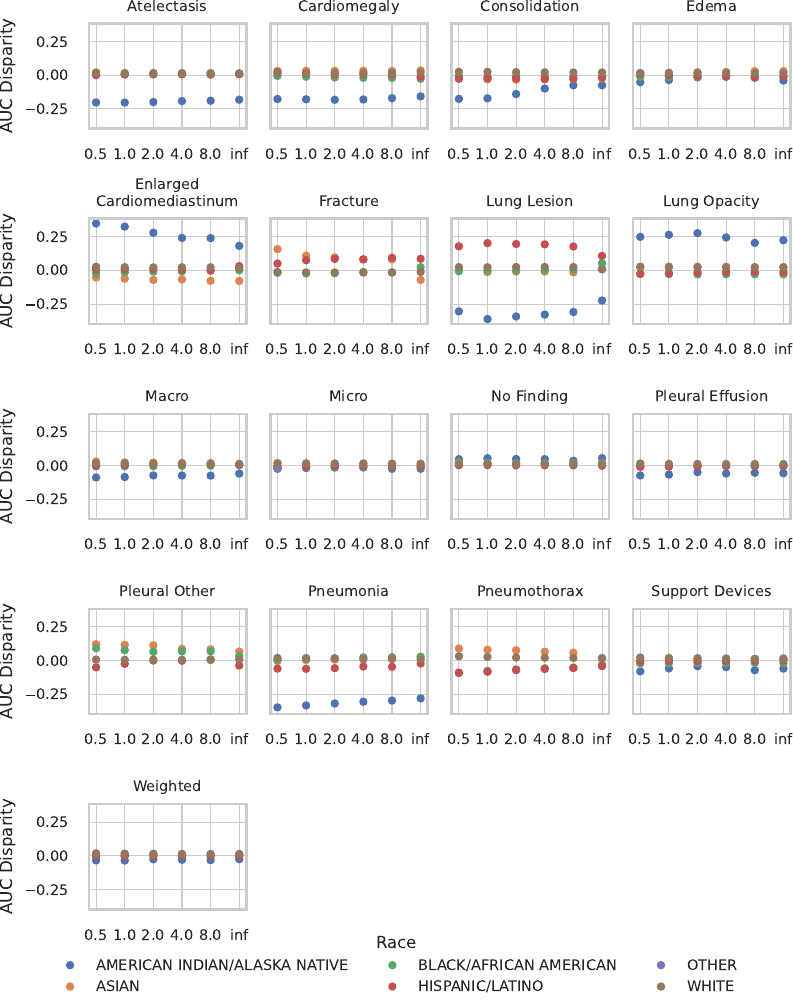}
    \caption{\small
        AUC disparities on MIMIC-CXR for the various race subgroups, as a function of $\varepsilon$. 
    }
    \label{fig:cxr_disp_all_mimic_3}
\end{figure}

\begin{figure}[htpb]
    \centering
    \includegraphics{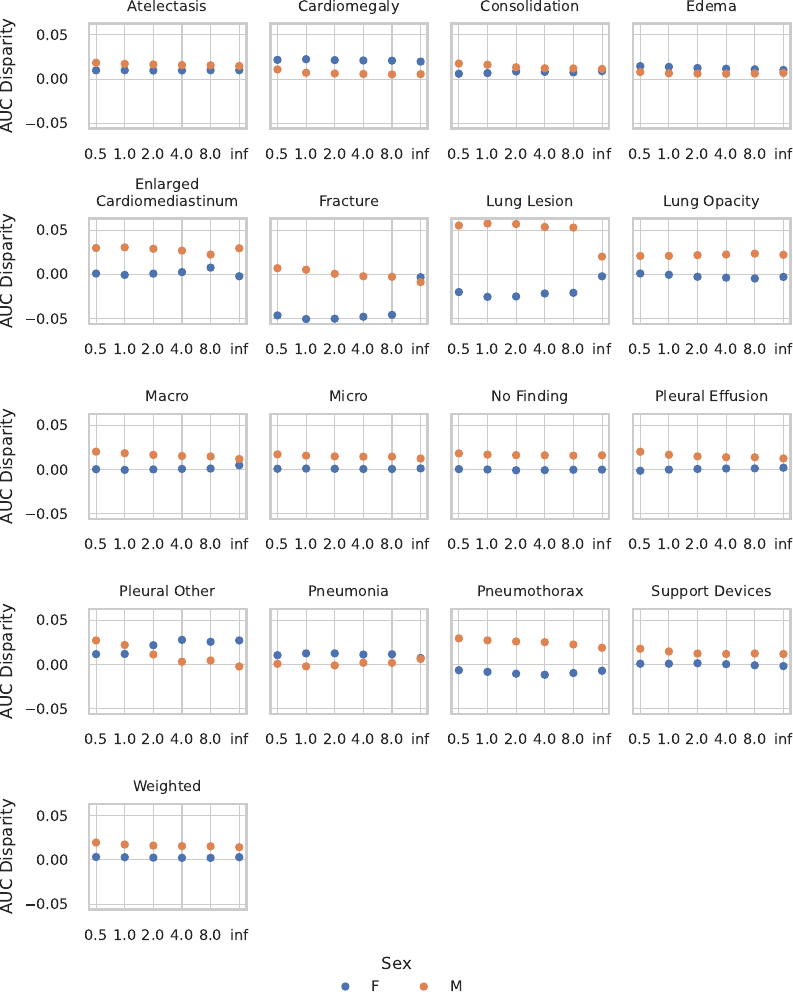}
    \caption{\small
        AUC disparities on MIMIC-CXR for the various sex subgroups, as a function of $\varepsilon$. 
    }
    \label{fig:cxr_disp_all_mimic_4}
\end{figure}

\begin{figure}[htpb]
    \centering
    \includegraphics{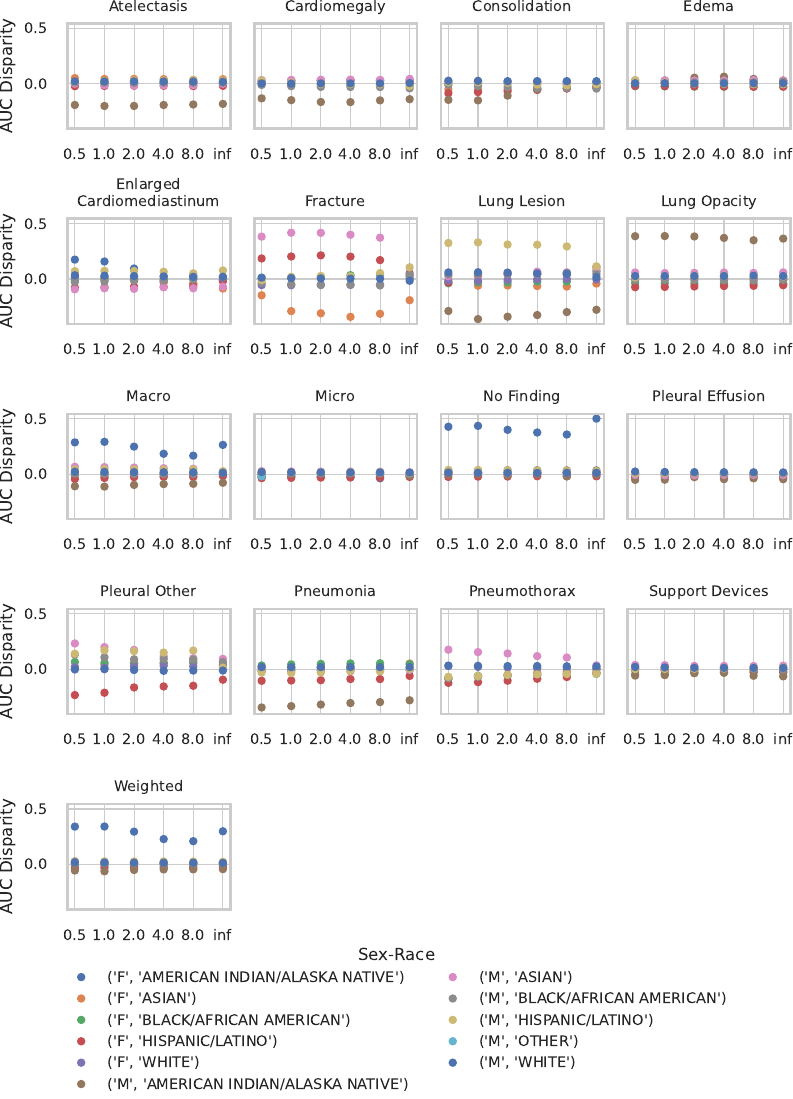}
    \caption{\small
        AUC disparities on MIMIC-CXR for the various race-sex subgroups, as a function of $\varepsilon$. 
    }
    \label{fig:cxr_disp_all_mimic_5}
\end{figure}

\begin{figure}[htpb]
    \centering
    \includegraphics{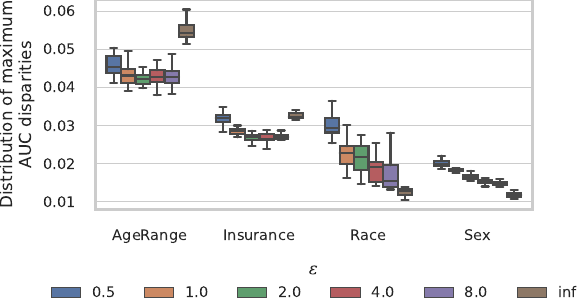}
    \caption{\small Maximum AUC disparities across sub-groups of different types (age, insurance, race and sex) as a function of $\varepsilon$ on MIMIC-CXR. The distribution shown is over 20 random seeds for the model training procedure. The maximum disparity reduces as $\varepsilon$ grows from $0.5$ to $8$ -- for public models (i.e.\ $\varepsilon = \infty$) disparities are higher than in private models for age and insurance groups, and smaller for race and sex groups.}
    \label{fig:cxr_distr_disp_mimic}
\end{figure}

To measure the fairness effects of DP training on the MIMIC-CXR dataset, we analyze the differences in AUC disparities between the private models and non-private baselines (cf.\ Section~\ref{sec:details-mimic}).
In \Cref{fig:main-analysis} we present the results of this evaluation for private models with $\varepsilon = 8$ compared to non-private models using the average AUC across all 14 labels present in the dataset. \Cref{fig:cxr_disp_all_mimic_0}-\ref{fig:cxr_disp_all_mimic_5} present a more fine-grained view of these results using AUC across individual labels and a range $\varepsilon$'s. To understand how the level of privacy guarantee (i.e.~$\varepsilon$), \Cref{fig:cxr_distr_disp_mimic} provides a view on the distribution (over random seeds) of the maximum AUC disparity across sub-groups defined by single demographic attributes. 

\subsection{CheXpert}
\begin{figure}[htpb]
    \centering
    \includegraphics{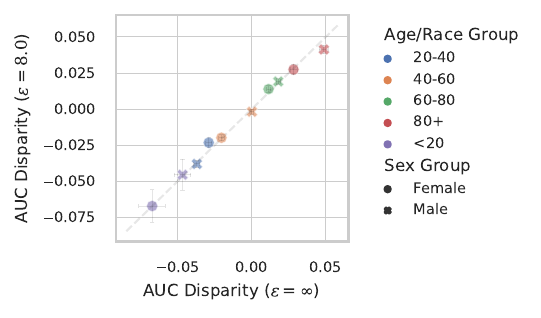}
    \includegraphics{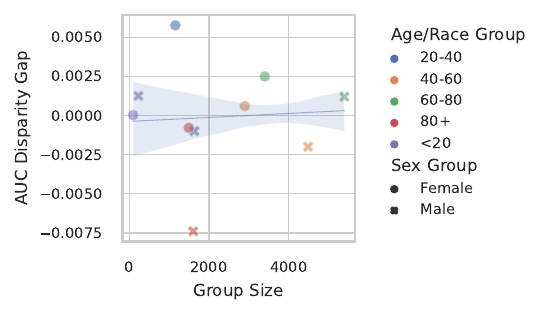}
    \caption{\small AUC disparities (i.e. population AUC - sub-group AUC) in private models are comparable to the disparities we observe on non-private models with comparable overall accuracy on CheXpert. (Top) For the private ($\varepsilon=8$) and non-private baseline models from Figure 3, comparing AUC disparities by sub-group between both models. Disparities are averaged over 20 independent runs, and gray crosses represent standard deviation. (Bottom) Stratification of differences in disparities between private and non-private models (averaged over 20 independent runs) by sub-group size. We include an OLS regression line to predict disparity gap as a function of group size (and 95\% confidence interval based on 1000 bootstrap resamples). The slope of the regression model lies in the 95\% confidence interval [-6.63e-07, 6.37e-07].}
    \label{fig:chexpert_basic}
\end{figure}

\Cref{fig:chexpert_basic} presents the results of evaluating disparities between the private models and non-private baseline using the same procedure as in \Cref{fig:main-analysis}. We note that CheXpert only contains sex and age demographic attributes, so the evaluation is limited to intersectional groups based on these two attributes. Qualitatively we observe the same behavior as in the MIMIC-CXR dataset.

\begin{figure}[htpb]
    \centering
    \includegraphics{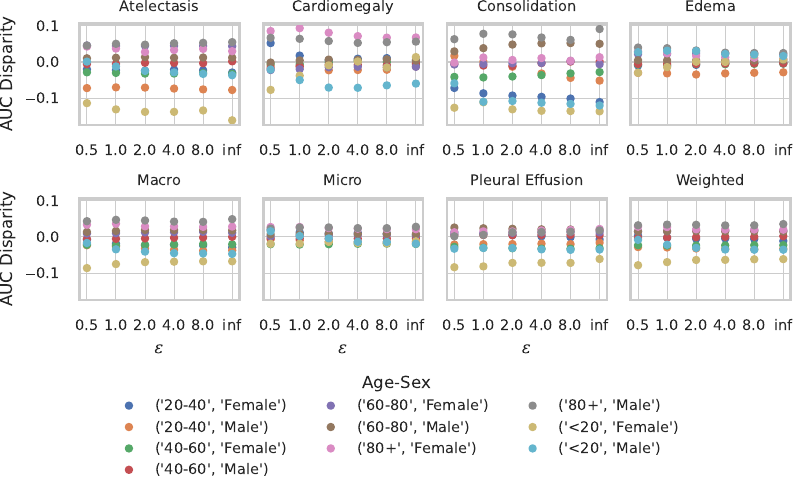}
    \caption{\small
        AUC disparities on CheXpert for the various age-sex subgroups, as a function of $\varepsilon$. 
    }
    \label{fig:cxr_disp_all_chexpert_0}
\end{figure}

\begin{figure}[htpb]
    \centering
    \includegraphics{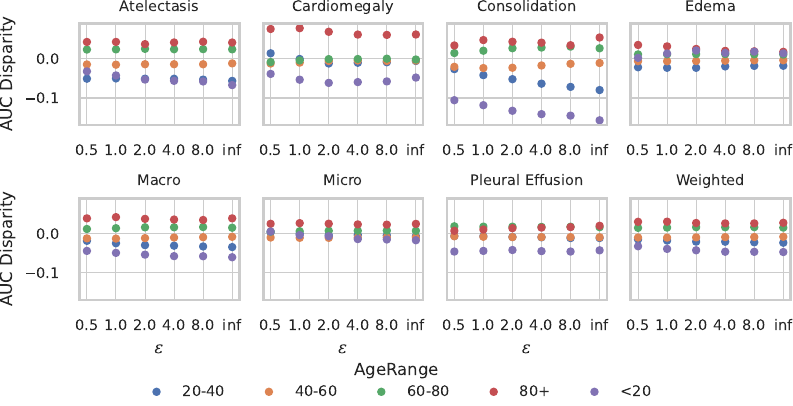}
    \caption{\small
        AUC disparities on CheXpert for the various age subgroups, as a function of $\varepsilon$. 
    }
    \label{fig:cxr_disp_all_chexpert_1}
\end{figure}

\begin{figure}[htpb]
    \centering
    \includegraphics{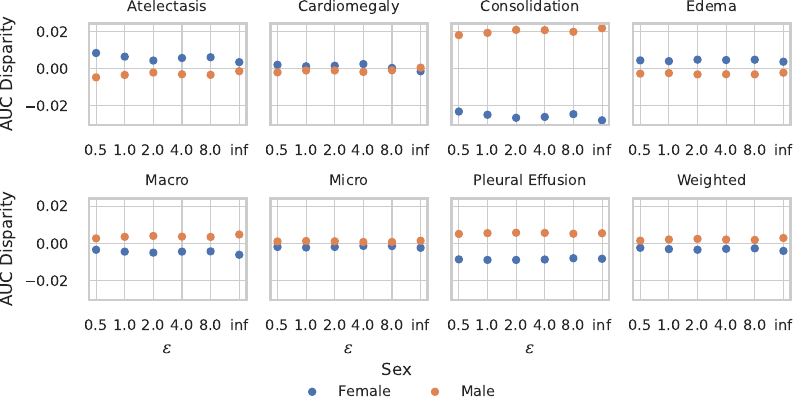}
    \caption{\small
        AUC disparities on CheXpert for the various sex subgroups, as a function of $\varepsilon$. 
    }
    \label{fig:cxr_disp_all_chexpert_2}
\end{figure}

\begin{figure}[htpb]
    \centering
    \includegraphics{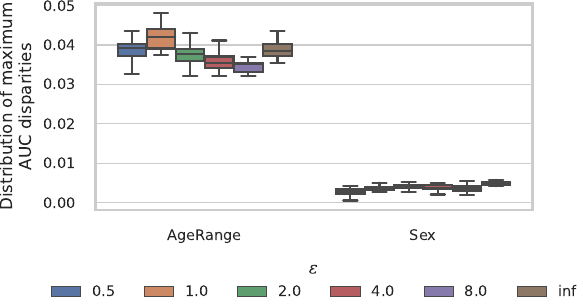}
    \caption{\small Maximum AUC disparities across sub-groups of different types (age, insurance, race and sex) as a function of $\varepsilon$ on CheXpert. The distribution shown is over 20 random seeds for the model training procedure. The maximum disparity reduces or stays roughly the same as $\varepsilon$ grows from $0.5$ to $8$ -- for public models (i.e.\ $\varepsilon = \infty$) disparities are higher than in private models.}
    \label{fig:cxr_distr_disp_chexpert}
\end{figure}

\section{Additional Results on Training From Scratch on CIFAR-10}

\subsection{Per-Class Disparities}

While the prior sections focused on measuring disparities in models fine-tuned with differential privacy, this section will be dedicated to understand disparities in privacy-preserving models trained from scratch. We build on prior work examining private model disparities at lower overall accuracy by extending analysis to higher model accuracy regimes. Our results also complement theoretical impossibility results about the limits of private learning. We look at a practical regime of model performance that does not yet require memorisation of individual examples \cite{DBLP:conf/stoc/Feldman20}. The significance of our findings in this section lies in the re-framing of privacy-fairness trade-offs into privacy-accuracy trade-offs. When private model accuracy is much worse than non-private models, closing the accuracy gap will also lead to narrowing the disparity gap.

\textbf{Experimental setup.}
We train a Wide Residual Network (WRN) on CIFAR-10, where we use 45k examples for training, 5k for validation and 10k for testing. The input images are standardized per channel using pre-computed average and standard deviation per channel, aggregated over the dataset. For our two main models for comparison: we use the same architecture to train a non-private model to achieve an overall 93\% Top 1 accuracy\footnote{While some non-private models can achieve up to 99.9\% accuracy on CIFAR-10, our accuracy is a reasonable comparison to models without pre-training, architecture search, ensembling, or attention  \url{https://benchmarks.ai/cifar-10}} and for a private model $\varepsilon=8$ with DP-SGD to an overall 81\% accuracy following techniques from \cite{de2022unlocking}. We describe specific procedures for different experiments in the subsequent sections. 

\textbf{Defining subgroups.}
When examining accuracy disparities across classes for our CIFAR-10 dataset, we measure the difference in accuracy of one class compared to another; the subgroups of interest are defined by the true class label. We note that notions of fairness such as demographic parity (independence) has no meaning when subgroups are defined by the true label. However, we study this type of disparity across classes following a series of prior work measuring class disparities  \cite{bagdasaryan2019differential,xu2020removing}. This notion of disparity has also been used in adjacent areas such as robustness \cite{benz2021robustness, xu2021robust}. Examining disparities between classes gives more insight beyond just accuracy in terms of understanding the strengths and weaknesses of a model. However, since class labels are not independent, a high accuracy for one class may arise from a model predicting too many examples to be that class and consequently reducing the accuracy of another class. 

\textbf{Better measurement of private model disparities.}
One question that naturally arises is whether the per-class accuracy differences between models can be replicated across different random seeds. We would require disparity measurement to be stable across random seeds before concluding that private models have more disparity or a different type of disparity. To answer this question, we trained 50 different models at each $\varepsilon$ over different random seeds and plot the spread of class conditional accuracy observed. 

\begin{figure}[htpb]
  \subfloat{
	\begin{minipage}[c]{
	   0.48\textwidth}
	   \centering
	   \includegraphics[width=1\textwidth]{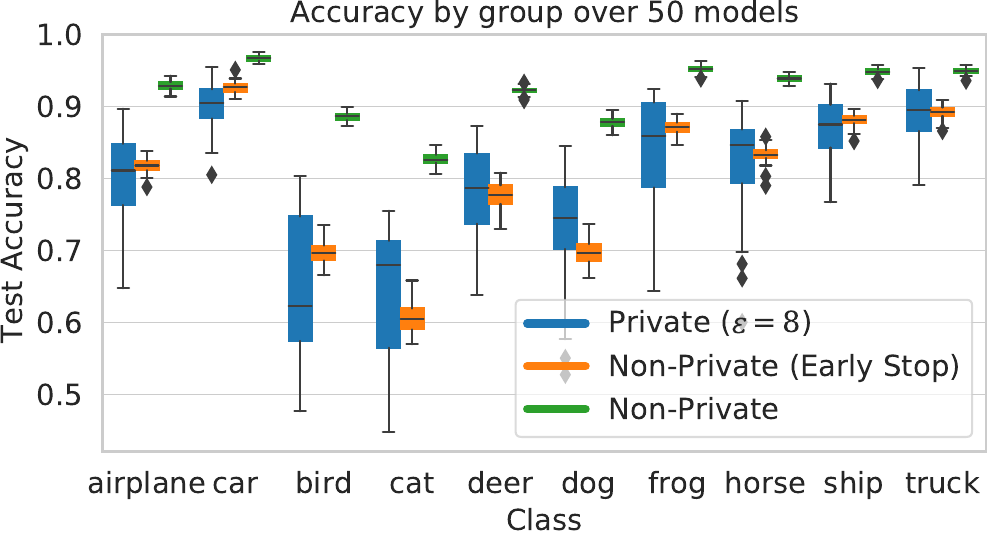}
	\end{minipage}}
 \hfill 	
  \subfloat{
	\begin{minipage}[c]{
	   0.48\textwidth}
	   \centering
	   \includegraphics[width=1\textwidth]{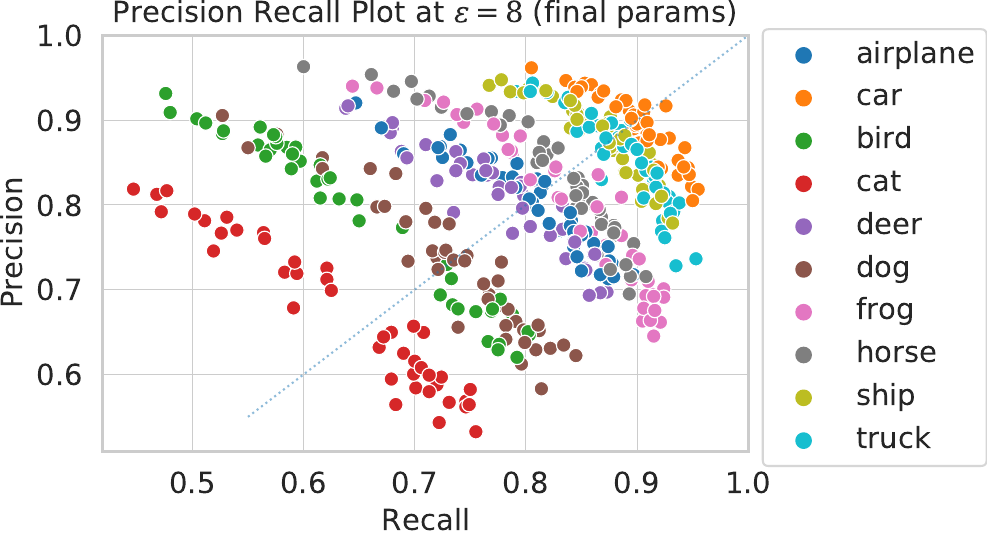}
	\end{minipage}}
\caption{\small (Left) Box plot of class conditional accuracy shows significant variation across random seeds. (Right) Class-wise precision vs recall plot across random seeds. The variation in class conditional accuracy is due to private models over and under predicting certain classes.}
\label{fig:noisy_per_class}
\end{figure} 

While all non-private and private models were within the same overall accuracy range ($\varepsilon=1$: 50\% - 55\% accuracy, $\varepsilon=8$: 77\% - 81\% accuracy), \Cref{fig:noisy_per_class} (left) shows how much class conditional disparity can vary from run to run across different random seeds. There is enough variation that it is not clear what the worst performing class is. Furthermore, when we stop training of a non-private model early to be similar in accuracy to the $\varepsilon=8$ model, the non-private models exhibit substantially less variation across random seeds than the private models. 

To explain this phenomenon, we can look at the precision-recall plot. In \Cref{fig:noisy_per_class} (right), there is a large spread between where each class lies in the precision-recall plot across random seeds. For example, in some runs, the \textit{cat} class has precision $\approx0.82$ and recall $\approx0.45$ recall which means that most images that are predicted to be cats are indeed cats but there are also many cat images that are predicted to be another class. High precision and low recall causes under-prediction of a class while low precision and high recall causes over-prediction of a class. Over-predicting one class implies that the model is also under-predicting another class; for one model with high precision on the \textit{cat} class would then necessarily have low precision on the \textit{dog} class since the two classes of images are often confused for one another. In \Cref{fig:noisy_per_class} (right), each model is illustrated by a set of 10 dependent plotted points, one for each class.

This over- and under- prediction is not simply an artifact of a less accurate model. For an early-stopped non-private model with the same accuracy as the $\varepsilon=8$ models, the precision-recall plot shows all classes to be on or near the diagonal across all runs. Class conditional accuracy of the last checkpoint in a model trained with DP-SGD does not provide stable estimates of disparity for multi-class classification. When subgroups defined by classes are not independent (e.g. CIFAR-10 / MNIST), the additional noise introduced by DP-SGD may cause over and under prediction of classes across runs.

\begin{figure}[htpb]
  \subfloat{
	\begin{minipage}[c]{
	   0.48\textwidth}
	   \centering
	   \includegraphics[width=1\textwidth]{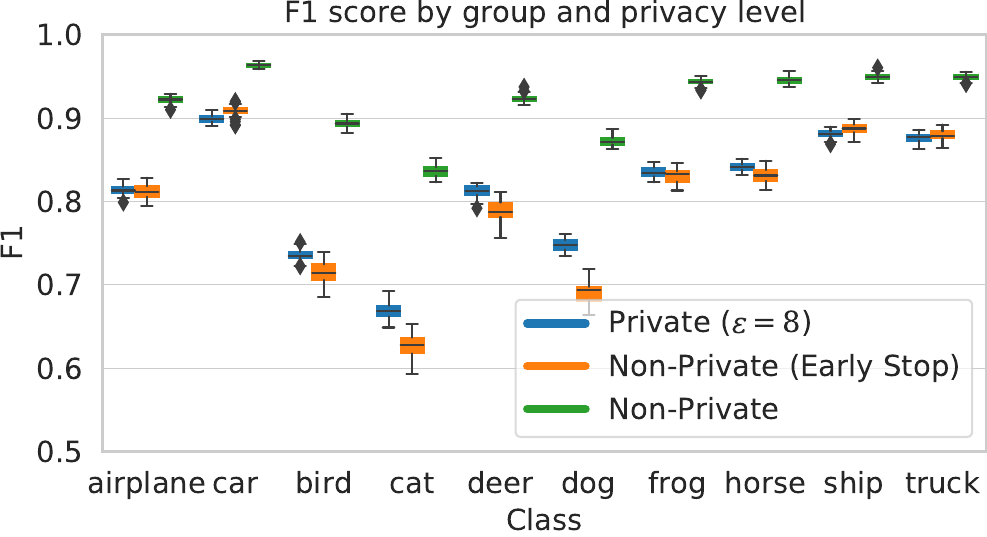}
	\end{minipage}}
 \hfill 	
  \subfloat{
	\begin{minipage}[c]{
	   0.48\textwidth}
	   \centering
	   \includegraphics[width=1\textwidth]{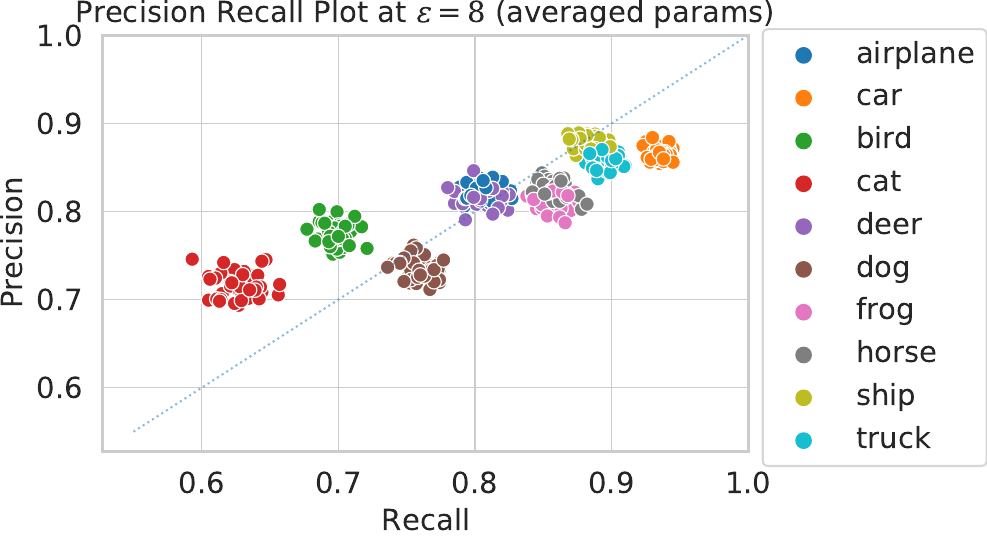}
	\end{minipage}}
\caption{\small (Left) Using F1 score instead of accuracy drastically reduces variation in per class performance for private models. (Right) Using checkpoint averaging like Exponential Moving Average (EMA) also reduces variation.}
\label{fig:variance_corrected}
\end{figure}

We observe that the reduction in the variance of accuracy under private training when using EMA directly mitigates the noisiness in these measurements, thus producing more robust disparity measurements because significant fluctuations as a function of the random seed are removed (\Cref{fig:variance_corrected} - right). Alternatively, we observe that using F1 score as a measure of accuracy (without EMA) also helps reduce the variations of accuracy measurements across random seeds (\Cref{fig:variance_corrected} - left).

\begin{figure}[htpb]
  \subfloat{
	\begin{minipage}[c]{
	   0.48\textwidth}
	   \centering
	   \includegraphics[width=\columnwidth]{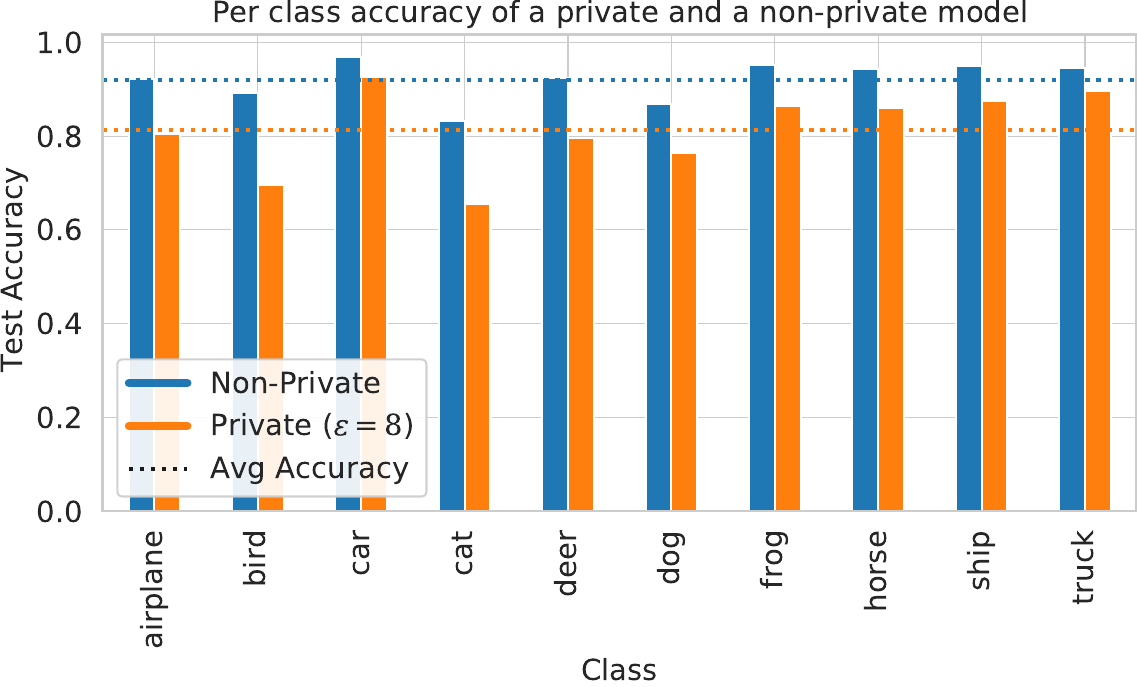}
	\end{minipage}}
 \hfill 	
  \subfloat{
	\begin{minipage}[c]{
	   0.48\textwidth}
	   \centering
    \includegraphics[width=\textwidth]{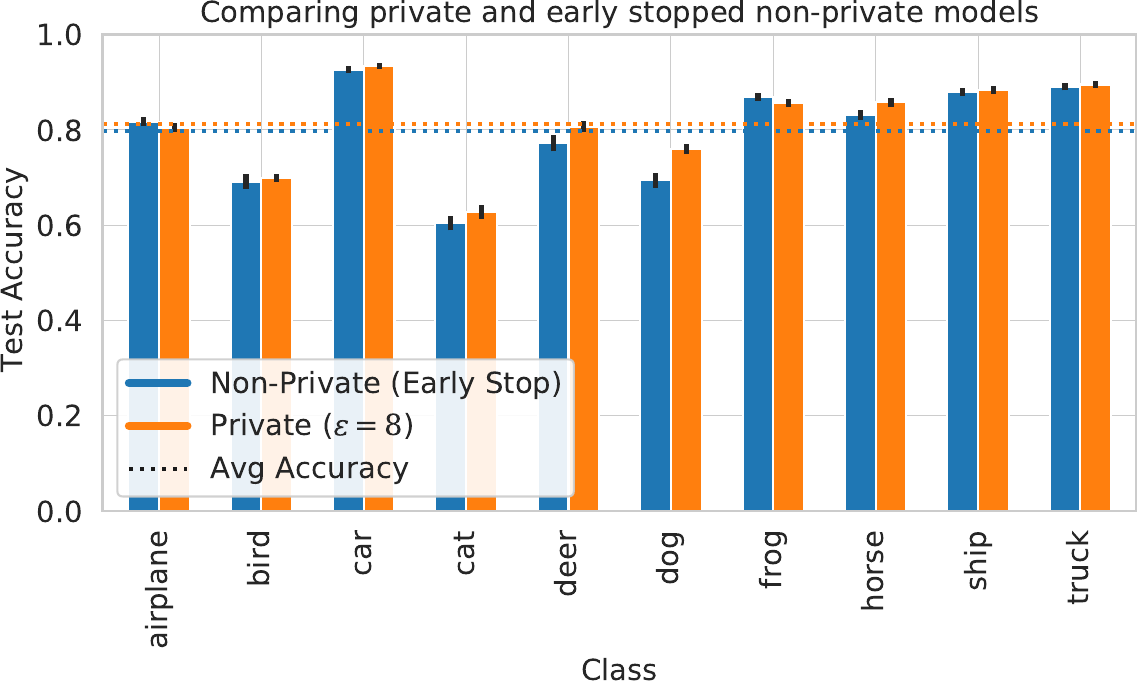}
	\end{minipage}}
\caption{\small (Left) Class conditional accuracy of a non-private model (93\% Top 1 accuracy) and a $\varepsilon=8$ private model (81\% Top 1 accuracy) trained on CIFAR-10 (balanced classes). Dotted lines represent the overall Top 1 accuracy. (Right) Comparison of $\varepsilon=8$ model with non-private models at a similar accuracy (EMA). We see that the per class disparities are similar.}
\label{fig:pub_priv_compare}
\end{figure}

\textbf{Results: Balanced Classes.} To disentangle the role of class imbalance and lower overall accuracy, we first compare private and non-private models trained with balanced data. When comparing a $\varepsilon=8$ model to a non-private model that is $\sim12\%$ more accurate, we observe an absolute disparity gap between private and non-private models (\Cref{fig:pub_priv_compare} - left). However, it is unclear whether private training hurts specific classes as suggested by previous work \cite{bagdasaryan2019differential, tran2021differentially}, or whether these disparities arise from the model being overall less accurate. We compare models at $\varepsilon=8$ (around $\sim81\%$ overall accuracy) with non-private models of the same architecture where training was stopped early (around $\sim80\%$ overall accuracy). We see in \Cref{fig:pub_priv_compare} (right) that worst class performance is comparable between private and non-private models.

\begin{figure}
    \centering
    \includegraphics[width=0.8\textwidth]{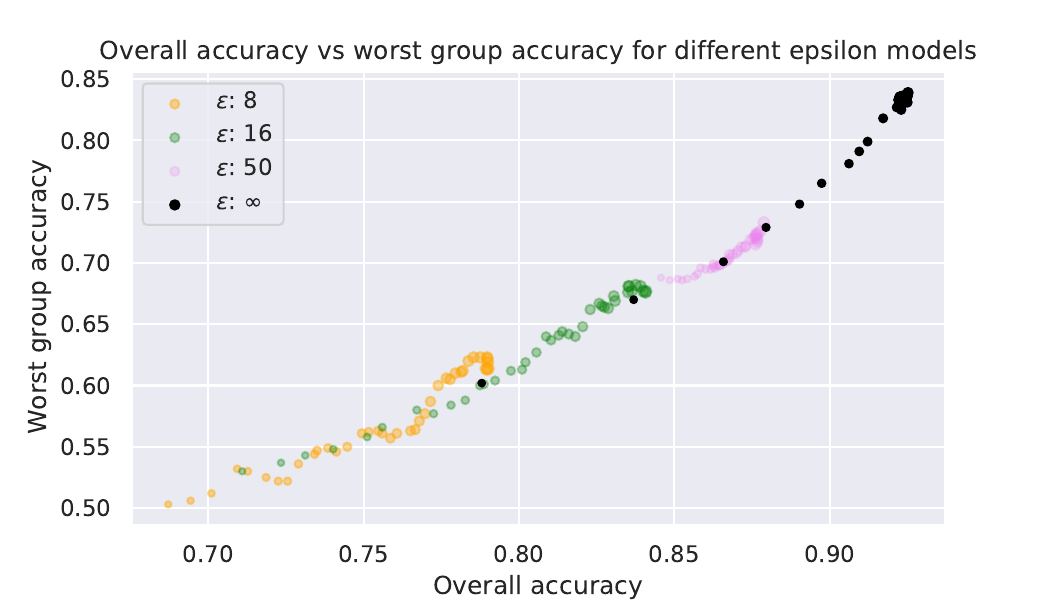}
    \caption{\small Overall accuracy (EMA) vs worst class accuracy (EMA) as private and non-private models are trained. We see that while private models stop at a lower overall accuracy, worst class accuracy of private models are similar to or higher than the non-private model throughout the training trajectory. Each point corresponds to a (EMA) checkpoint in the training process.}
    \label{fig:training_progression}
\end{figure}
Beyond two specific accuracy levels, we can also look at the progression of the worst class accuracy as the overall accuracy increases during model training. \Cref{fig:training_progression} shows the worst class (\textit{cat}) accuracy as training progresses for $\varepsilon=\{8, 16, 50, \infty\}$. What would DP-SGD induced disparity look like in this plot? We would expect to see private models diverging from the trajectory of non-private models by having a smaller slope. This would indicate that the worst group suffers from lower accuracy at the same level of overall accuracy. However, this is not what we observe. We observe private models matching or surpassing non-private models in terms of worst group accuracy at the same overall accuracy. In the setting of balanced data for the CIFAR-10 dataset, we do not observe any evidence that DP-SGD exacerbates disparity at in privacy preserving models with a similar accuracy. 

\textbf{Results: Imbalanced Classes.}
A crucial scenario we also consider is when certain classes in the training dataset are underrepresented. This is the case when \cite{bagdasaryan2019differential} compare private and non-private model performance on a 60,000-image subset of iNaturalist \cite{van2018inaturalist}. The species classes are highly unbalanced with the smallest class being only 1,000 examples. The works \cite{xu2020removing} and \cite{uniyal2021dp} also create an artificially unbalanced dataset by only including 10\% of class 8 images. In our experiments, we look at two classes: \textit{car} and \textit{cat}. We artificially create unbalanced training and validation sets of 50\%, 20\% and 10\% of the \textit{cat} class while leaving the rest of the classes at 100\% representation. We repeat the same setup with the \textit{car} class. We leave the test set to be the same 10,000 balanced examples.  

\begin{figure}
    \centering
    \includegraphics[width=1.0\textwidth]{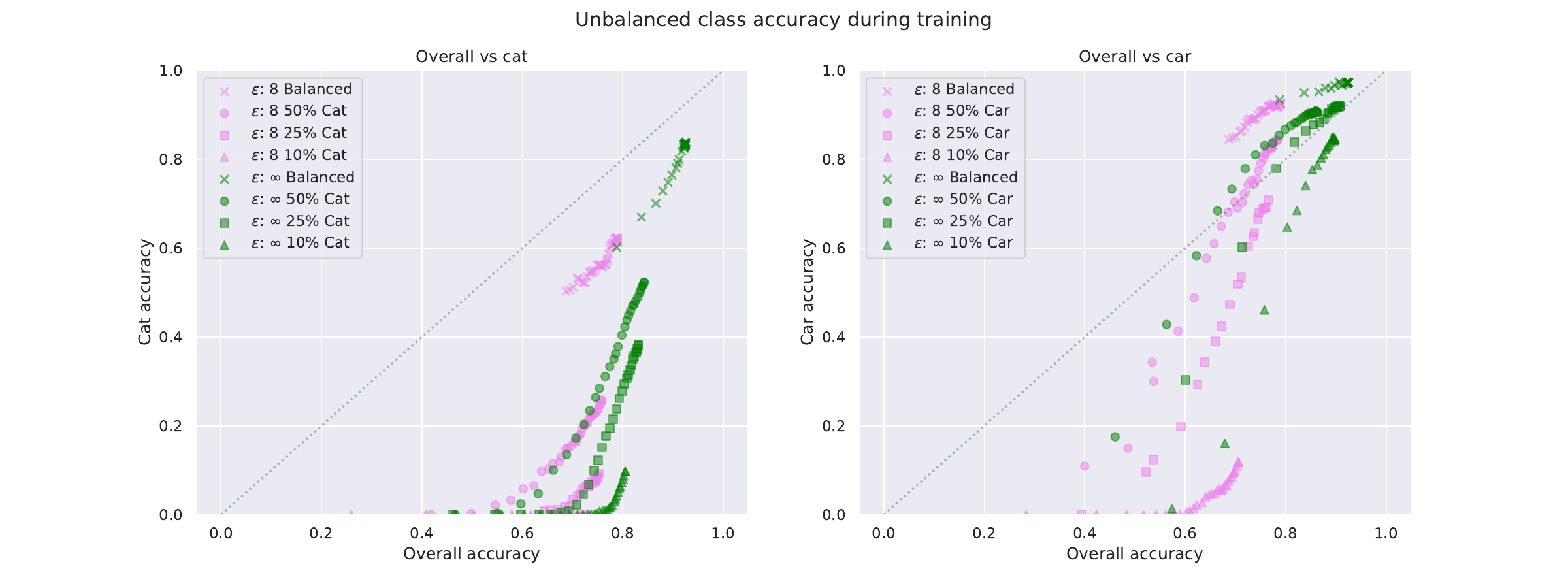}
    \caption{\small (Left) Overall accuracy (EMA) vs cat class accuracy (EMA) at different levels of dataset imbalance for private and non-private models. (Right) Overall accuracy (EMA) vs cat class accuracy (EMA) at different levels of dataset imbalance for private and non-private models. We see that while private models stop at a lower overall accuracy, the ratio (relative disparity) is similar throughout the training trajectory.}
    \label{fig:training_progression_inbalance}
\end{figure}

For various levels of artificially unbalanced data, \Cref{fig:training_progression_inbalance} (left) shows the training progression of $\varepsilon=8$ private models compared to non-private models. We see that in terms of overall accuracy, private models achieve lower accuracy by about 10\%. In terms of unbalanced class accuracy, we see that representation in the training set drastically affects the accuracy of the \textit{cat} class. For 50\% and 25\% representation of the cats in the training data, private models are comparable to non-private models in the regime where non-private model accuracy is between less than 75\%. At 10\% ($\sim$500) cats in the training data, the private model achieves zero accuracy on cats despite a $\sim75$\% overall accuracy. This is due to the resulting model never predicting any example in the \textit{cat} class. When a subclass is small enough, private models may not predict the label at all. In the limit, if only 1 example had the class label \textit{cat}, then we would expect our DP models to guarantee to never predict said example class. While differential privacy guarantees tell us what happens when there is only 1 example, we do not know what level of class under-representation causes a private model at a fixed $\varepsilon$ to eschew predicting said class altogether in real world datasets. 

While the \textit{cat} class is a difficult class for private and non-private models, the \textit{car} class is an easy class. In \Cref{fig:pub_priv_compare}, we see that the car class consistently achieves a higher accuracy than the overall accuracy. To disentangle the effect of class difficulty and class representation in private training, we also train $\varepsilon=8$ models on CIFAR-10 with the cars class artificially under-sampled to 10\%, 25\%, 50\%. In \Cref{fig:training_progression_inbalance} (right), the green cluster in the top right shows that non-private models achieve a very similar final overall/car class accuracies for all the unbalanced datasets. The $\varepsilon=8$ models at 25\% and 50\% imbalance have the same car (underrepresented) class accuracy as non-private models at the same overall accuracy. It is only at 10\% under representation that we observe non-private models at the same overall accuracy have much higher underrepresented class accuracy. This demonstrates that while DP-SGD \textit{does not necessarily imply} worsened disparity when classes are underrepresented, DP-SGD \textit{can} introduce worsened disparities at extreme levels of data imbalance. 

When measuring per-class disparities in private models, the mere presence of data imbalance does not dictate disparity in private models. We compare two classes with different levels of ``difficulty" and find that different levels of under-representation are required for observing differences between private and non-private model disparity.  While our results here are not particularly surprising, they do motivate more systematic empirical analyses of regimes where private model disparities are comparable to non-private models in terms of disparity. The presence of data imbalance should not deter practitioners from applying privacy preserving machine learning due to the risk of class disparities; the level of actual disparity depends on the dataset, the class, and the level of data imbalance. 

\subsection{Picking $\varepsilon$: Rethinking Example Difficulty}
\begin{figure}
    \centering
    \includegraphics[width=1.0\textwidth]{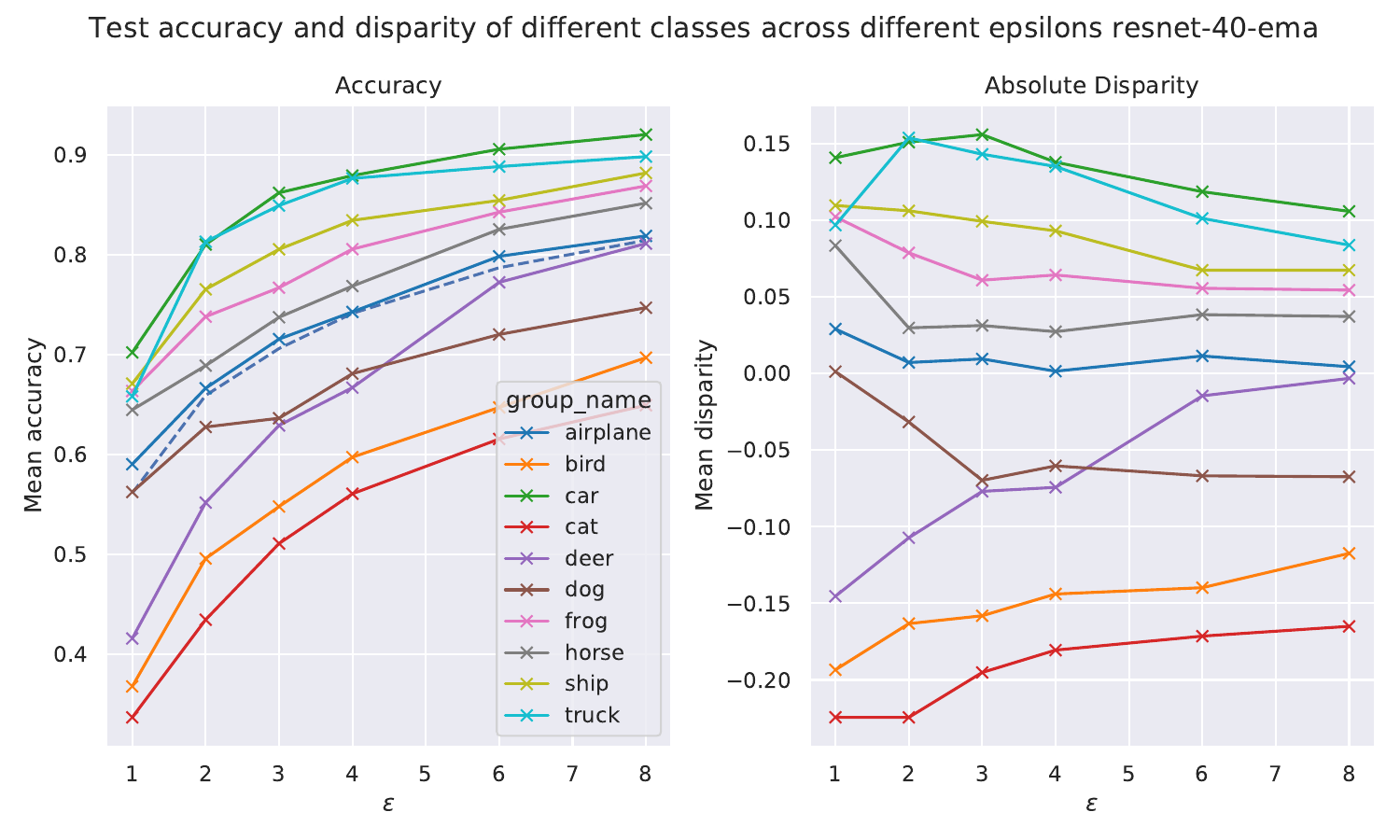}
    \caption{\small Class conditional accuracy (EMA) and absolute disparity averaged across 5 random seeds for different values of $\varepsilon$.}
    \label{fig:across_eps}
\end{figure}

We have focused primarily on $\varepsilon=8$ CIFAR-10 models to make the case the training models with DP-SGD does not necessarily imply more disparity than non-private models. Different settings may require different levels of privacy guarantees. A natural question that arises is what kind of disparities stronger levels of privacy may produce. When comparing private models at different levels of privacy, \Cref{fig:across_eps} illustrates that as $\varepsilon$ increases from 1 to 8, the overall accuracy also increases. This is expected since as $\varepsilon$ approaches $\infty$ we would expect the same accuracy as a non-private model. This is gives further insight into \Cref{fig:training_progression} where we only compare the worst class accuracy against the overall accuracy; in the CIFAR-10 dataset, \textit{cat} is the worst class across different $\varepsilon$. 

Suppose both the \textit{cat} and \textit{bird} classes were not a part of the dataset, then it appears that the most disadvantaged class would depend on $\varepsilon$. When $\varepsilon \leq 4$, we see that \textit{deer} is the worst class while for $\varepsilon \geq 4$, \textit{dog} becomes the worst performing class. There is a cross-over between the class conditional accuracy of \textit{deer} and \textit{dog} as $\varepsilon$ increases. A naive practitioner may observe that \textit{deer} is the worst class at a small $\varepsilon$ and may then focus on improving the performance of the class without realising that the ordering of example difficulty may change. The transition of the \textit{deer} class being more difficult at small $\varepsilon$ and easier at larger $\varepsilon$ is consistent across random seeds and model architectures in our experiments. 

In contrast, during the training progression of non-private models, all classes progress in per-class accuracy without any crossover in ranking. In fact, prior work examining the difficulty of examples \cite{carlini2019distribution, feldman2020neural, jiang2020characterizing} rely on the assumption that example difficulty is not model/algorithm dependent in order to give useful score estimates of difficulty. \cite{kaplun2022deconstructing} observed one type of non-monotonic behaviour by finding examples in the CIFAR-10 dataset which correlate inversely with overall model performance for increasingly complex non-private models. However, this swapping of class difficulty we observe suggests that training private models at increasing $\varepsilon$ may not behave the same as training non-private models of increasing complexity (in terms of parameters or training iterations). Future investigation into measuring and modelling changes in the relative difficulty of examples, and consequently classes, as the level of privacy changes is also crucial for understanding how to select privacy parameters.

\end{appendices}

\end{document}